\documentclass[journal]{IEEEtran}

\usepackage[american]{babel}

\usepackage[caption=false]{subfig}
\usepackage{graphicx}
\usepackage{amsmath,amssymb}
\usepackage{algorithm,algorithmic}

\usepackage[breaklinks=true,colorlinks,bookmarks=false]{hyperref}

\begin{document}

\title{Learning Correspondence Structures for Person Re-identification}

\author{Weiyao Lin,~Yang Shen,~Junchi Yan,~Mingliang Xu,~Jianxin Wu,~Jingdong Wang,~and Ke Lu%
\thanks{The basic idea of this paper appeared in our conference version~\cite{41}. In this
version, we extend our approach by introducing a multi-structure strategy,
carry out detailed analysis, and present more performance results.}
\thanks{W. Lin and Y. Shen are with the Department of Electronic Engineering, Shanghai Jiao Tong University, China (email: \{wylin, shenyang1715\}@sjtu.edu.cn). }
\thanks{J. Yan is with Software Engineering Institute, East China Normal University, Shanghai, China (email: jcyan@sei.ecnu.edu.cn).}
\thanks{M. Xu is with the School of Information Engineering, Zhengzhou University, China (email: iexumingliang@zzu.edu.cn)}
\thanks{J. Wu is with the National Key Laboratory for Novel Software Technology, Nanjing University, China (email: wujx2001@nju.edu.cn).}
\thanks{J. Wang is with Microsoft Research, Beijing, China (email: jingdw@microsoft.com).}
\thanks{K. Lu is with the College of Computer and Information Technology, China Three Gorges University, Yichang, China, also with the University of Chinese Academy of Sciences, Beijing, China (email: luk@ucas.ac.cn).}
}

\maketitle

\begin{abstract}
This paper addresses the problem of handling spatial misalignments due to camera-view changes or human-pose variations in person re-identification. We first introduce a boosting-based approach to learn a correspondence structure which indicates the patch-wise matching probabilities between images from a target camera pair. The learned correspondence structure can not only capture the spatial correspondence pattern between cameras but also handle the viewpoint or human-pose variation in individual images. We further introduce a global constraint-based matching process. It integrates a global matching constraint over the learned correspondence structure to exclude cross-view misalignments during the image patch matching process, hence achieving a more reliable matching score between images. Finally, we also extend our approach by introducing a multi-structure scheme, which learns a set of local correspondence structures to capture the spatial correspondence sub-patterns between a camera pair, so as to handle the spatial misalignments between individual images in a more precise way. Experimental results on various datasets demonstrate the effectiveness of our approach. The project page for this paper is available at \url{http://min.sjtu.edu.cn/lwydemo/personReID.htm}
\end{abstract}

\begin{IEEEkeywords}
Person Re-identification, Correspondence Structure Learning, Spatial Misalignment
\end{IEEEkeywords}

\section{Introduction} \label{section:introduction}

Person re-identification (Re-ID) is of increasing importance in visual surveillance. The goal of person Re-ID is to identify a specific person indicated by a probe image from a set of gallery images captured from cross-view cameras (i.e., cameras that are non-overlapping and different from the probe image's camera)~\cite{SalientColor, rankboost, color_in}. It remains challenging due to the large appearance changes in different camera views and the interferences from background or object occlusion.

\begin{figure}[t]
  \centering
  \vspace{-2mm}
  \subfloat[]{\includegraphics[width=2.88cm,height=2.7cm]{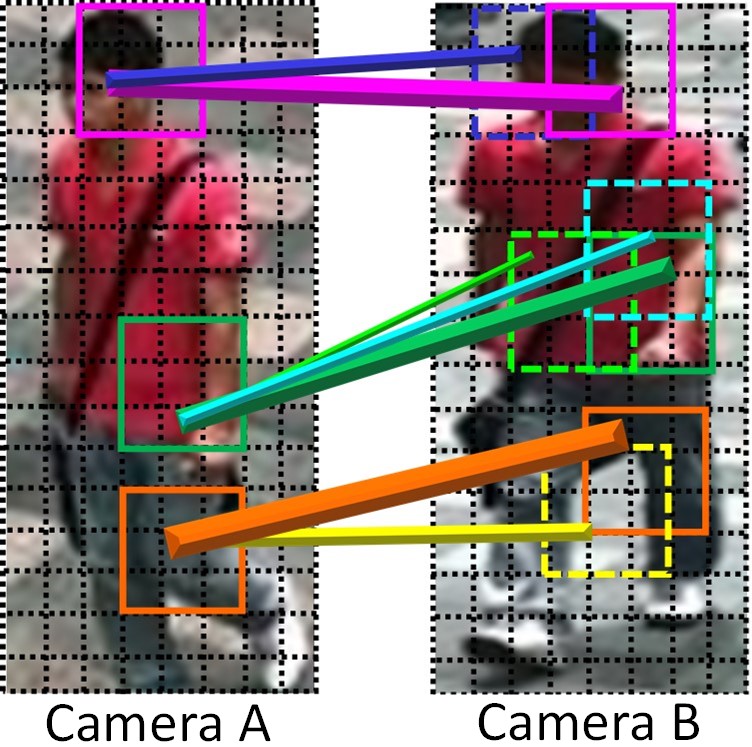}   \label{fig:mapping structure_example_a}}
  \hspace{0.1mm}
  \subfloat[]{\includegraphics[width=2.88cm,height=2.7cm]{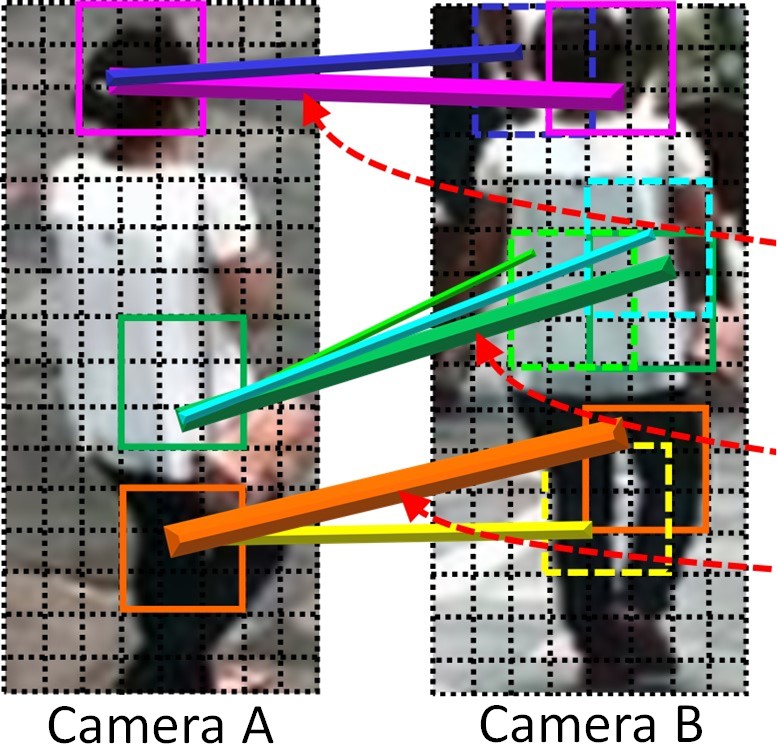}   \label{fig:mapping structure_example_b}}
  \hspace{-2.8mm}
  \subfloat[]{\includegraphics[width=2.53cm,height=2.7cm]{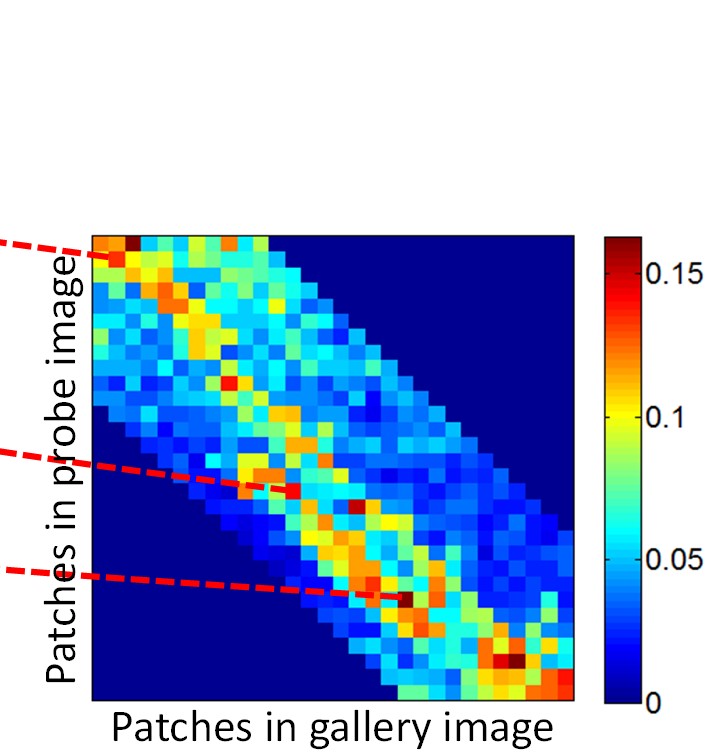}   \label{fig:mapping structure_example_c}}
  \caption{(a) and (b): Two examples of using a correspondence structure to handle spatial misalignments between images from a camera pair. Images are obtained from the same camera pair: $A$ and $B$. The colored squares represent sample patches in each image while the lines between images indicate the matching probability between patches (line width is proportional to the probability values). (c): The correspondence structure matrix including all patch matching probabilities between $A$ and $B$ (the matrix is down-sampled for a clearer illustration). (Best viewed in color)}   \label{fig:mapping structure_example}
\end{figure}

One major challenge for person Re-ID is the uncontrolled spatial misalignment between images due to camera-view changes or human-pose variations. For example, in Fig.~\ref{fig:mapping structure_example_a}, the green patch located in the lower part in camera $A$'s image corresponds to patches from the upper part in camera $B$'s image. However, most existing works \cite{SalientColor, rankboost, color_in,3,eiml,1,kernel-based,prid,zheng2015iccv,chen2015cvpr} focus on handling the overall appearance variations between images, while the spatial misalignment among images' local patches is not addressed. Although some patch-based methods \cite{part,12,6,part2} address the spatial misalignment problem by decomposing images into patches and performing an online patch-level matching, their performances are often restrained by the online matching process which is easily affected by the mismatched patches due to similar appearance or occlusion.

In this paper, we argue that due to the stable setting of most cameras (e.g., fixed camera angle or location), each camera has a stable constraint on the spatial configuration of its captured images. For example, images in Fig.~\ref{fig:mapping structure_example_a} and~\ref{fig:mapping structure_example_b} are obtained from the same camera pair: $A$ and $B$. Due to the constraint from camera angle difference, body parts in camera $A$'s images are located at lower places than those in camera $B$, implying a lower-to-upper correspondence pattern between them. Meanwhile, constraints from camera locations can also be observed. Camera $A$ (which monitors an exit region) includes more side-view images, while camera $B$ (monitoring a road) shows more front or back-view images. This further results in a high probability of side-to-front/back correspondence pattern.

Based on this intuition, we propose to learn a correspondence structure (i.e., a matrix including \emph{all} patch-wise matching probabilities between a camera pair, as Fig.~\ref{fig:mapping structure_example_c}) to encode the spatial correspondence pattern constrained by a camera pair, and utilize it to guide the patch matching and matching score calculation processes between images. With this correspondence structure, spatial misalignments can be suitably handled and patch matching results are less interfered by the confusion from appearance or occlusion. In order to model human-pose variations or local viewpoint changes inside a camera view, the correspondence structure for each patch is described by a one-to-many graph whose weights indicate the matching probabilities between patches, as in Fig.~\ref{fig:mapping structure_example}. Besides, a global constraint is also integrated during the patch matching process, so as to achieve a more reliable matching score between images. 

Moreover, since people often show different poses inside a camera view, the spatial correspondence pattern between a camera pair may be further divided and modeled by a set of sub-patterns according to these pose variations (e.g., Fig.~\ref{fig:mapping structure_example} can be divided into a left-to-front correspondence sub-pattern \subref{fig:mapping structure_example_a} and a right-to-back correspondence sub-pattern \subref{fig:mapping structure_example_b}). Therefore, we further extend our approach by introducing a set of local correspondence structures to capture various correspondence sub-patterns between a camera pair. In this way, the spatial misalignments between individual images can be modeled and handled in a more precise way.

In summary, our contributions to Re-ID are four folds.
\begin{enumerate}
 \item We introduce a correspondence structure to encode cross-view correspondence pattern between cameras, and develop a global constraint-based matching process by combining a global constraint with the correspondence structure to exclude spatial misalignments between images. These two components establish a novel framework for addressing the Re-ID problem.
 \item Under this framework, we propose a boosting-based approach to learn a suitable correspondence structure between a camera pair. The learned correspondence structure can not only capture the spatial correspondence pattern between cameras but also handle the viewpoint or human-pose variation in individual images.
 \item We further extend our approach by introducing a multi-structure scheme, which first learns a set of local correspondence structures to capture the spatial correspondence sub-patterns between a camera pair, and then adaptively selects suitable correspondence structures to handle the spatial misalignments between individual images. Thus, image-wise spatial misalignments can be modeled and excluded in a more precise way.
 \item We release a new and challenging benchmark \textsc{Road dataset} for person Re-ID which includes large variation of human pose and camera angle.
\end{enumerate}

The rest of this paper is organized as follows. Section~\ref{section:related_work} reviews related works. Section~\ref{section:framework} describes the framework of the proposed approach. Sections~\ref{section:mapping structure learning} to~\ref{section:learning} describe the details of our proposed global constraint-based matching process and boosting-based learning approach, respectively. Section~\ref{section:multiple correspondence structure} describes the details of extending our approach with multi-structure scheme. Section~\ref{section:experimental evaluation} shows the experimental results and Section~\ref{section:conclusion} concludes the paper.

\section{Related Works} \label{section:related_work}

Many person re-identification methods have been proposed. Most of them focus on developing suitable feature representations about humans' appearance \cite{SalientColor, rankboost, color_in, 3, OTF}, or finding proper metrics to measure the cross-view appearance similarity between images \cite{eiml, 1, kernel-based, prid, ma2014tip, chen2015tip, zhang2016learning}. Recently, due to the effectiveness of deep neural networks in learning discriminative features, many deep Re-ID methods are proposed which utilize deep neural networks to enhance the reliability of feature representations or similarity metrics~\cite{faqiang2016cvpr, zhang2015tip, chen2016tip, ahmed2015cvpr, ding2015pr, cheng2016person, xiao2016learning, wu2016enhanced}. Since most of these works do not effectively model the spatial misalignment among local patches inside images, their performances are still impacted by the interferences from viewpoint or human-pose changes.

In order to address the spatial misalignment problem, some patch-based methods are proposed \cite{2,part,part2,12,6,8,10,SLEPK,added[1],added[2]} which decompose images into patches and perform an online patch-level matching to exclude patch-wise misalignments. In \cite{2,10,added[1]}, a human body in an image is first parsed into semantic parts (e.g., head and torso) or patch clusters. Then, similarity matching is performed between the corresponding semantic parts or patch clusters. Since these methods are highly dependent on the accuracy of body parser or patch clustering, they have limitations in scenarios where the body parser does not work reliably or the patch clustering results are less satisfactory.

In \cite{part}, Oreifej et al. divide images into patches according to appearance consistencies and utilize the Earth Movers Distance (EMD) to measure the overall similarity among the extracted patches. However, since the spatial correlation among patches are ignored during similarity calculation, the method is easily affected by the mismatched patches with similar appearance. Although Ma et al. \cite{12} introduce a body prior constraint to avoid mismatching between distant patches, the problem is still not well addressed, especially for the mismatching between closely located patches.

To reduce the effect of patch-wise mismatching, some saliency-based approaches \cite{6,8,added[2],added[3]} are recently proposed, which estimate the saliency distribution relationship between images and utilize it to control the patch-wise matching process. Following the similar line, Bak and Carr~\cite{added[2]} introduce a deformable model to obtain a set of weights to guide the patch matching process. Bak et al.~\cite{added[3]} further introduce a hand-crafted Epanechnikov kernel to determine the weight distribution of patches. Although these methods consider the correspondence constraint between patches, our approach differs from them in: (1) our approach focuses on constructing a correspondence structure where patch-wise matching parameters are jointly decided by both matched patches. Comparatively, the matching weights in most of the saliency-based approaches \cite{8,added[2]} are only controlled by patches in the probe-image (probe patch). (2) Our approach models patch-wise correspondence by a one-to-many graph such that each probe patch will trigger multiple matches during the patch matching process. In contrast, the saliency-based approaches only select one best-matched patch for each probe patch. (3) Our approach introduces a global constraint to control the patch-wise matching result while the patch matching result in saliency-based methods is locally decided by choosing the best matched within a neighborhood set.

Besides patch-based approaches, some other methods are also proposed which aim to simultaneously find discriminative features and handle the spatial misalignment problem. In \cite{added[4]}, the correspondence relationship between images is captured by a hierarchical matching process integrated in an end-to-end deep learning framework. In \cite{zhang2014eccv, PRISM}, image-wise similarities are measured by kernelized visual word co-occurrence matrices where a latent spatial kernel is integrated to handle spatial displacements. Since these methods give more weights on small displacements and less weights on large displacements, they are more suitable to handle spatial misalignments with relatively small displacements and may have limitations for very large displacements.

\section{Overview of The Approach} \label{section:framework}

\begin{figure}[t]
  \centering
  \includegraphics[width=8.7 cm, height=4.1 cm]{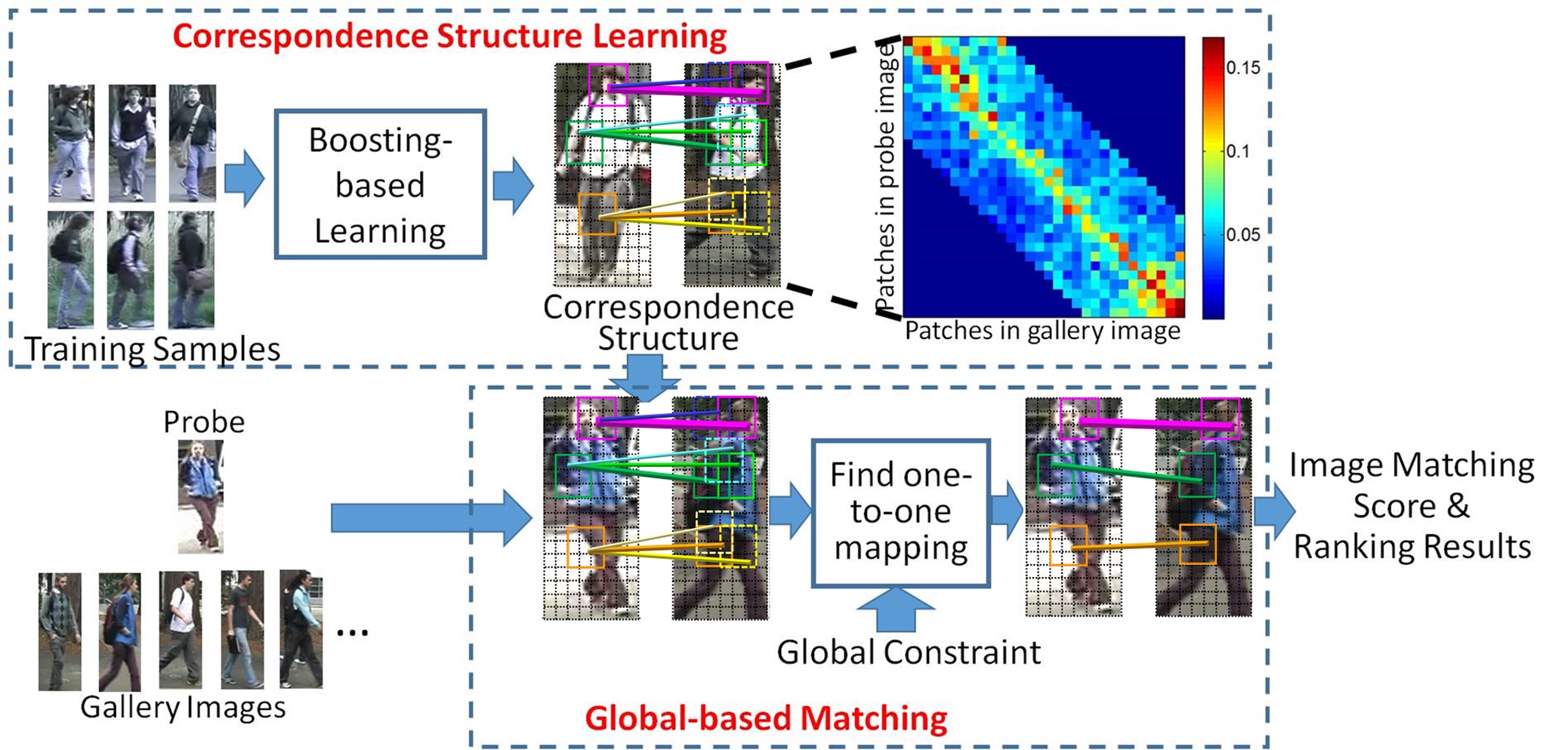}
  \caption{Framework of the proposed approach.}  \label{fig:framework}
\end{figure}

The framework of our approach is shown in Fig.~\ref{fig:framework}. During the training process, which is detailed in Section~\ref{section:learning}, we present a boosting-based process to learn the correspondence structure between the target camera pair. During the prediction stage, which is detailed in Section~\ref{section:mapping structure learning}, given a probe image and a set of gallery images, we use the correspondence structure to evaluate the patch correlations between the probe image and each gallery image, and find the optimal one-to-one mapping between patches, and accordingly the matching score. The Re-ID result is achieved by ranking gallery images according to their matching scores.

\section{Person Re-ID via Correspondence Structure} \label{section:mapping structure learning}

This section introduces the concept of correspondence structure, show the scheme of computing the patch correlation using the correspondence structure, and finally present the patch-wise mapping method to compute the matching score between the probe image and a gallery image.

\subsection{Correspondence structure}

The correspondence structure, $\mathbf{\Theta}_{A,B}$, encodes the spatial correspondence distribution between a pair of cameras, $A$ and $B$. In our problem, we adopt a discrete distribution, which is a set of patch-wise matching probabilities, $\mathbf{\Theta}_{A, B} = \{P_{i,B}\}_{i=1}^{N_A}$, where $N_A$ is the number of patches of an image in camera $A$. $P_{i,B} = \{P_{i1}, P_{i2}, \dots, P_{i{N_B}}\}$ describes the correspondence distribution in an image from camera $B$ for the $i$th patch $x_i$ of an image captured from camera $A$, where $N_B$ is the number of patches of an image in $B$. The illustrations of the correspondence distribution are shown on the top-right of Fig.~\ref{fig:framework}.

The definition of the matching probabilities in the correspondence structure only depends on a camera pair and are independent of specific images. In the correspondence structure, it is possible that one patch in camera $A$ is highly correlated to multiple patches in camera $B$, so as to handle human-pose variations and local viewpoint changes. 

\subsection{Patch correlation}

Given a probe image $U$ in camera $A$ and a gallery image $V$ in camera $B$, the patch-wise correlation between $U$ and $V$, $C(x_i,y_j)$ ($x_i\in U$, $y_j\in V$), is computed from both the correspondence structure between two cameras and the visual features:
\begin{equation}
 C(x_i, y_j) = \lambda_{T_c}(P_{ij}) \cdot \log \Phi(\mathbf{f}_{x_i},\mathbf{f}_{y_j}; x_i, y_j).
\label{equation:equ1}
\end{equation}
Here $x_i$ and $y_j$ are $i$th and $j$th patch in images $U$ and $V$; $\mathbf{f}_{x_i}$ and $\mathbf{f}_{y_j}$ are the feature vectors for $x_i$ and $y_j$, respectively. {$P_{ij}$ is the correspondence structure between $U$ and $V$. $\lambda_{T_c}(P_{ij}) = 1$, if $P_{ij}>T_c$}, and $0$ otherwise,
and $T_c=0.05$ is a threshold. $\Phi(\mathbf{f}_{x_i},\mathbf{f}_{y_j};x_i, y_j)$ is the correspondence-structure-controlled similarity
between $x_i$ and $y_j$,
\begin{equation}
\Phi(\mathbf{f}_{x_i},\mathbf{f}_{y_j}; x_i, y_j) = \Phi_{ij} ( \mathbf{f}_{x_i},\mathbf{f}_{y_j})  P_{ij} ,
\label{equation:equ2}
\end{equation}
where $\Phi_{ij} ( \mathbf{f}_{x_i},\mathbf{f}_{y_j})$ is the similarity between $x_i$ and $y_j$.

The correspondence structure $P_{ij}$ in Equations~\ref{equation:equ1} and~\ref{equation:equ2} is used to adjust the appearance similarity $\Phi_{ij}(\mathbf{f}_{x_i},\mathbf{f}_{y_j})$ such that a more reliable patch-wise correlation strength can be achieved. The thresholding term $\lambda_{T_c}(P_{ij})$ is introduced to exclude the patch-wise correlation with a low correspondence probability, which effectively reduces the interferences from mismatched patches with similar appearance.

The patch-wise appearance similarity $\Phi_{ij} ( \mathbf{f}_{x_i},\mathbf{f}_{y_j})$ in Eq.~\ref{equation:equ2} can be achieved by many off-the-shelf methods~\cite{6,8,patch_match}. In this paper, we extract Dense SIFT and Dense Color Histogram~\cite{6} from each patch and utilize the KISSME metric~\cite{1} as the basic distance metric to compute $\Phi_{ij} ( \mathbf{f}_{x_i},\mathbf{f}_{y_j})$. Note that in order to achieve better performance~\cite{added[2]}, we learn multiple metrics so as to adaptively model the local patch-wise correlations at different locations. Specifically, we train an independent KISSME metric $\Phi_{ij}$ for each set of patch-pair locations $\{x_i,y_j\}$. Besides, it should also be noted that both the feature extraction and distance metric learning parts can be replaced by other state-of-the-art methods. For example, in the experiments in Section~\ref{section:experimental evaluation}, we also show the results by replacing KISSME with the kLFDA~\cite{kernel-based} and KMFA-$R_{\chi^2}$~\cite{chen2015mirror} metrics to demonstrate the effectiveness of our approach.

\subsection{Patch-wise mapping} \label{Patch-wise mapping}

\begin{figure}[t]
  \centering
  \subfloat[]{\includegraphics[width=3.1cm, height=2.8cm]{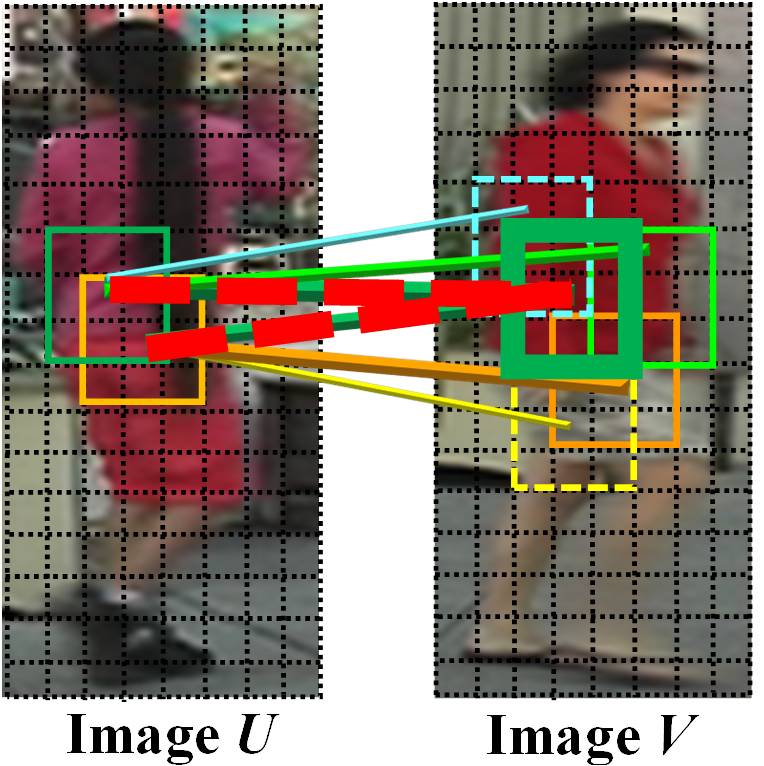}     \label{fig:one to one matching_a}}
  \hspace{2mm} \vrule  \hspace{3mm}
  \subfloat[]{\includegraphics[width=3.1cm, height=2.8cm]{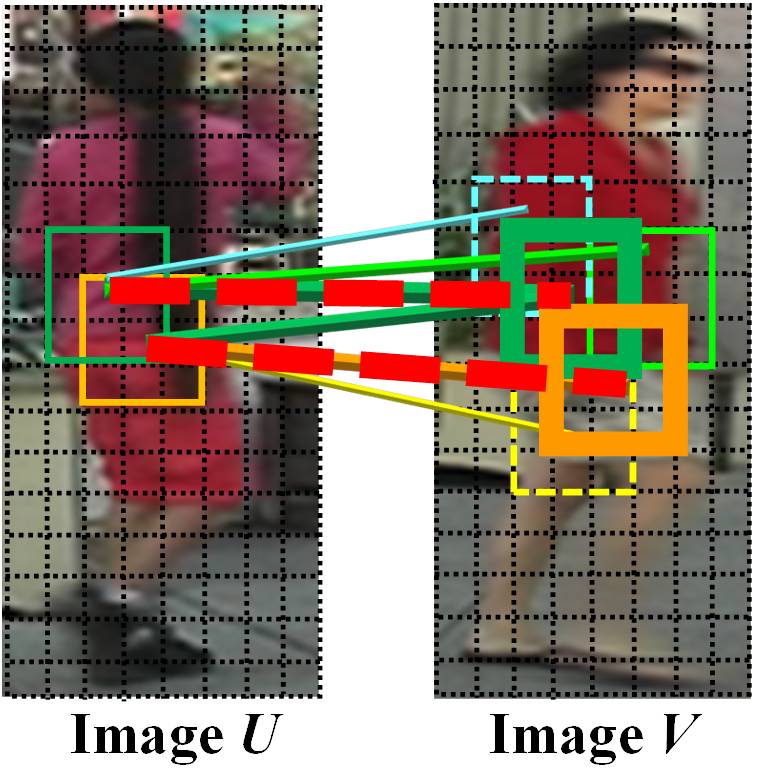}     \label{fig:one to one matching_b}}
  \caption{Patch matching result (a) by locally finding the largest correlation strength $C(x_i,y_j)$ for each patch and (b) by using a global constraint. The red dashed lines indicate the final patch matching results and the colored solid lines are the matching probabilities in the correspondence structure. (Best viewed in color)}   \label{fig:one to one matching}
\end{figure}

After obtaining the alignment-enhanced correlation strength $C(x_i,y_j)$, we can find a best-matched patch in image $V$ for each patch in $U$ and herein calculate the final image matching score. To compute $C(x_i,y_j)$ of testing image pairs, we only consider the potential matching patches within a searching range $\mathbf{R}(x_i)$ around the chosen probe patch $x_i$. That is, $\mathbf{R}(x_i)=\{y_j|d(y_i,y_j)<T_d\}$, where $y_i$ is $x_i$'s co-located patch in camera $B$, $d(y_i,y_j)$ is the distance between patches $y_i$ and $y_j$ (cf. Eq.~\ref{equation:equ5a}), and $T_d=32$ in this paper. However, locally finding the largest $C(x_i,y_j)$ may still create mismatches among patch pairs with high matching probabilities. For example, Fig.~\ref{fig:one to one matching_a} shows an image pair $U$ and $V$ containing different people. When locally searching for the largest $C(x_i,y_j)$, the yellow patch in $U$ will be mismatched to the bold-green patch in $V$ since they have both large appearance similarity and high matching probability. This mismatch unsuitably increases the matching score between $U$ and $V$.

To address this problem, we introduce a global one-to-one mapping constraint and solve the resulting linear assignment task~\cite{hungary} to find the best matching:
\begin{align}
&\mathbf{\Omega}^*_{U,V} = \mathop{\arg\max}_{\mathbf{\Omega}_{U,V}}{\sum_{\{x_i,y_j\}\in\mathbf{\Omega}_{U,V}}{C(x_i,y_j)}}
\label{equation:equ3}\\
\text{s.t.}~& x_i\neq x_s, y_j\neq y_t ~~~\forall~ \{x_i,y_j\}, \{x_s,y_t\}\in\mathbf{\Omega}_{U,V} \notag
\end{align}
where $\mathbf{\Omega}^*_{U,V}$ is the set of the best patch matching result between images $U$ and $V$. $\{x_i,y_j\}$ and $\{x_s,y_t\}$ are two matched patch pairs in $\mathbf{\Omega}$. According to Eq.~\ref{equation:equ3}, we want to find the best patch matching result $\mathbf{\Omega}^*_{U,V}$ that maximizes the total image matching score
\begin{align}
\psi_{U,V}=\sum_{\{x_i,y_j\}\in\mathbf{\Omega}_{U,V}}{C(x_i,y_j)}, \label{equation:matchingscore}
\end{align}
given that each patch in $U$ can only be matched to one patch in $V$ and vice versa.

Eq.~\ref{equation:equ3} can be solved by the Hungarian method~\cite{hungary}. Fig.~\ref{fig:one to one matching_b} shows an example of the patch matching result by Eq.~\ref{equation:equ3}. From Fig.~\ref{fig:one to one matching_b}, it is clear that by the inclusion of a global constraint, local mismatches can be effectively reduced and a more reliable image matching score can be achieved. Based on the above process, we can calculate the image matching scores $\psi$ between a probe image and all gallery images in a cross-view camera, and rank the gallery images accordingly to achieve the final Re-ID result~\cite{12}.

\section{Correspondence Structure Learning} \label{section:learning}

\subsection{Objective function}

Given a set of probe images $\{U_\alpha\}$ from camera $A$ and their corresponding cross-view images $\{V_\beta\}$ from camera $B$ in the training set, we learn the optimal correspondence structure $\mathbf{\Theta}_{A,B}^*$ between cameras $A$ and $B$ so that the correct match image is ranked before the incorrect match images
in terms of the matching scores. The formulation is:
\begin{align}
  \mathop{\min}_{\mathbf{\Theta}_{A,B}}
    \sum_{U_{\alpha}} R(V_{\alpha'}; \psi_{U_\alpha,V_{\alpha'}}(\mathbf{\Theta}_{A,B}),  \mathbf{\Psi}_{U_\alpha,V_{\beta\neq\alpha'}} (\mathbf{\Theta}_{A,B})), \label{equation:equ4}
\end{align}
in which $V_{\alpha'}$ is the correct match gallery image of the probe image ${U_{\alpha}}$. $\psi_{U_\alpha,V_{\alpha'}}(\mathbf{\Theta}_{A,B})$ (as computed from Eq.~\ref{equation:matchingscore}) is the matching score between $U_\alpha$ and $V_{\alpha'}$ and $\mathbf{\Psi}_{U_\alpha,V_{\beta\neq\alpha'}} (\mathbf{\Theta}_{A,B})$
is the set of matching scores of all incorrect matches. $R(V_{\alpha'}; \psi_{U_\alpha,V_{\alpha'}}(\mathbf{\Theta}_{A,B}), \mathbf{\Psi}_{U_\alpha,V_{\beta\neq\alpha'}} (\mathbf{\Theta}_{A,B}))$ is the rank of $V_{\alpha'}$ among all the gallery images according to the matching scores. Intuitively, the penalty is the smallest if the rank is $1$, i.e., the matching score of $V_{\alpha'}$ is the greatest.

According to Eq.~\ref{equation:equ4}, the correspondence structure that can achieve the best Re-ID result on the training set will be selected as the optimal correspondence structure. However, the optimization is not easy as the matching score (Eq.~\ref{equation:matchingscore}) is complicated. We present an approximate solution, a boosting-based process, to solve this problem, which utilizes an iterative update process to gradually search for better correspondence structures fitting Eq.~\ref{equation:equ4}.

\subsection{Boosting-based learning} \label{section:boost-learning}

\begin{figure}[t]
  \centering
  \subfloat[]{\includegraphics[width=0.3\columnwidth]{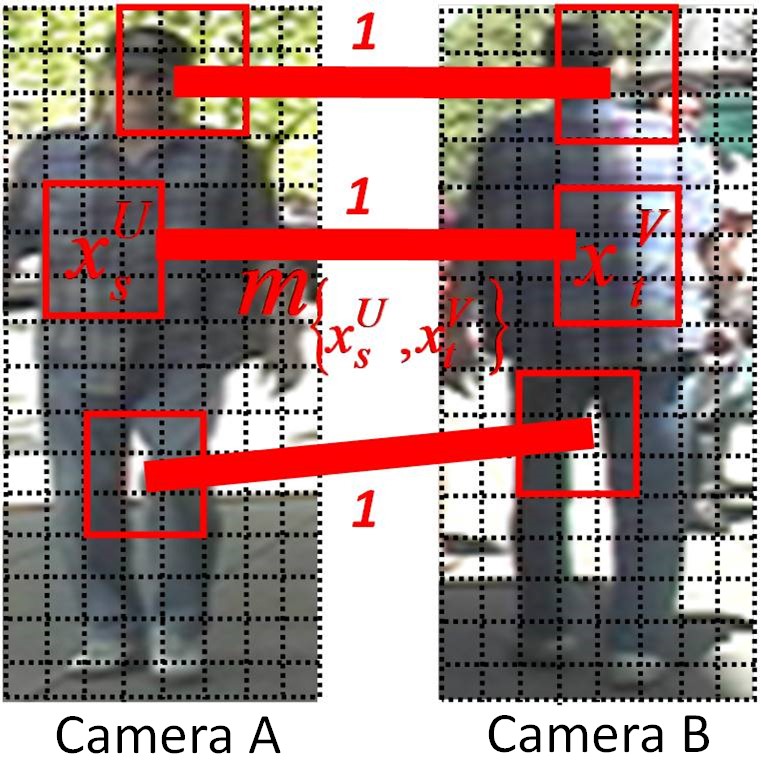}   \label{fig:one to many matching a}}
  \vrule \hspace{.5mm}
  \subfloat[]{\includegraphics[width=0.3\columnwidth]{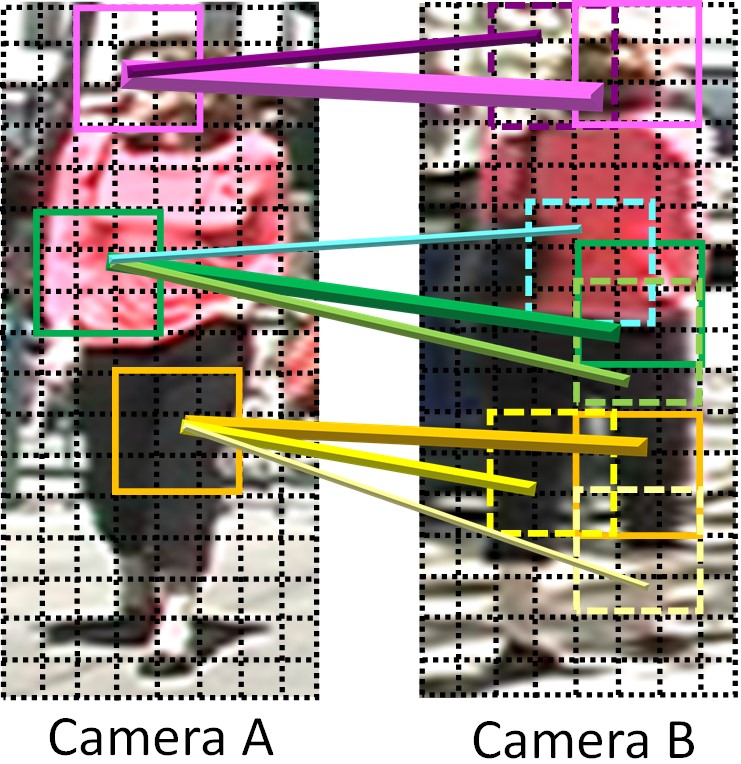}   \label{fig:one to many matching b}}
  \vrule \hspace{.5mm}
  \subfloat[]{\includegraphics[width=0.3\columnwidth]{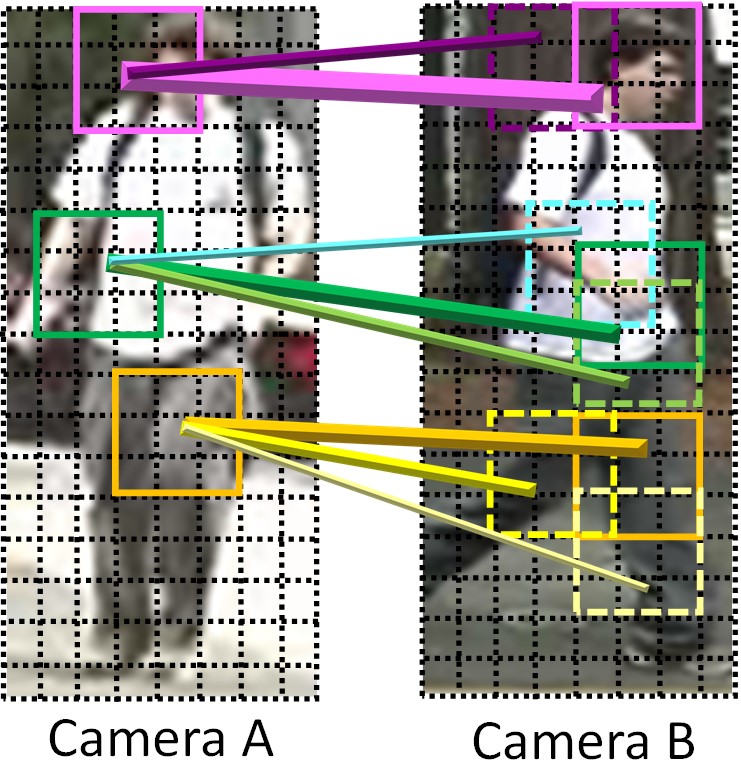}   \label{fig:one to many matching c}}
  \caption{(a): An example of binary mapping structure (the red lines with weight $1$ indicate that the corresponding patches are connected). (b)-(c): Examples of the correspondence structures learned by our approach for the VIPeR~\cite{viper} data. The line widths in (b)-(c) are proportional to the patch-wise probability values. (Best viewed in color)}    \label{fig:one to many matching}
\end{figure}

The boosting-based approach utilizes a progressive way to find the best correspondence structure with the help of \emph{binary mapping structures}. A binary mapping structure is similar to the correspondence structure except that it simply uses $0$ or $1$ instead of matching probabilities to indicate the connectivity or linkage between patches, cf. Fig.~\ref{fig:one to many matching a}. It can be viewed as a simplified version of the correspondence structure which includes rough information about the cross-view correspondence pattern.

Since binary mapping structures only include simple connectivity information among patches, their optimal solutions are tractable for individual probe images. Therefore, by searching for the optimal binary mapping structures for different probe images and using them to progressively update the correspondence structure, suitable cross-view correspondence patterns can be achieved.

The entire boosting-based learning process can be describe by the following steps, which are summarized in Algorithm~\ref{algorithm:progressive updating}.

\textbf{Finding the optimal binary mapping structure.} For each training probe image $U_{\alpha}$, we aim to find the optimal binary mapping structure $\mathbf{M}_\alpha$ such that the rank order of $U_{\alpha}$'s correct match image $V_{\alpha'}$ is minimized under $\mathbf{M}_\alpha$. In order to reduce the computation complexity of finding $\mathbf{M}_\alpha$, we introduce an approximate strategy. Specifically, we first create multiple candidate binary mapping structures under different searching ranges (from $26$ to $32$) by adjacency-constrained search ~\cite{6}, and then select the one which minimizes the rank order of $V_{\alpha'}$ and use it as the approximated optimal binary mapping structure $\mathbf{M}_\alpha$. Note that we find one optimal binary mapping structure for each probe image such that the obtained binary mapping structures can include local cross-view correspondence clues in different training samples.

\begin{figure*}[t]
  \vspace{-4mm}
  \centering
   \subfloat[]{\includegraphics[width=2.62cm,height=2.5cm]{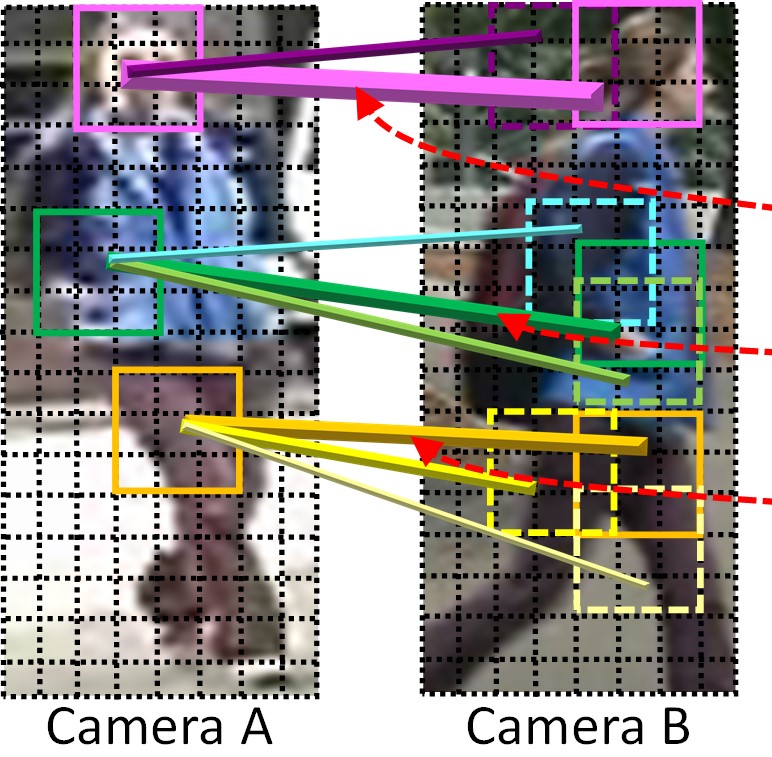}   \label{fig:learned structure a}}
  \hspace{-2.61mm}
  \subfloat[]{\includegraphics[width=2.62cm,height=2.5cm]{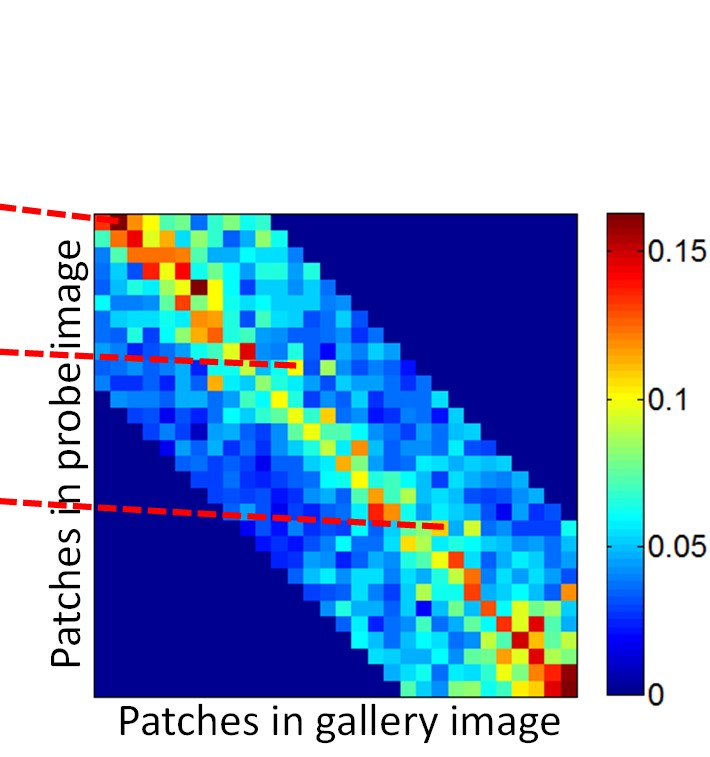}    \label{fig:learned structure b}}
  \hspace{0.3mm}
  \subfloat[]{\includegraphics[width=2.6cm,height=1.9cm]{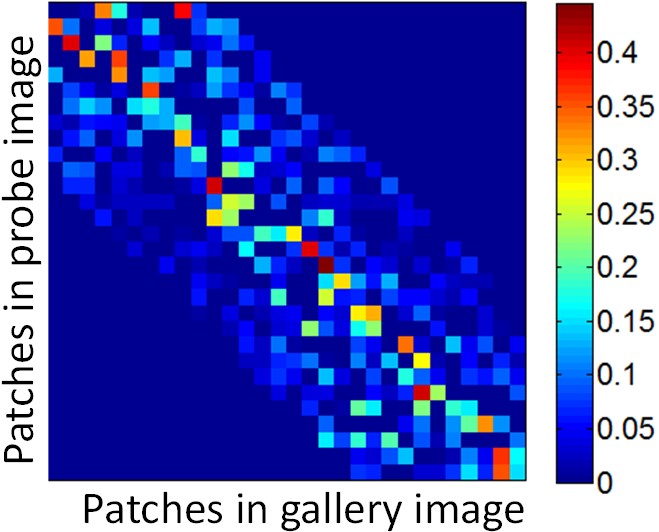}   \label{fig:learned structure c}}
  \hspace{0.3mm}
  \subfloat[]{\includegraphics[width=2.62cm,height=2.5cm]{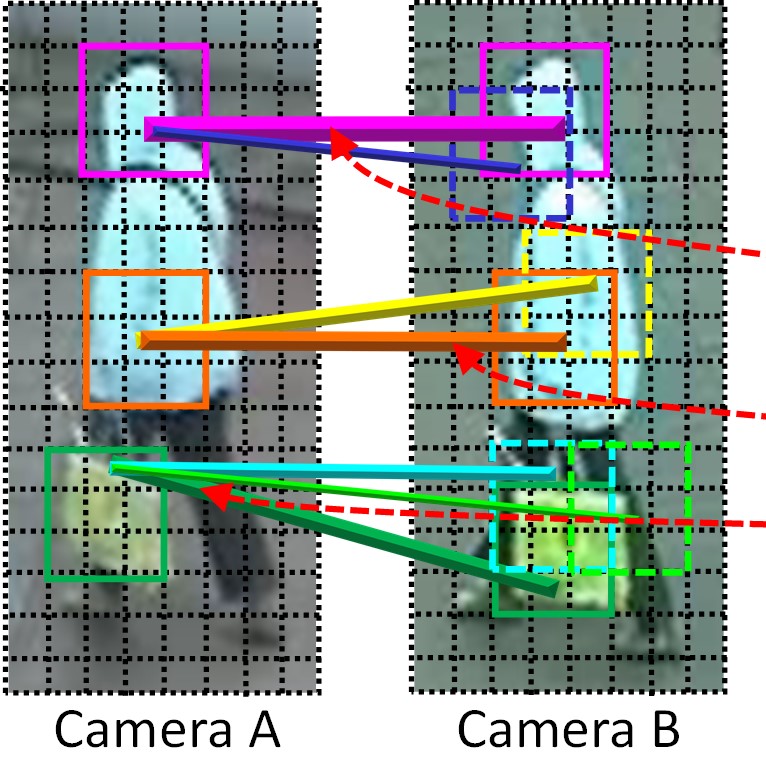}    \label{fig:learned structure d}}
  \hspace{-2.61mm}
  \subfloat[]{\includegraphics[width=2.62cm,height=2.5cm]{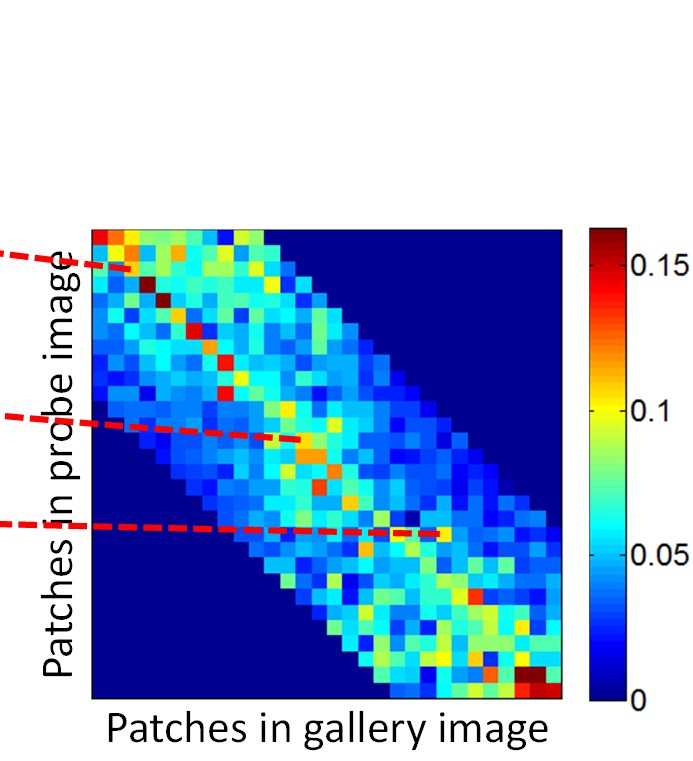}    \label{fig:learned structure e}}
  \hspace{0.3mm}
  \subfloat[]{\includegraphics[width=2.6cm,height=1.9cm]{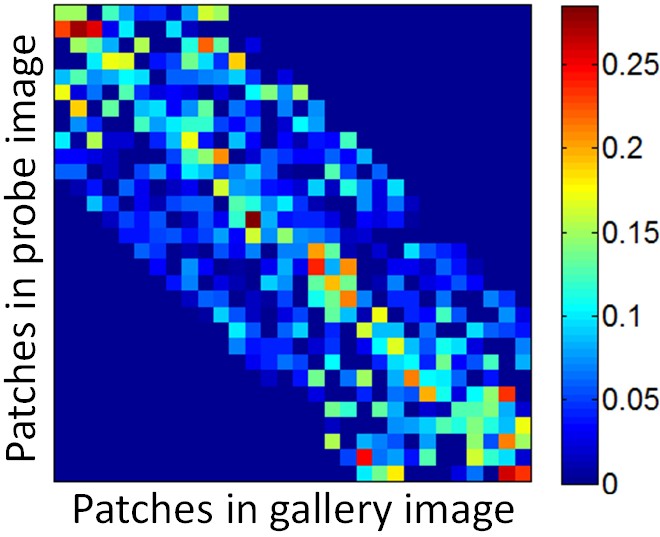}   \label{fig:learned structure f}}
  \\
  \subfloat[]{\includegraphics[width=2.62cm,height=2.5cm]{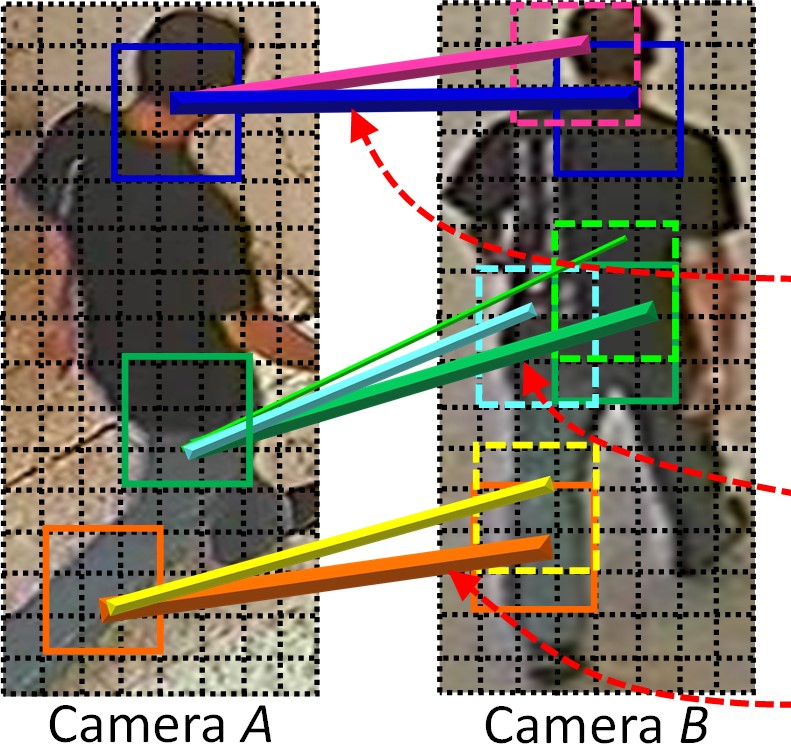}    \label{fig:learned structure g}}
  \hspace{-2.61mm}
  \subfloat[]{\includegraphics[width=2.62cm,height=2.5cm]{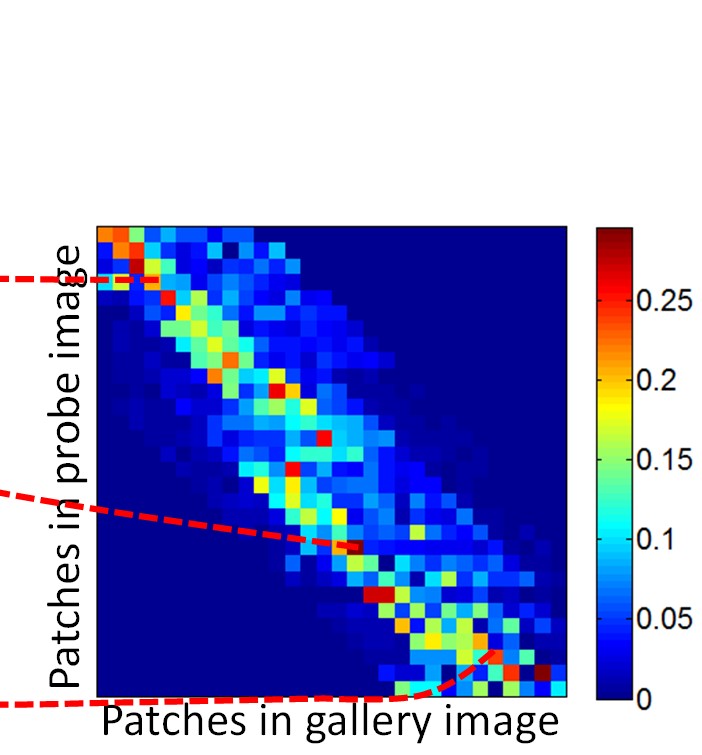}    \label{fig:learned structure h}}
  \hspace{0.3mm}
  \subfloat[]{\includegraphics[width=2.6cm,height=1.9cm]{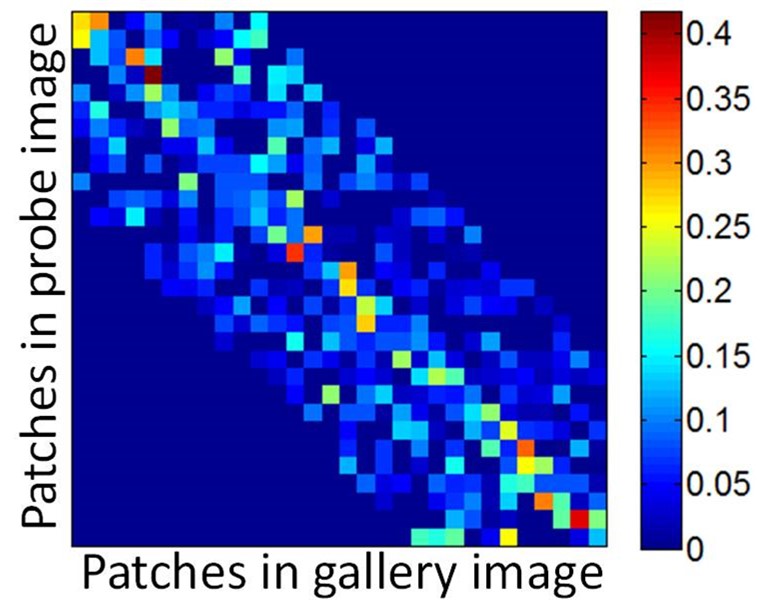}   \label{fig:learned structure i}}
  \hspace{0.3mm}
  \subfloat[]{\includegraphics[width=2.62cm,height=2.5cm]{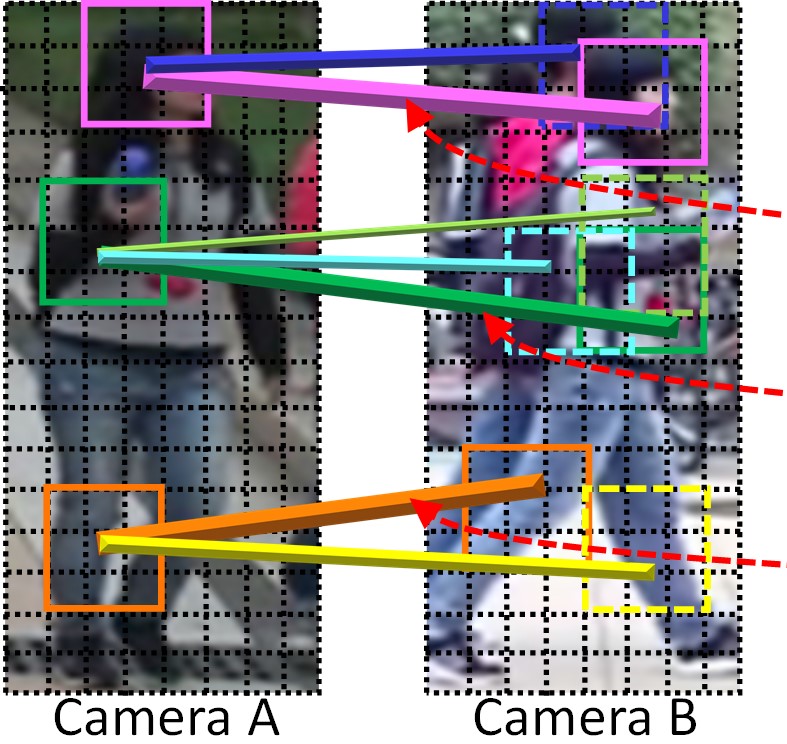}     \label{fig:learned structure j}}
  \hspace{-2.61mm}
  \subfloat[]{\includegraphics[width=2.62cm,height=2.5cm]{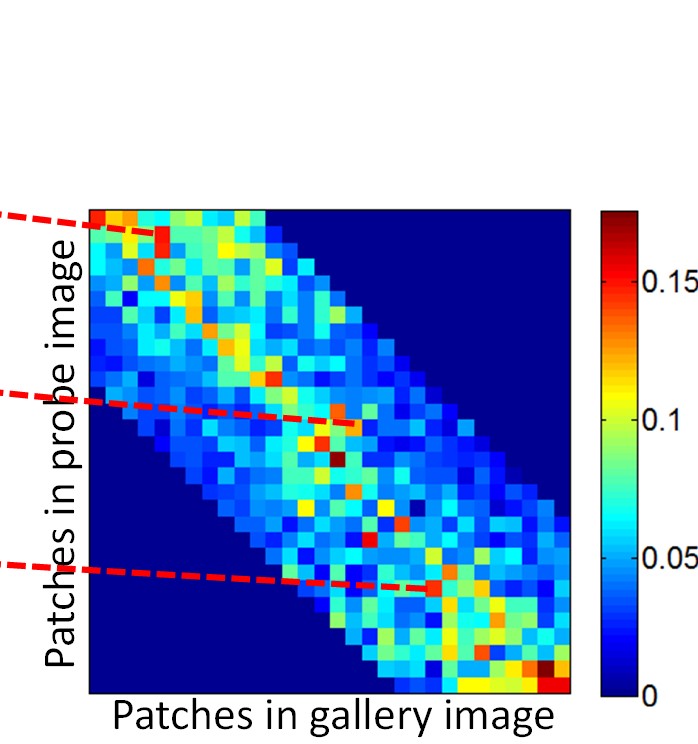}     \label{fig:learned structure k}}
  \hspace{0.3mm}
  \subfloat[]{\includegraphics[width=2.6cm,height=1.9cm]{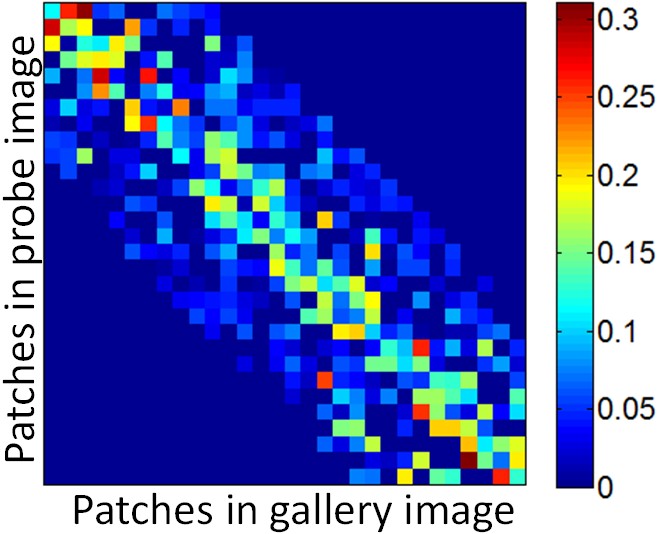}   \label{fig:learned structure l}}
  \caption{The learned correspondence structures for various datasets. (a, d, g, j): The correspondence structures learned by our approach (with the KISSME metric) for the VIPeR~\cite{viper}, PRID 450S~\cite{prid}, 3DPeS~\cite{3dpes}, and  SYSU-sReID~\cite{guo2014multi} datasets, respectively (the correspondence structure for our ROAD dataset is shown in Figure~\ref{fig:mapping structure_example}). The line widths are proportional to the patch-wise probability values. (b, e, h, k): The complete correspondence structure matrices of (a, d, g, j) learned by our approach. (c, f, i, l): The correspondence structure matrices of (a, d, g, j)'s dataset obtained by the simple-average method. (Patches in (b, e, h, k) and (c, f, i, l) are organized by a row-first scanning order. All the correspondence structure matrices are down-sampled for a clearer illustration of the correspondence pattern). (Best viewed in color)}   \label{fig:learned structure}
\end{figure*}

\textbf{Correspondence Structure Initialization.} In this paper, patch-wise matching probabilities $P_{ij}$ in the correspondence structure are initialized by:
\begin{equation}
   P^0_{ij} \propto \left\{
    \begin{aligned}
      &0,~~~~~~~~~~~~\text{if}~~~d{(y_i,y_j)}\geq T_d\\
      &\frac {1}{d{(y_i,y_j)}+1},~~~~~\text{otherwise}
    \end{aligned}
  \right. \,,   \label{equation:equ5a}
\end{equation}
where $y_i$ is {$x_i$}'s co-located patch in camera $B$, such as the two blue patches in Fig.~\ref{fig:learned structure g}. $d{(y_i,y_j)}$ is the distance between patches $y_i$ and $y_j$ which is used to measure the distance between patches $x_i$ and $y_j$ from two different images. The distance $d{(y_i,y_j)}$ is defined as the number of strides to move from $y_i$ to $y_j$ (i.e., the $\ell_1$ distance). We use $d{(y_i,y_j)}+1$ to avoid dividing by zero. $T_d$ is a threshold which is set to be $32$ in this paper. According to Eq.~\ref{equation:equ5a}, a patch pair whose patches are located close to each other (i.e., small $d{(y_i,y_j)}$) will be initialized with a large $P^0_{ij}$ value. On the contrary, patch pairs with large patch-wise distances will be initialized with small $P^0_{ij}$ values. Moreover, if patches in a patch pair have extremely large distance (i.e., $d{(y_i,y_j)}\geq T_d$), we will simply set their corresponding matching probability to be $0$.

\textbf{Binary mapping structure selection.} During each iteration $k$ in the learning process, we first apply correspondence structure $\mathbf{\Theta}^{k-1}_{A,B}=\{P^{k-1}_{ij}\}$ from the previous iteration to calculate the rank orders of all correct match images $V_{\alpha'}$ in the training set. Then, we randomly select $20$ $V_{\alpha'}$ where half of them are ranked among top $50\%$ (implying better Re-ID results) and another half are ranked among the last $50\%$ (implying worse Re-ID results). Finally, we extract binary mapping structures corresponding to these selected images and use them to update and boost the correspondence structure.

Note that we select binary mapping structures for both high- and low-ranked images in order to include a variety of local patch-wise correspondence patterns. In this way, the final obtained correspondence structure can suitably handle the variations in human-pose or local viewpoints.

\textbf{Calculating the updated matching probability.} With the introduction of the binary mapping structure $\mathbf{M}_\alpha$, we can model the updated matching probability in the correspondence structure by combining the estimated matching probabilities from multiple binary mapping structures:
\begin{equation}
\hat{P}^k_{ij}=\sum_{\mathbf{M}_\alpha\in\mathbf{\Gamma}^k}{\hat{P}(x_i,y_j|\mathbf{M_\alpha})\cdot P(\mathbf{M}_\alpha)} \,,
\label{equation:equ5}
\end{equation}
where $\hat{P}^k_{ij}$ is the updated matching probability between patches $x_i$ and $y_j$ in the $k$-th iteration. $\hat{P}(x_i,y_j|\mathbf{M_\alpha})$ is the estimated matching probability between $x_i$ and $y_j$ when including the local correspondence pattern information of binary mapping structure $\mathbf{M}_\alpha$. $\mathbf{\Gamma}^k$ is the set of binary mapping structures selected in the $k$-th iteration. $P(\mathbf{M}_\alpha)=\frac{\tilde{\mathcal{R}}_n(\mathbf{M}_\alpha)}{\sum_{\mathbf{M}_\gamma\in\mathbf{\Gamma}^k}\tilde{\mathcal{R}}_n(\mathbf{M}_\gamma)}$ is the prior probability for binary mapping structure $\mathbf{M}_\alpha$, where $\tilde{\mathcal{R}}_n(\mathbf{M}_\alpha)$ is the CMC score at rank $n$~\cite{cmc} when using $\mathbf{M}_\alpha$ as the correspondence structure to perform person Re-ID over the training images. $n$ is set to be $5$ in our experiments. Similar to $C(x_i,y_j)$, when calculating matching probabilities, we only consider patch pairs whose distances are within a range (cf. Eq.~\ref{equation:equ5a}), while probabilities for other patch pairs are simply set to $0$.

According to Eq.~\ref{equation:equ5}, the updated matching probability $\hat{P}^k_{ij}$ is calculated by integrating the estimated matching probability under different binary mapping structures (i.e., $\hat{P}(x_i,y_j|\mathbf{M_\alpha})$). Moreover, binary mapping structures that have better Re-ID performances (i.e., larger $P(\mathbf{M}_\alpha)$) will have more impacts on the updated matching probability result.

$\hat{P}(x_i,y_j|\mathbf{M_\alpha})$ in Eq.~\ref{equation:equ5} is further calculated by combining a correspondence strength term and a patch importance term:
\begin{equation}
\hat{P}(x_i,y_j|\mathbf{M}_\alpha)=\hat{P}(y_j|x_i,\mathbf{M}_\alpha)\cdot \hat{P}(x_i|\mathbf{M}_\alpha) \,, \label{equation:equ6}
\end{equation}
in which $\hat{P}(y_j|x_i,\mathbf{M}_\alpha)$ is the correspondence strength term reflecting the correspondence strength from $x_i$ to $y_j$ when including $\mathbf{M}_\alpha$. $\hat{P}(x_i|\mathbf{M}_\alpha)$ is the patch importance term reflecting the impact of $\mathbf{M}_\alpha$ to patch $x_i$. $\hat{P}(y_j|x_i,\mathbf{M}_\alpha)$ is calculated as
\begin{equation}
 \hat{P}(y_j|x_i,\mathbf{M}_\alpha) \propto \left\{
   \begin{aligned}
    &1,~~~~~\text{if}~~m_{ij}\in\mathbf{M}_\alpha \\
    &\tilde{\mathcal{A}}_{y_j|x_i,\mathbf{M}_\alpha},~~~\text{otherwise}
  \end{aligned}
 \right. \,,   \label{equation:equ7}
\end{equation}
in which $m_{ij}$ is a patch-wise link connecting $x_i$ and $y_j$. $\tilde{\mathcal{A}}_{y_j|x_i,\mathbf{M}_\alpha}=\frac{\overline{\Phi}_{ij}(x_i,y_j)}{\sum_{y_t,m_{it}\in\mathbf{M}_\alpha}{\overline{\Phi}_{it}(x_i,y_t)}}$, where $\overline{\Phi}_{ij}(x_i,y_j)$ is the average appearance similarity~\cite{6,1} between patches $x_i$ and $y_j$ over all correct match image pairs in the training set. $y_t$ is a patch that is connected to $x_i$ in the binary mapping structure $\mathbf{M}_\alpha$.

According to Eq.~\ref{equation:equ6}, given a specific binary mapping structure $\mathbf{M_\alpha}$, the estimated matching probability $\hat{P}(x_i,y_j|\mathbf{M_\alpha})$ between a probe patch $x_i$ and a gallery patch $y_j$ is calculated by jointly considering the importance of the probe patch $x_i$ under $\mathbf{M_\alpha}$ (i.e., $\hat{P}(x_i|\mathbf{M}_\alpha)$) and the correspondence strength from the probe patch $x_i$ to the gallery patch $y_j$ under $\mathbf{M_\alpha}$ (i.e., $\hat{P}(y_j|x_i,\mathbf{M}_\alpha)$).

Moreover, from Eq.~\ref{equation:equ7}, $\hat{P}(y_j|x_i,\mathbf{M}_\alpha)$ is mainly decided by the relative appearance similarity strength between patch pair $\{x_i, y_j\}$ and all patch pairs which are connected to $x_i$ in the binary mapping structure $\mathbf{M}_\alpha$. This enables patch pairs with larger appearance similarities to have stronger correspondences. Besides, we also include a constraint that $\hat{P}(y_j|x_i,\mathbf{M}_\alpha)$'s value will be the largest (equal to $1$) if the binary mapping structure $\mathbf{M_\alpha}$ includes a link between $x_i$ and $y_j$. In this way, we are able to bring $\mathbf{M_\alpha}$'s impact into full play when estimating the matching probabilities under it.

\begin{algorithm}[t]
   \small
   \caption{Boosting-based Learning Process}
   \small{
     {\bf Input}: A set of training probe images $\{U_\alpha\}$ from camera $A$ and their corresponding cross-view images $\{V_\beta\}$ from camera $B$\\
     {\bf Output}: $\mathbf{\Theta}_{A,B}=\{P_{ij}\}$, the correspondence structure between $\{U_\alpha\}$ and $\{V_\beta\}$
   }
     \begin{algorithmic}[1]
     \small{
      \STATE Find an optimal binary mapping structure $\mathbf{M}_\alpha$ for each probe image $U_{\alpha}$, as described in Sec~\ref{section:boost-learning}
      \STATE Set $k=1$. Initialize $P^0_{ij}$ by Eq.~\ref{equation:equ5a}.
      \STATE Use the previous correspondence structure $\{P^{k-1}_{ij}\}$ to perform Re-ID on $\{U_\alpha\}$ and $\{V_\beta\}$, and select $20$ binary mapping structures $\mathbf{M}_\alpha$ based on the Re-ID result, as described in Sec~\ref{section:boost-learning}
      \STATE Compute updated matching probability $\hat{P}^k_{ij}$ by Eq.~\ref{equation:equ5}
      \STATE Compute new candidate correspondence structure $\{\tilde{P}^k_{ij}\}$ by Eq.~\ref{equation:equ11}
      \STATE Use \{$\tilde{P}^k_{ij}$\} to calculate the Re-ID result on the training set by Eq.~\ref{equation:equ4}, and compare it with the result of the previous structure $\{P^{k-1}_{ij}\}$
      \STATE If \{$\tilde{P}^k_{ij}$\} has better Re-ID result than $\{P^{k-1}_{ij}\}$, set $\{{P}^k_{ij}\}=\{\tilde{P}^k_{ij}\}$; Otherwise, $\{{P}^k_{ij}\}=\{P^{k-1}_{ij}\}$
      \STATE Set $k=k+1$ and go back to step $3$ if not converged or not reaching the maximum iteration number
      \STATE Output $\{P_{ij}\}$
     }
     \end{algorithmic}
     \label{algorithm:progressive updating}
\end{algorithm}

Furthermore, the probe patch importance probability $\hat{P}(x_i|\mathbf{M}_\alpha)$ in Eq.~\ref{equation:equ6} can be further calculated by accumulating the impact of each individual link in $\mathbf{M}_\alpha$ on patch $x_i$:
\begin{align}
 \label{equation:equ8}
 \hat{P}(x_i|\mathbf{M}_\alpha)=\sum_{m_{st}\in\mathbf{M}_\alpha}{}&{\hat{P}(x_i|m_{st},\mathbf{M}_\alpha)}\notag\\
 &\cdot\hat{P}(m_{st}|\mathbf{M}_\alpha) \,,
\end{align}
where $m_{st}$ is a patch-wise link in $\mathbf{M}_\alpha$, as the red lines in Fig.~\ref{fig:one to many matching a}. $\hat{P}(m_{st}|\mathbf{M}_\alpha)$ is the importance probability of link $m_{st}$ which is defined similar to $P(\mathbf{M}_\alpha)$:
\begin{equation}
 \hat{P}(m_{st}|\mathbf{M}_\alpha)=\frac{\tilde{\mathcal{R}}_n (m_{st})}{\sum_{m_{hg}\in\mathbf{M}_\alpha}\tilde{\mathcal{R}}_n (m_{hg})} \,,
\label{equation:equ9}
\end{equation}
where $\tilde{\mathcal{R}}_n (m_{st} )$ is the rank-$n$ CMC score \cite{cmc} when only using a single link $m_{st}$ as the correspondence structure to perform Re-ID.
$\hat{P}(x_i|m_{st}, \mathbf{M}_\alpha)$ in Eq.~\ref{equation:equ8} is the impact probability from link $m_{st}$ to patch $x_i$, defined as:
\begin{equation}
 \hat{P}(x_i|m_{st}, \mathbf{M}_\alpha) \propto \left\{
   \begin{aligned}
     &0,~~~~~~~~~~\text{if } d{(x_i,x_s)}\geq T_d\\
     &\frac {1}{d{(x_i,x_s)}+1},~~\text{otherwise}
   \end{aligned}
 \right.  \label{equation:equ10}
\end{equation}
where $x_s$ is link $m_{st}$'s end patch in camera A. $d(\cdot)$ and $T_d$ are the same as in Eq.~\ref{equation:equ5a}.

According to Eq.~\ref{equation:equ8}, a probe patch $x_i$'s importance probability $\hat{P}(x_i|\mathbf{M}_\alpha)$ under $\mathbf{M}_\alpha$ is calculated by integrating the impact probability $\hat{P}(x_i|m_{st}, \mathbf{M}_\alpha)$ from each individual link in $\mathbf{M}_\alpha$ to $x_i$. Since $\hat{P}(x_i|m_{st}, \mathbf{M}_\alpha)$ is modeled by the distance between link $m_{st}$ and $x_i$ (cf. Eq.~\ref{equation:equ10}), it enables patches that are closer to the links in $\mathbf{M}_\alpha$ to have larger importance probabilities. Moreover, Eq.~\ref{equation:equ9} further guarantees links with better Re-ID performances (i.e., larger $\hat{P}(m_{st}|\mathbf{M}_\alpha)$) to have more impacts on the calculated importance probability $\hat{P}(x_i|\mathbf{M}_\alpha)$.

\textbf{Correspondence structure update.} After obtaining the updated matching probability $\hat{P}^k_{ij}$ in Eq.~\ref{equation:equ5}, we can calculate matching probabilities of the new candidate correspondence structure \{$\tilde{P}^k_{ij}$\} by:

\begin{equation}
\tilde{P}^k_{ij} = (1-\varepsilon)P^{k-1}_{ij}+\varepsilon\hat{P}^k_{ij} \,,
\label{equation:equ11}
\end{equation}
where $P^{k-1}_{ij}$ is the matching probability in iteration $k-1$. $\varepsilon$ is the update rate which is set $0.2$ in our paper.

In order to guarantee that the boosting-based learning approach keeps optimizing Eq.~\ref{equation:equ4} during the iteration process, an evaluation module is further included to decide whether to accept the new candidate correspondence structure \{$\tilde{P}^k_{ij}$\}. Specifically, we use \{$\tilde{P}^k_{ij}$\} to calculate the Re-ID result on the training set by Eq.~\ref{equation:equ4}, and compare it with the Re-ID result of the previous structure $\{P^{k-1}_{ij}\}$. If \{$\tilde{P}^k_{ij}$\} has better Re-ID result than $\{P^{k-1}_{ij}\}$, we will select $\{\tilde{P}^k_{ij}\}$ as the correspondence structure in the $k$-th iteration (i.e., $\{{P}^k_{ij}\}=\{\tilde{P}^k_{ij}\}$); Otherwise, we will still keep the correspondence structure in the previous iteration (i.e., $\{{P}^k_{ij}\}=\{P^{k-1}_{ij}\}$).


From Equations~\ref{equation:equ5}--\ref{equation:equ11}, our update process integrates multiple variables (i.e., binary mapping structure, individual links, patch-link correlation) into a unified probability framework. In this way, various information cues that affect the objective function in Equation~\ref{equation:equ4} (such as appearances, ranking results, and patch-wise correspondence patterns) can be effectively included, such that suitable candidate correspondence structures for the objective function can be created during the iterative model updating process. Therefore, by selecting the best structure among these candidate structures (with the evaluation module), a satisfactory correspondence structure can be obtained.

Figures~\ref{fig:mapping structure_example},~\ref{fig:one to many matching}, and~\ref{fig:learned structure} show some examples of the correspondence structures learned from different cross-view datasets. From these figures, we can see that the correspondence structures learned by our approach can suitably indicate the matching correspondence between spatial misaligned patches. For example, in Figures~\ref{fig:learned structure g}- \ref{fig:learned structure h}, the large lower-to-upper misalignment between cameras is effectively captured. Besides, the matching probability values in the correspondence structure also indicate the correlation strengths between patch pairs, which are displayed as the colored blocks in Figures~\ref{fig:learned structure b}, \ref{fig:learned structure e}, \ref{fig:learned structure h} and \ref{fig:learned structure k}.

Furthermore, comparing Figures~\ref{fig:mapping structure_example_a} and~\ref{fig:mapping structure_example_b}, we can see that human-pose variation is also suitably handled by the learned correspondence structure. More specifically, although images in Fig.~\ref{fig:mapping structure_example} have different human poses, patches of camera $A$ in both figures can correctly find their corresponding patches in camera $B$ since the one-to-many matching probability graphs in the correspondence structure suitably embed the local correspondence variation between cameras. Similar observations can also be obtained from Figures~\ref{fig:one to many matching b} and~\ref{fig:one to many matching c}. 

\section{Extending The Approach with Multiple Correspondence Structures} \label{section:multiple correspondence structure}

As mentioned before, since people often show different poses inside a camera view, the spatial correspondence pattern between a camera pair can be further divided and modeled by a set of sub-patterns according to these pose variations. Therefore, we extend our approach by introducing a multi-structure scheme, which first learns a set of local correspondence structures to capture various spatial correspondence sub-patterns between a camera pair, and then adaptively selects suitable correspondence structures to handle the spatial misalignments between individual images.

\begin{figure}[t]
  \centering
  \includegraphics[width=0.48\textwidth]{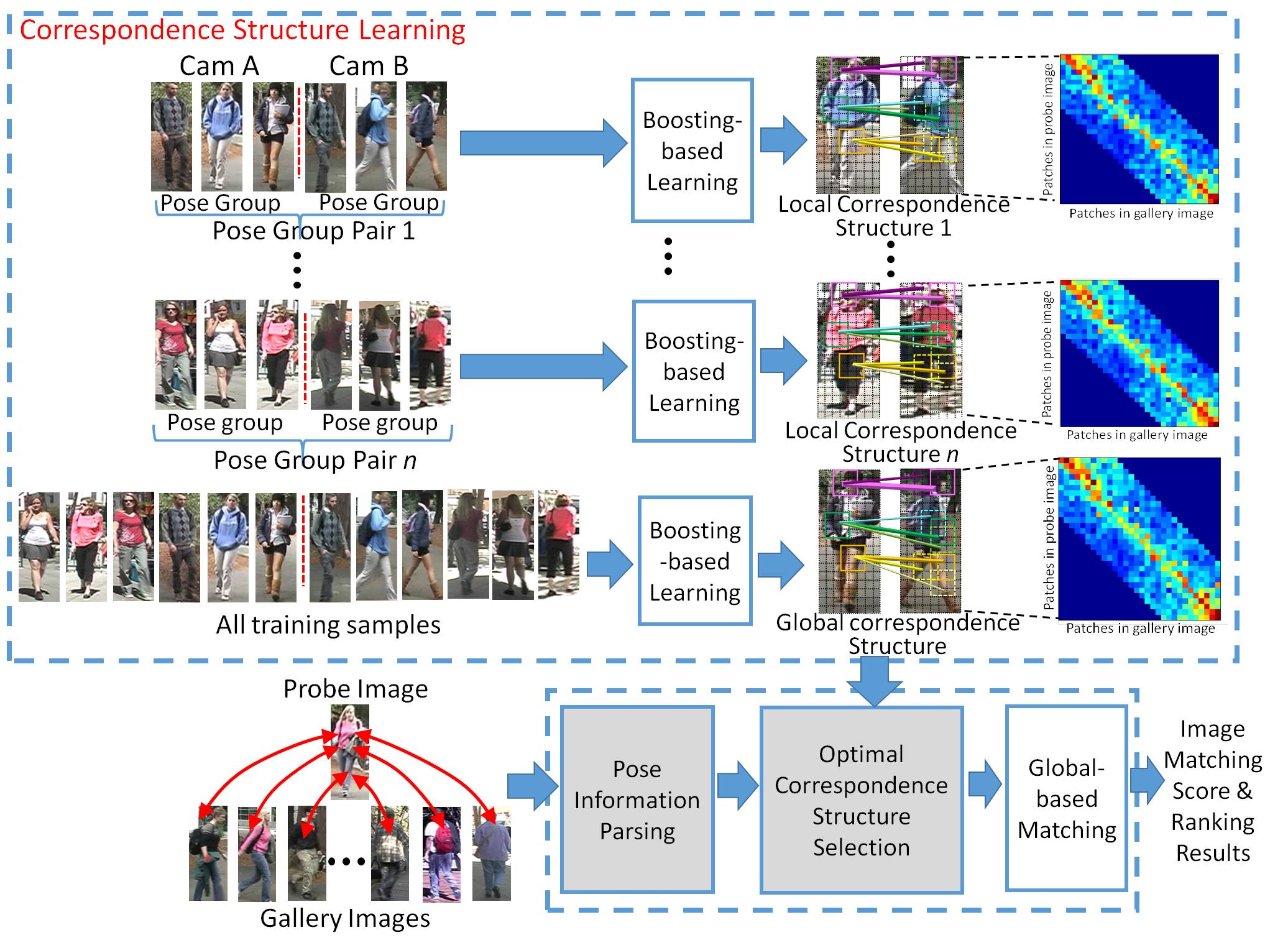}
  \caption{Framework of the proposed approach with multiple correspondence structures.}    \label{fig:framework of multiple structures}
\end{figure}

Fig.~\ref{fig:framework of multiple structures} shows our extended person Re-ID framework with multiple correspondence structures. During the training stage, we first divide images in each camera into different pose groups, where images in each pose group includes people with similar poses. Then, we construct a set of pose-group pairs. Each pose-group pair is composed of two pose groups from different cameras and reflects one spatial correspondence sub-pattern between cameras. Finally, we apply our boosting-based learning process over the constructed pose-group pairs and obtain a set of local correspondence structures to capture the correspondence sub-patterns between a camera pair.

During the prediction stage, we parse the human pose information for each pair of images being matched, and adaptively select the optimal correspondence structure to calculate the image-wise matching score for person Re-ID. More specifically, for a pair of test images (a probe image and a gallery image), we first parse the human pose information for both images~\cite{xiaogang2013iccv} and classify them into their closest pose groups, respectively. Then, we are able to identify the pose-group pair for the test images and use its corresponding correspondence structure as the optimal structure for image matching.

Moreover, the following settings need to be mentioned about our multi-structure scheme in Fig.~\ref{fig:framework of multiple structures}.

\begin{enumerate}
 \item During the prediction stage, since we only need to obtain a rough estimation of human poses (e.g., differentiate front pose, side pose, or back pose), we utilize a simple but effective method which parses human pose from his/her head orientations. Specifically, we first utilize semantic region parsing~\cite{xiaogang2013iccv} to obtain head and hair regions from a test image. Then, we construct a feature vector including the size and relative location information of head and hair regions, and apply a linear SVM~\cite{SVM} to classify this feature vector into one of the pose groups. According to our experiments, this method can achieve satisfactory (over $90\%$) pose classification accuracy on the datasets in our experiments. Note that our framework is general, and in practice, more sophisticated pose estimation methods~\cite{wohlhart2015cvpr, dantone2013cvpr, added[3]} can be included to handle pose information parsing under more complicated scenarios.
\item During the training stage, we utilize two ways to divide pose groups \& pose group pairs: (1) Defining and dividing pose groups \& pose group pairs manually; (2) Dividing pose groups \& pose group pairs automatically, where we first cluster training images~\cite{lu2014csvt} in each camera into a set of pose groups (using the same feature vector as described in point $1$), and then form camera-wise pose group pairs based on the pose group set including the largest number of pose groups. Note that in order to automatically decide the number of clusters, we utilize a spectral clustering scheme~\cite{lu2014csvt} which is able to find the optimal cluster number by integrating a clustering-coherency metric to evaluate the results under different cluster numbers. Since the automatic pose group pair division scheme can properly catch the major pose correspondence patterns between a camera pair, it can also achieve satisfactory performances. Experimental results show that methods with this automatic pose group pair division scheme can achieve similar Re-ID results to the ones with manual pose group pair division (cf. Section~\ref{section:multiple correspondence structure2}).
 \item Note that in Fig.~\ref{fig:framework of multiple structures}, besides the \emph{local} correspondence structures learned from pose-group pairs, we also construct a \emph{global} correspondence structure learned from all training images (cf. Fig.~\ref{fig:framework}). Thus, if a test image pair cannot find suitable local correspondence structure (e.g., a test image cannot be confidently classified into any pose group, or the correspondence structure of a specific pose-group pair is not constructed due to limited training samples), this global correspondence structure will be applied to calculate the matching score of this test image pair.
\end{enumerate}

\section{Experimental Results} \label{section:experimental evaluation}

\subsection{Datasets and Experimental Settings} \label{dataset setting}

\begin{figure}[t]
  \centering
  \subfloat[]{\includegraphics[width=2.1cm,height=2.0cm]{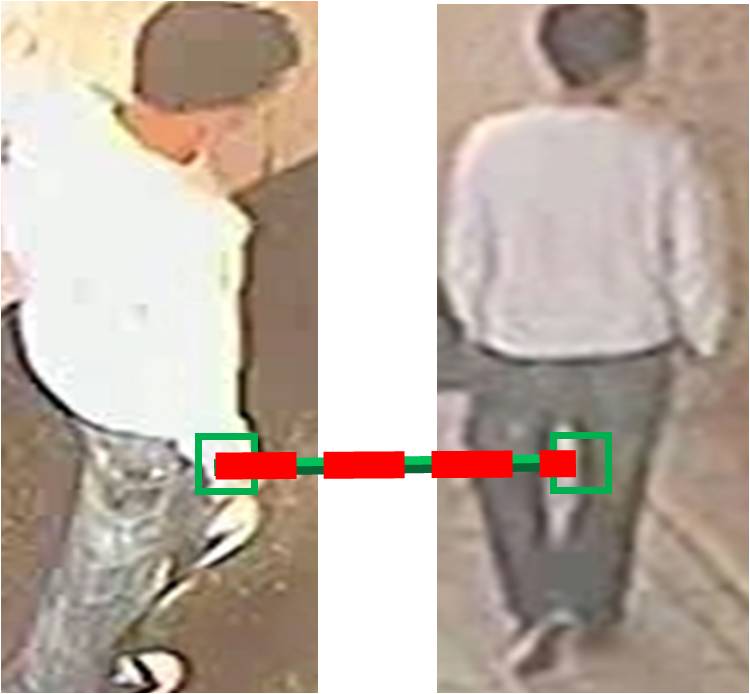}     \label{fig:different mapping structures a}}
    \hspace{2mm}
  \subfloat[]{\includegraphics[width=2.1cm,height=2.0cm]{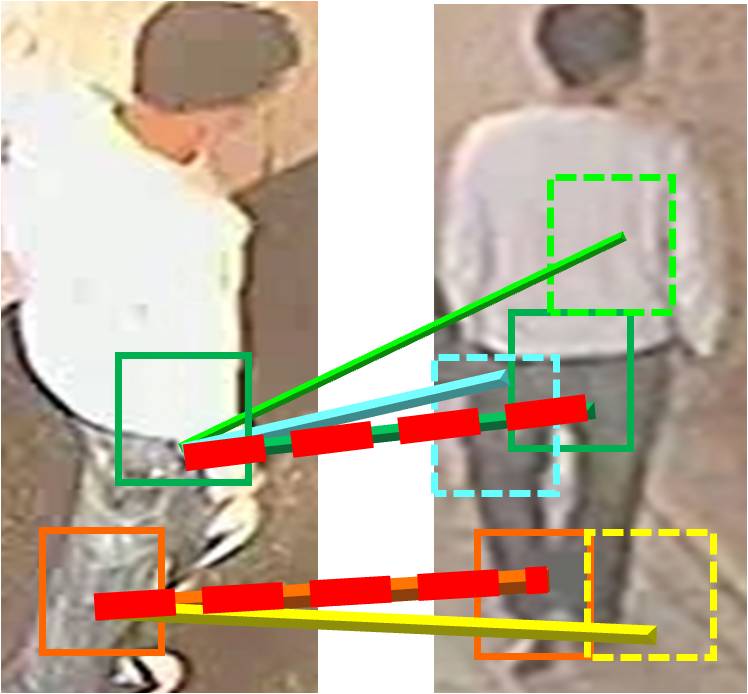}     \label{fig:different mapping structures b}}
    \hspace{2mm}
  \subfloat[]{\includegraphics[width=2.1cm,height=2.0cm]{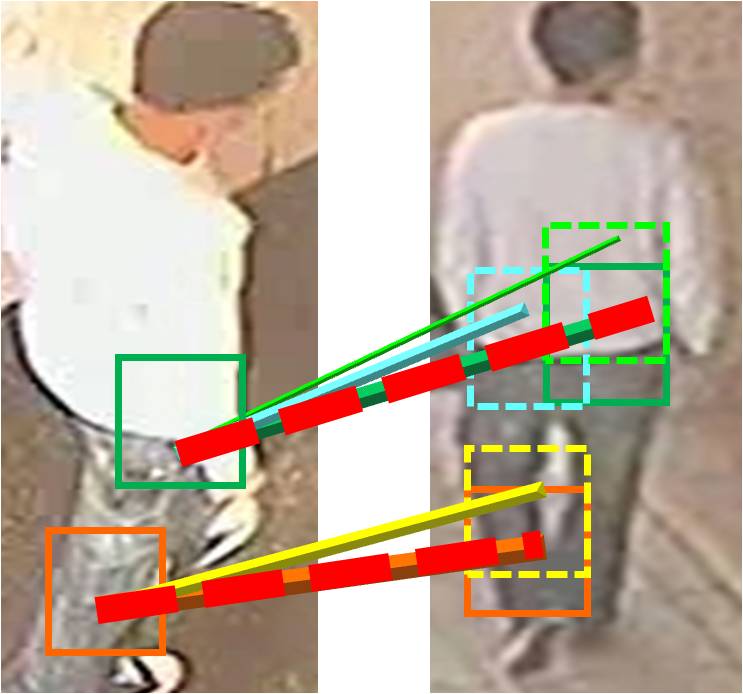}     \label{fig:different mapping structures c}}
  \\
    \subfloat[]{\includegraphics[width=2.1cm,height=2.0cm]{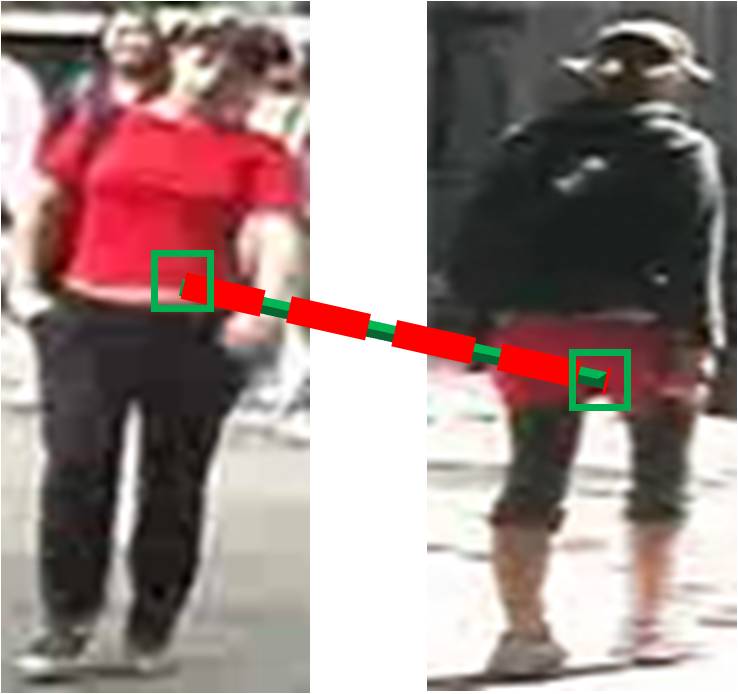}   \label{fig:different mapping structures d}}
    \hspace{2mm}
  \subfloat[]{\includegraphics[width=2.1cm,height=2.0cm]{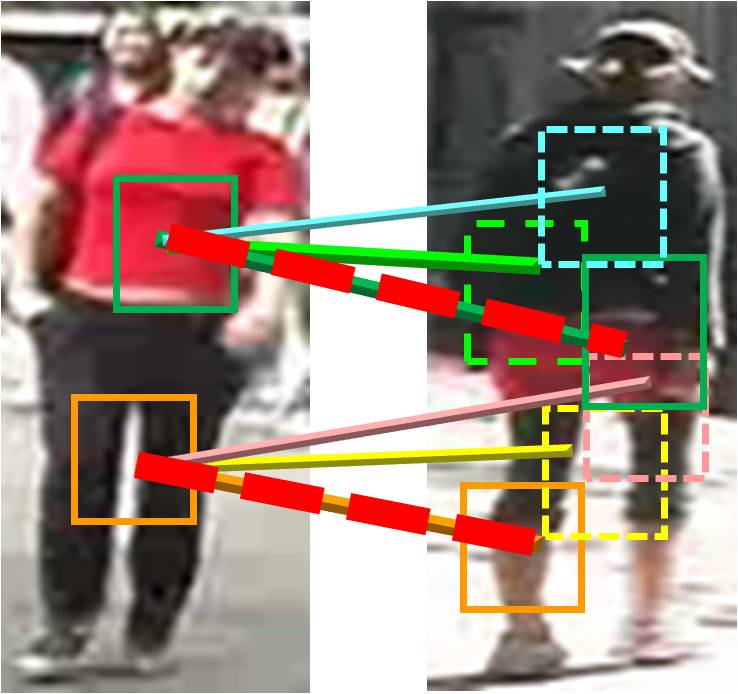}     \label{fig:different mapping structures e}}
    \hspace{2mm}
  \subfloat[]{\includegraphics[width=2.1cm,height=2.0cm]{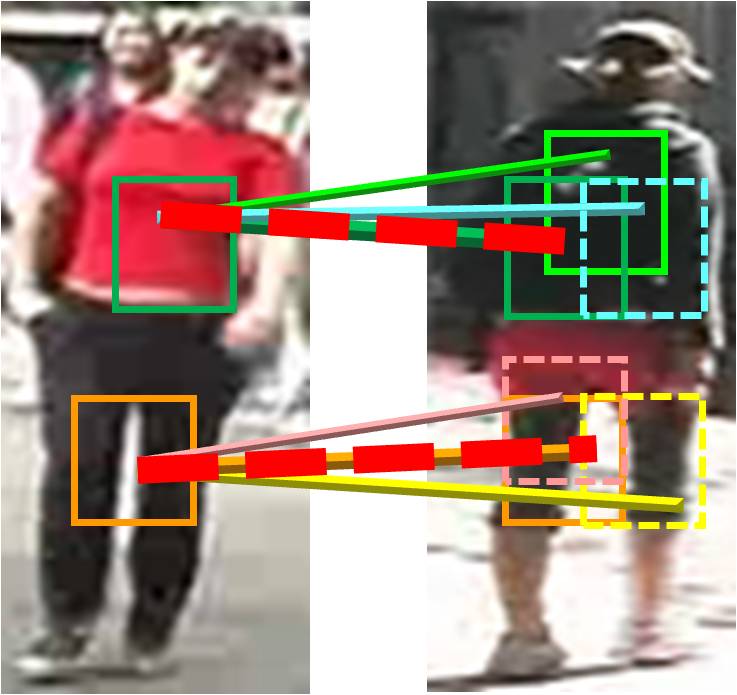}     \label{fig:different mapping structures f}}
  \\
  \subfloat[]{\includegraphics[width=2.1cm,height=2.0cm]{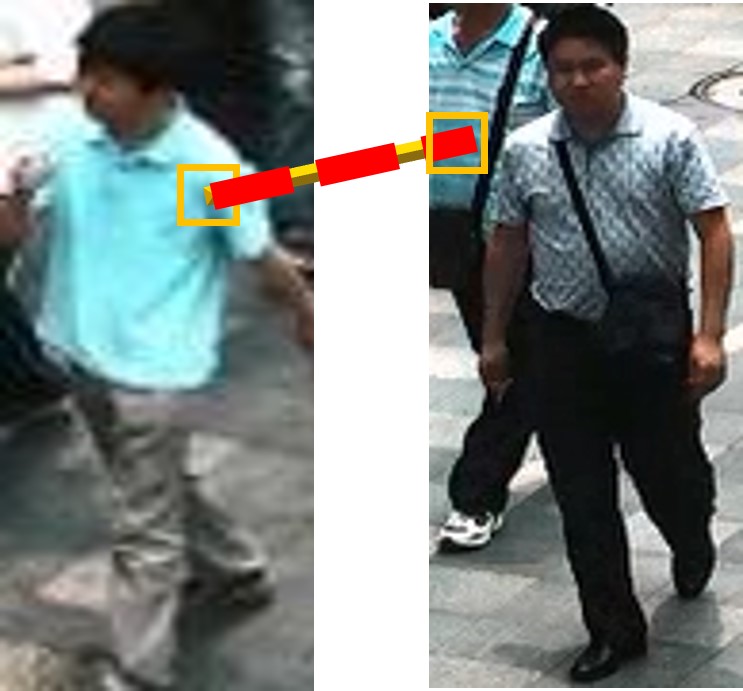}     \label{fig:different mapping structures g}}
    \hspace{2mm}
  \subfloat[]{\includegraphics[width=2.1cm,height=2.0cm]{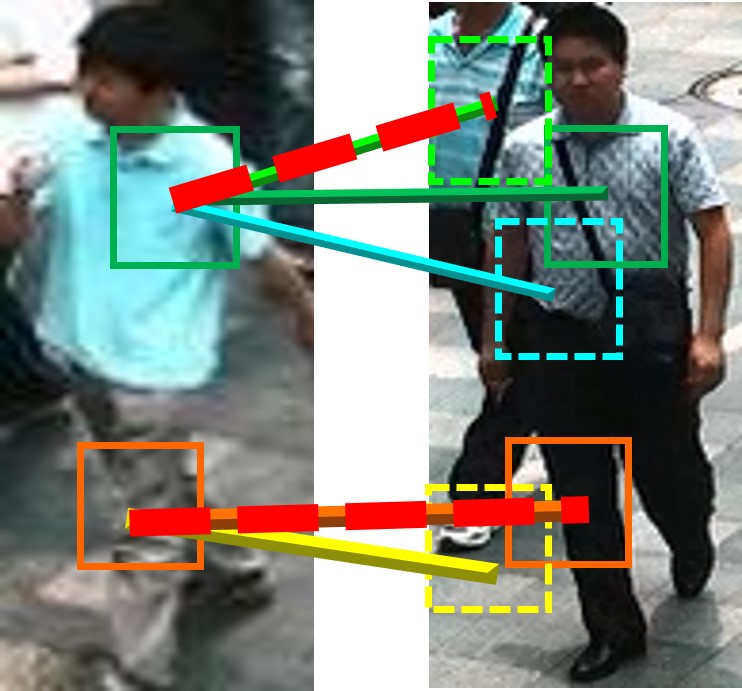}     \label{fig:different mapping structures h}}
    \hspace{2mm}
  \subfloat[]{\includegraphics[width=2.1cm,height=2.0cm]{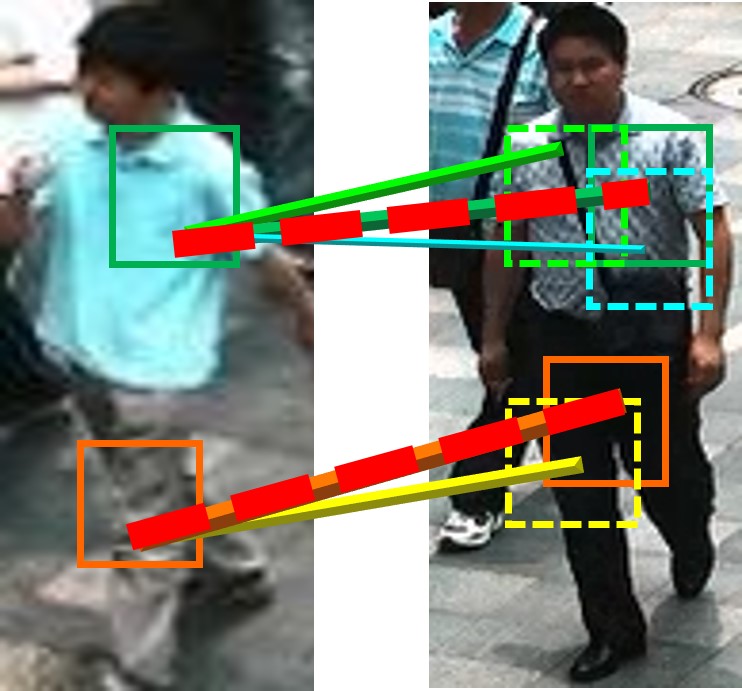}     \label{fig:different mapping structures i}}
  \\
  \subfloat[]{\includegraphics[width=2.1cm,height=2.0cm]{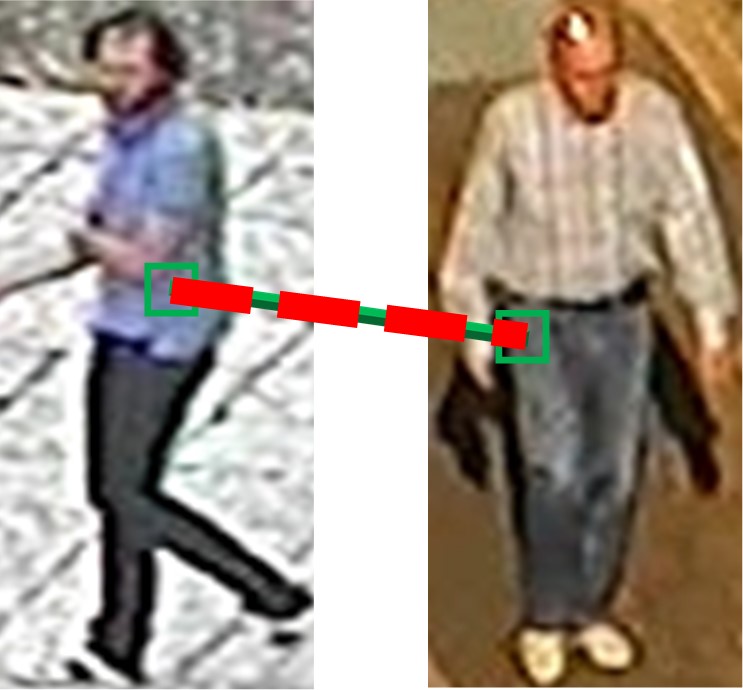}     \label{fig:different mapping structures j}}
    \hspace{2mm}
  \subfloat[]{\includegraphics[width=2.1cm,height=2.0cm]{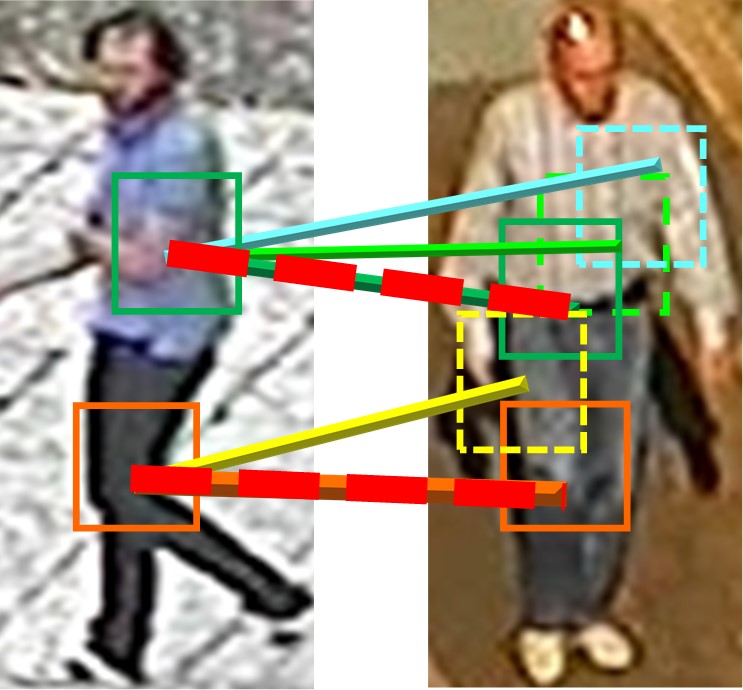}     \label{fig:different mapping structures k}}
    \hspace{2mm}
  \subfloat[]{\includegraphics[width=2.1cm,height=2.0cm]{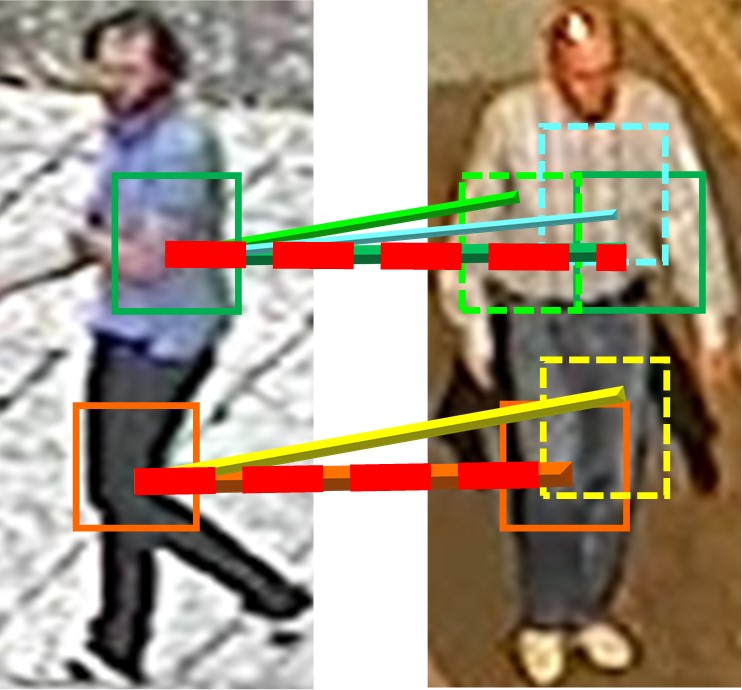}     \label{fig:different mapping structures l}}
  \caption{Comparison of different patch mapping methods. Left column: the adjacency-constrained method; Middle column: the simple-average method; Last column: our approach (with the KISSME metric). The solid lines represent matching probabilities in a correspondence structure and the red-dashed lines represent patch matching results. Note that the image pair in (a)-(c) includes the same person (i.e., correct match) while the image pairs in (d)-(l) include different people (i.e., wrong match). (Best viewed in color)}   \label{fig:different mapping structures}
\end{figure}

We perform experiments on the following five datasets:

\textbf{VIPeR.} The VIPeR dataset~\cite{viper} is a commonly used dataset which contains $632$ image pairs for $632$ pedestrians, as in Figures~\ref{fig:one to many matching a}-\ref{fig:one to many matching c} and~\ref{fig:different mapping structures d}. It is one of the most challenging datasets which includes large differences in viewpoint, pose, and illumination between two camera views. Images from camera $A$ are mainly captured from $0$ to $90$ degree while camera $B$ mainly from $90$ to $180$ degree.

\textbf{PRID 450S.} The PRID 450S dataset~\cite{prid} consists of $450$ person image pairs from two non-overlapping camera views. Some example images in PRID 450S dataset are shown in Fig.~\ref{fig:learned structure d}. It is also challenging due to low image qualities and viewpoint changes.

\textbf{3DPeS.} The 3DPeS dataset~\cite{3dpes} is comprised of $1012$ images from $193$ pedestrians captured by eight cameras, where each person has $2$ to $26$ images, as in Figures~\ref{fig:learned structure g} and~\ref{fig:different mapping structures a}. Note that since there are eight cameras with significantly different views in the dataset, in our experiments, we group cameras with similar views together and form three camera groups. Then, we train a correspondence structure between each pair of camera groups. Finally, three correspondence structures are achieved and utilized to calculate Re-ID performance between different camera groups. For images from the same camera group, we simply utilize adjacency-constrained search~\cite{6} to find patch-wise mapping and calculate the image matching score accordingly.

\textbf{Road.} The Road dataset is our own constructed dataset which includes $416$ image pairs taken by two cameras with camera $A$ monitoring an exit region and camera $B$ monitoring a road region, as in Figures~\ref{fig:mapping structure_example} and~\ref{fig:different mapping structures g}.\footnote{Available at \url{http://min.sjtu.edu.cn/lwydemo/personReID.htm}} This dataset has large variation of human pose and camera angle. Images in this dataset are taken from a realistic crowd road scene.

\textbf{SYSU-sReID.} The SYSU-sReID dataset~\cite{guo2014multi} contains $502$ individual pairs taken by two disjoint cameras in a campus environment, as in Fig.~\ref{fig:learned structure k}. This dataset includes more cross-camera pose correspondence patterns.

For all of the above datasets, we follow previous methods \cite{3,kernel-based,SalientColor} and perform experiments under $50\%$-training and $50\%$-testing. All images are scaled to $128\times 48$. The patch size in our approach is $24\times 18$. The stride size between neighboring patches is $6$ horizontally and $8$ vertically for probe images, and $3$ horizontally and $4$ vertically for gallery images. Note that we use smaller stride size in gallery images in order to obtain more patches. In this way, we can have more flexibilities during patch-wise matching.

\begin{figure}[t]
  \centering
  \subfloat[VIPeR dataset]{\includegraphics[width=3.95cm,height=2.9cm]{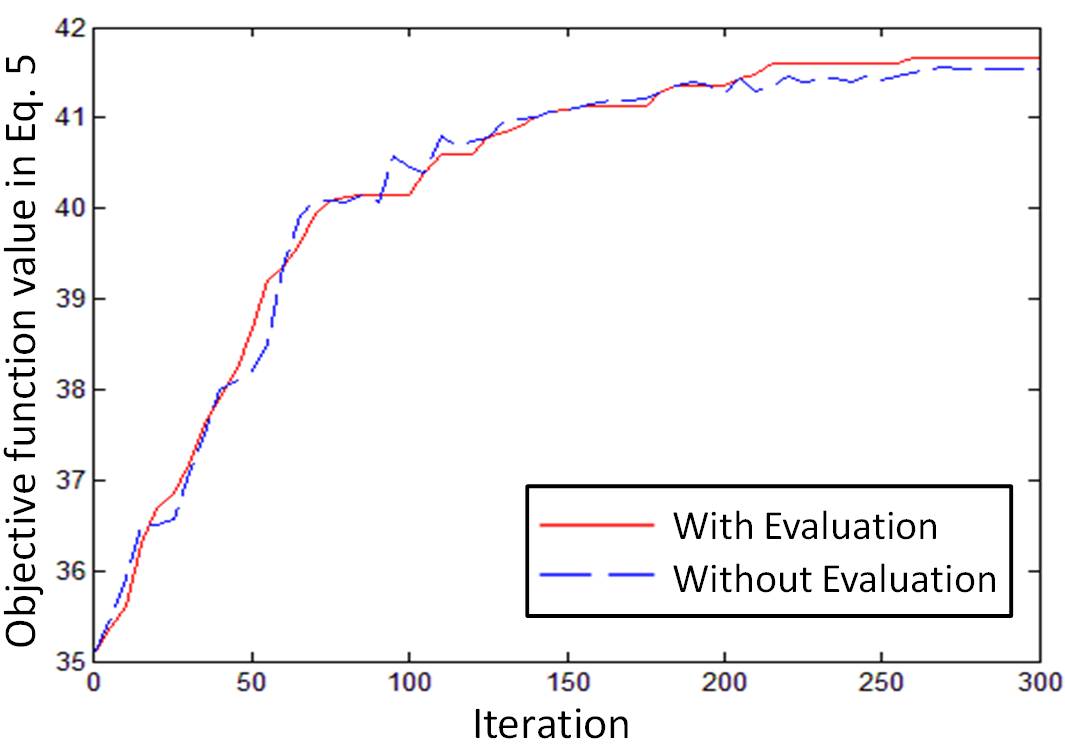}     \label{fig:convergence a}}
    \hspace{1mm}
  \subfloat[SYSU dataset]{\includegraphics[width=3.95cm,height=2.9cm]{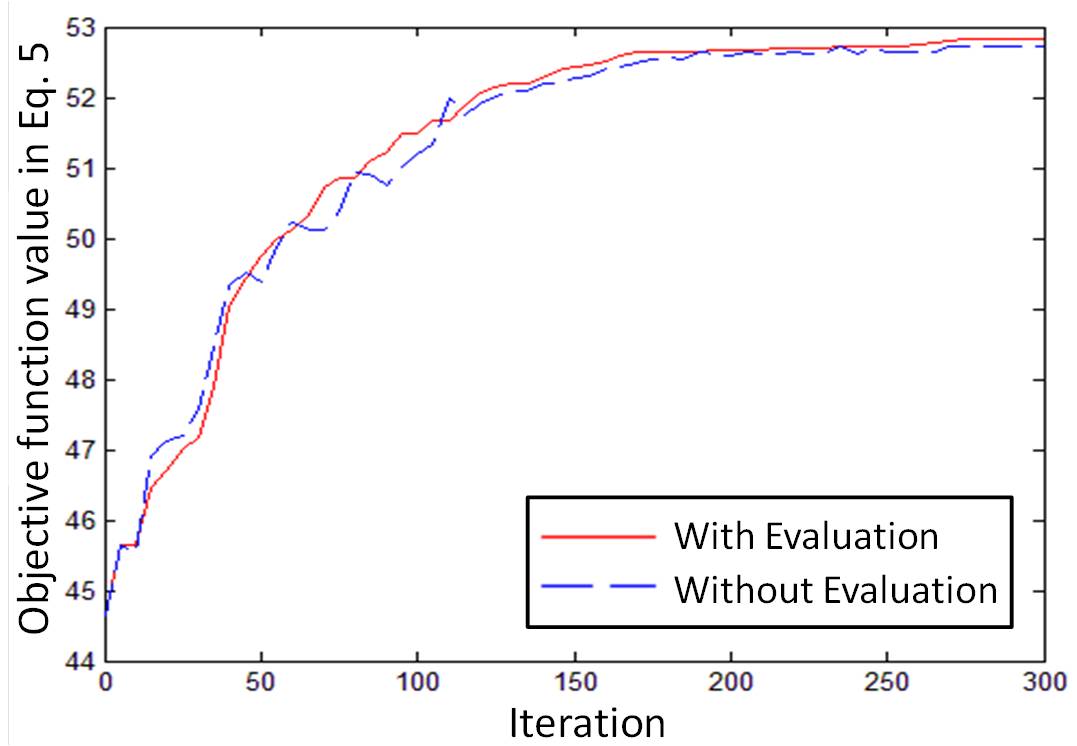}     \label{fig:convergence b}}
  \caption{The convergence curves when using our learning approach to find the correspondence structure in the VIPeR and SYSU datasets. (y-axis: the value of the objective function in Eq.~\ref{equation:equ4} when using the correspondence structure in a certain iteration; x-axis: the iteration number. Red solid curve: our learning approach with the evaluation module; Blue dashed curve: our learning approach without the evaluation module) (Best viewed in color)} \label{fig:convergence}
\end{figure}

Furthermore, in order to save computation complexity, we make the following simplifications on the learning process during our experiments:
\begin{enumerate}
 \item The global constraint in Eq.~\ref{equation:equ3} is not applied in the training stage (cf. step $3$ in Algorithm~\ref{algorithm:progressive updating}). Our experiments show that skipping the global constraint in training does not affect the final Re-ID results too much.
 \item The evaluation module (cf. step $7$ in Algorithm~\ref{algorithm:progressive updating}) is also skipped in the training stage, i.e., we directly use $\{\tilde{P}^k_{ij}\}$ in Eq.~\ref{equation:equ11} to be the correspondence structure in the $k$-th iteration without evaluating whether it is better than $\{{P}^{k-1}_{ij}\}$. Our experiments show that our learning process without the evaluation module can achieve similar Re-ID results to the one including the evaluation module while still obtaining stable correspondence structures within $300$ iterations. For example, Fig.~\ref{fig:convergence} shows the convergence curves when using our learning approach to find the correspondence structure in the VIPeR and SYSU datasets. We can see that our learning approach without the evaluation module (the blue dashed curves) can achieve stable results after $270$ iterations, which is similar to the one including the evaluation module (the red solid curves).
\end{enumerate}

\subsection{Results of using a single correspondence structure}

We first evaluate the performance of our approach by using a single correspondence structure to handle the spatial misalignments between a camera pair (cf. Fig.~\ref{fig:framework}). Note that although multiple correspondence structures are used to handle the view differences among the $8$ cameras in the 3DPeS dataset (cf. Section~\ref{dataset setting}), we only use one correspondence structure for each camera pair. Therefore, it also belongs to the situation of using a single correspondence structure.

\subsubsection{Patch matching performances} We compare the patch matching performance of three methods: (1) The adjacency-constrained search method~\cite{6,8,part2} which finds a best matched patch for each patch in a probe image (probe patch) by searching a fixed neighborhood region around the probe patch's co-located patch in a gallery image (\emph{Adjacency-constrained}). (2) The simple-average method which simply averages the binary mapping structures for different probe images (as in Fig.~\ref{fig:one to many matching a}) to be the correspondence structure and combines it with a global constraint to find the best one-to-one patch matching result (\emph{Simple-average}). (3) Our approach which employs a boosting-based process to learn the correspondence structure and combines it with a global constraint to find the best one-to-one patch matching result.

Fig.~\ref{fig:different mapping structures} shows the patch mapping results of different methods, where solid lines represent matching probabilities in a correspondence structure and red-dashed lines represent patch matching results. Besides, Figures~\ref{fig:learned structure b}, \ref{fig:learned structure e}, \ref{fig:learned structure h}, \ref{fig:learned structure k} and Figures~\ref{fig:learned structure c}, \ref{fig:learned structure f}, \ref{fig:learned structure i}, \ref{fig:learned structure l} also show the correspondence structure matrices obtained by our approach and the simple-average method, respectively. From Figures~\ref{fig:learned structure} and~\ref{fig:different mapping structures}, we can observe:

\begin{enumerate}
 \item Since the adjacency-constrained method searches a fixed neighborhood region without considering the correspondence pattern between cameras, it may easily be interfered by wrong patches with similar appearances in the neighborhood (cf. Figures~\ref{fig:different mapping structures d}, \ref{fig:different mapping structures g}, \ref{fig:different mapping structures j}). Comparatively, with the indicative matching probability information in the correspondence structure, the interference from mismatched patches can be effectively reduced (cf. Figures~\ref{fig:different mapping structures f}, \ref{fig:different mapping structures i}, \ref{fig:different mapping structures l}).

 \item When there are large misalignments between cameras, the adjacency-constrained method may fail to find proper patches as the correct patches may be located outside the neighborhood region, as in Fig.~\ref{fig:different mapping structures a}. Comparatively, the large misalignment pattern between cameras can be properly captured by our correspondence structure, resulting in a more accurate patch matching result (cf. Fig.~\ref{fig:different mapping structures c}).

 \item Comparing Figures~\ref{fig:learned structure b}, \ref{fig:learned structure e}, \ref{fig:learned structure h}, \ref{fig:learned structure k} and Figures~\ref{fig:learned structure c}, \ref{fig:learned structure f}, \ref{fig:learned structure i}, \ref{fig:learned structure l} with the last two columns in Fig.~\ref{fig:different mapping structures}, it is obvious that the correspondence structures by our approach is better than the simple average method. Specifically, the correspondence structures by the simple average method include many unsuitable matching probabilities which may easily result in wrong patch matches. In contrast, the correspondence structures by our approach are more coherent with the actual spatial correspondence pattern between cameras. This implies that reliable correspondence structure cannot be easily achieved without suitably integrating the information cues between cameras.
\end{enumerate}

\begin{figure*}[t]
  \centering
  \vspace{-2mm}
  \subfloat[the KISSME metric]{\includegraphics[width=5.0cm,height=2.6cm]{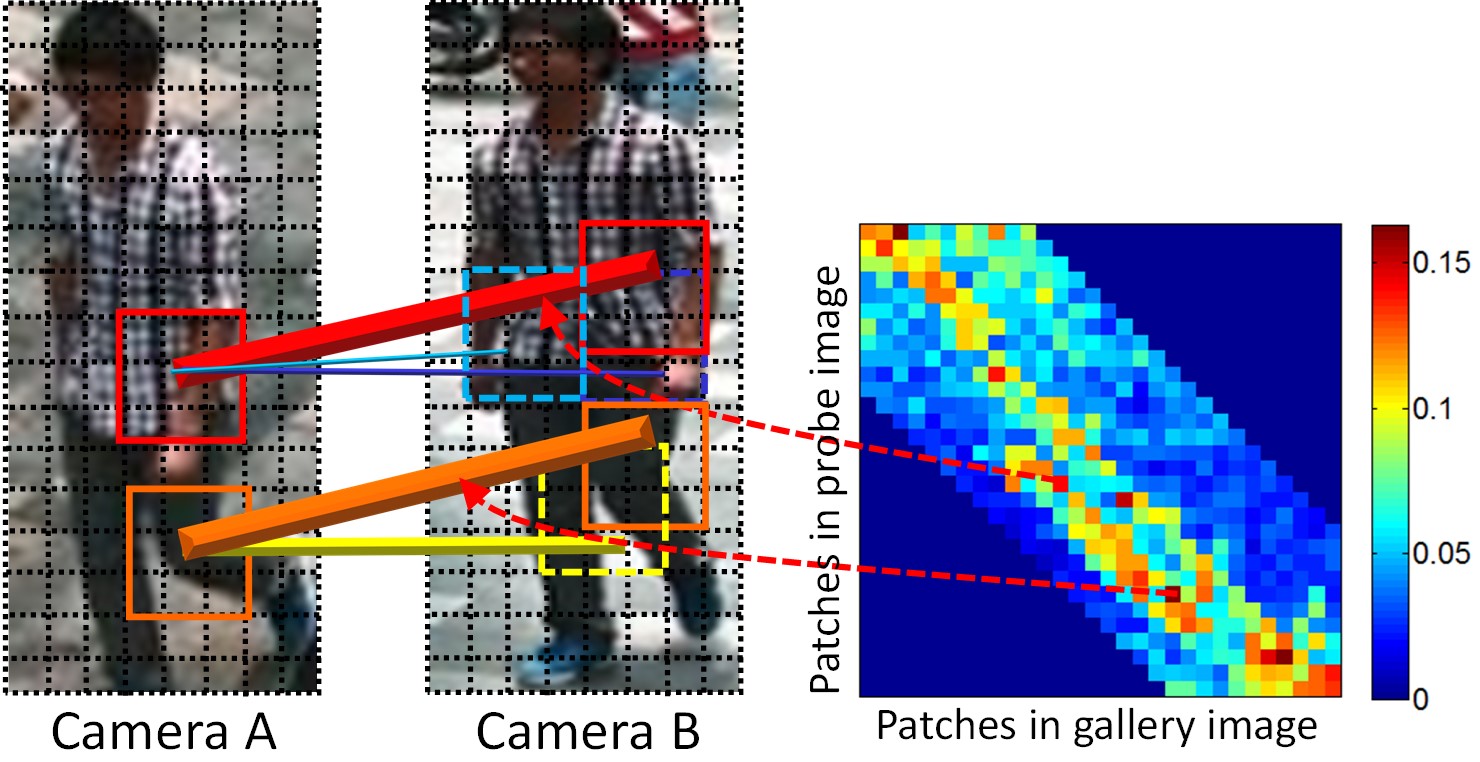}  \label{fig:structures with various dms a}}
  \hspace{4mm}
  \subfloat[the kLFDA metric]{\includegraphics[width=5.0cm,height=2.6cm]{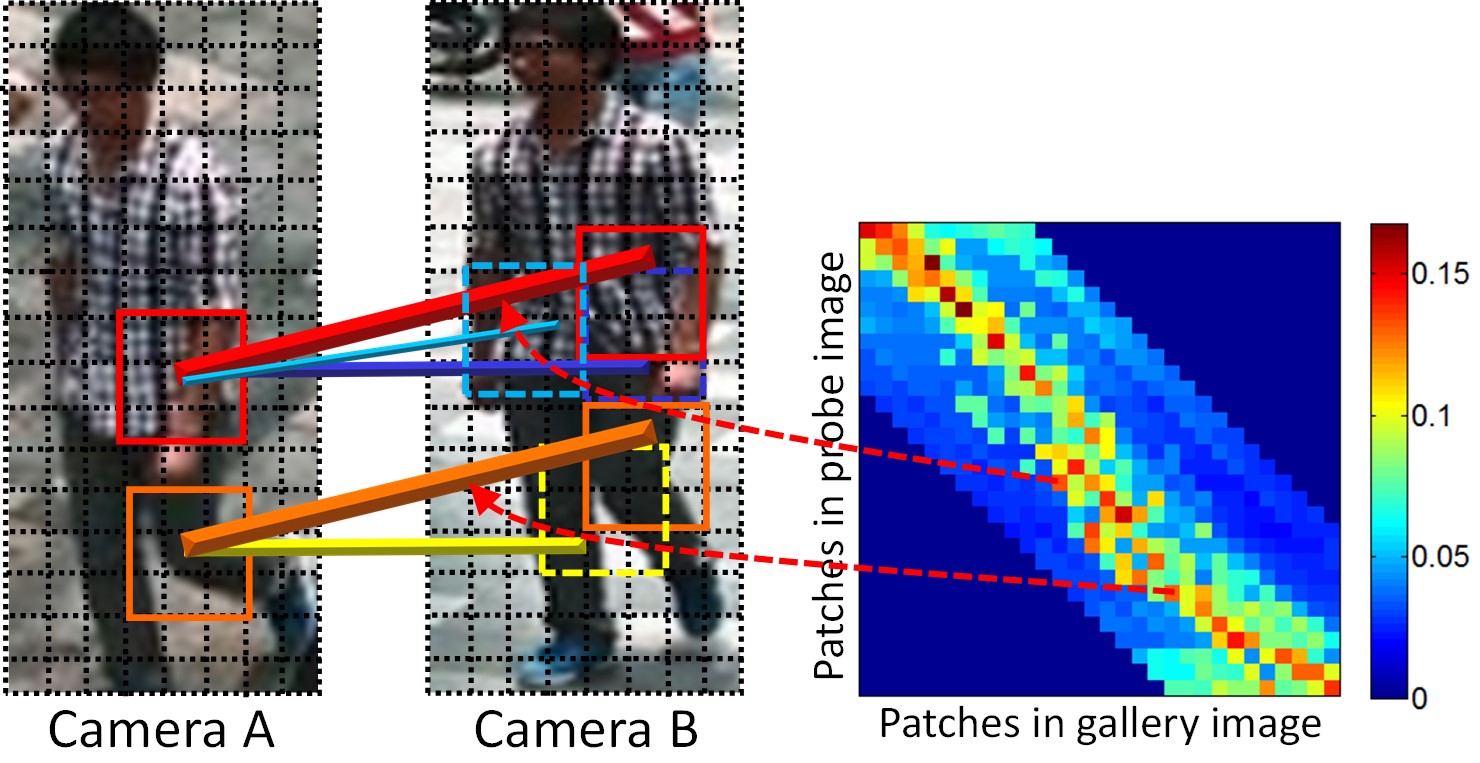}   \label{fig:structures with various dms b}}
  \hspace{4mm}
  \subfloat[the KMFA-$R_{\chi^2}$ metric]{\includegraphics[width=5.0cm,height=2.6cm]{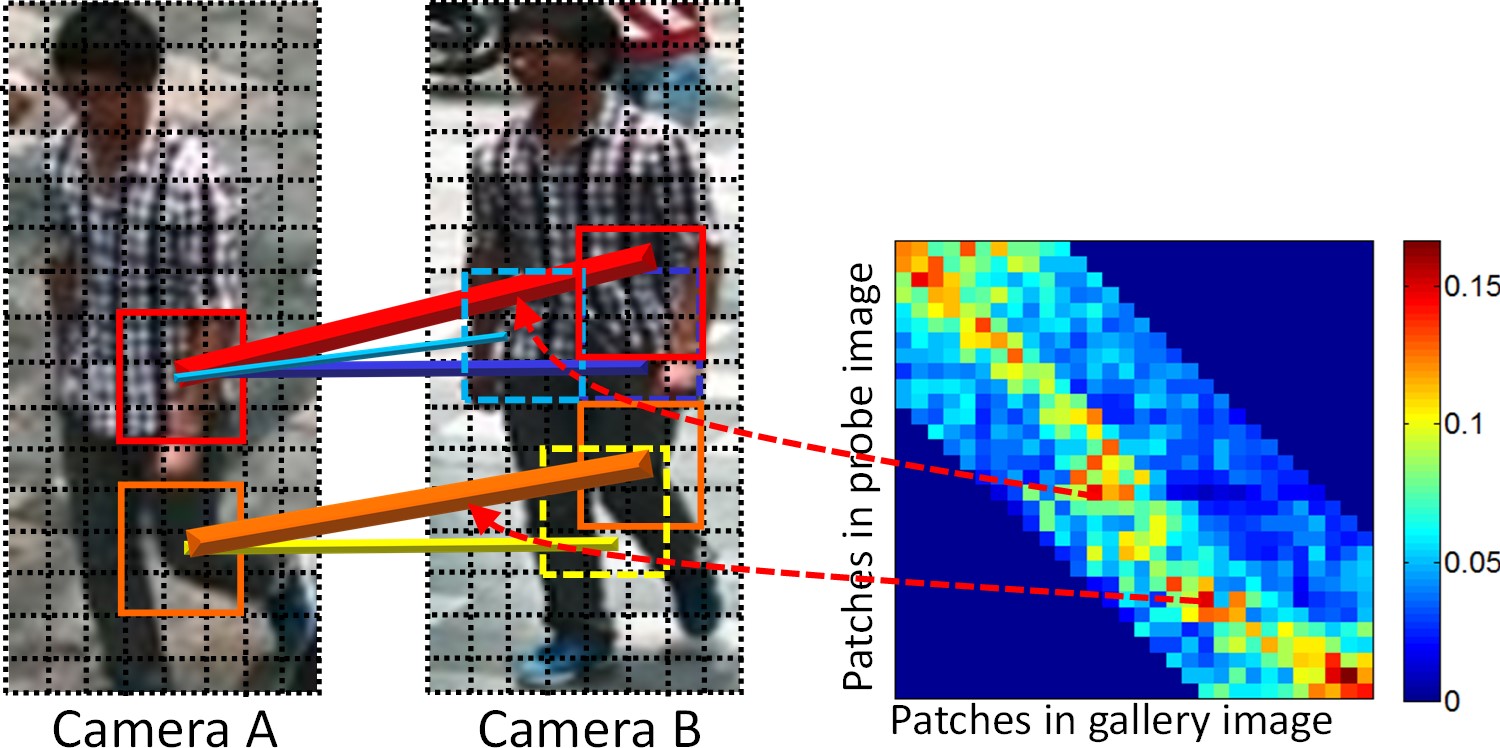}     \label{fig:structures with various dms c}}
  \caption{Comparison of correspondence structures for the Road dataset when using different distance metrics (KISSME~\cite{1}, kLFDA~\cite{kernel-based} and KMFA-$R_{\chi^2}$~\cite{chen2015mirror}) to measure the patch-wise similarity. The line widths are proportional to the patch-wise probability values. (Best viewed in color)}  \label{fig:structures with various dms}
\end{figure*}

Moreover, in order to further evaluate the effectiveness of our correspondence structure learning process, we also show the learned correspondence structures when using different distance metrics (KISSME~\cite{1}, kLFDA~\cite{kernel-based}, and KMFA-$R_{\chi^2}$~\cite{chen2015mirror}) to measure the patch-wise similarity (cf. Eq.~\ref{equation:equ2}), as in Fig.~\ref{fig:structures with various dms}. From Fig.~\ref{fig:structures with various dms}, we see that our approach can effectively capture the overall spatial correspondence patterns under different distance metrics (e.g., the large lower-to-upper misalignments between cameras are properly captured by all the correspondence structures in Fig.~\ref{fig:structures with various dms}).

\subsubsection{Performances of person Re-ID} We evaluate person Re-ID results by the standard Cumulated Matching Characteristic (CMC) curve~\cite{cmc} which measures the correct match rates within different Re-ID rank ranges. The evaluation protocols are the same as~\cite{3}. That is, for each dataset, we perform $10$ randomly-partitioned $50\%$-training and $50\%$-testing experiments and average the results.

To verify the effectiveness of each module of our approach, we compare results of four methods: (1) Not applying correspondence structure and directly using the appearance similarity between co-located patches for person Re-ID, (\emph{Non-structure}); (2) Simply averaging the binary mapping structures for different probe images as the correspondence structure and utilizing it for Re-ID (\emph{Simple-average}); (3) Using a correspondence structure where the correspondence probabilities for all neighboring patch-wise pairs are set as $1$  (i.e., $P_{ij}=1$, if $d(y_i,y_j) < T_d$ and $P_{ij}=0$ otherwise, cf. Eq.~\ref{equation:equ5a}), and applying a global constraint as Section~\ref{Patch-wise mapping} to calculate image-wise similarity (\emph{AC+global}); (4) Using the correspondence structure learned by our approach, but do not include global constraint when performing Re-ID (\emph{Non-global}); (5) Our approach (\emph{Proposed (Single)}).


Tables~\ref{tab:cmcTable01}--\ref{tab:cmcTable05} show the CMC results of different methods under three distance metrics: KISSME~\cite{1}, kLFDA~\cite{kernel-based}, and KMFA-$R_{\chi^2}$~\cite{chen2015mirror} (note that we use Dense SIFT and Dense Color Histogram features~\cite{6} for KISSME \& kLFDA metrics, and use RGB, HSV, YCbCr, Lab, YIQ, $16$ Gabor texture features~\cite{chen2015mirror} for KMFA-$R_{\chi^2}$ metric). From Tables~\ref{tab:cmcTable01}--\ref{tab:cmcTable05}, we notice that: 

\begin{enumerate}
 \item Our approach has obviously improved results than the non-structure method. This indicates that proper correspondence structures can effectively improve Re-ID performances by reducing patch-wise misalignments.
 \item The simple-average method has similar performance to the no-structure method. This implies that an inappropriate correspondence structure reduces the Re-ID performance.
 \item The AC+global method can be viewed as an extended version of the adjacency-constrained search method~\cite{6,8,part2} plus a global constraint. Compared with the AC+global method, our approach still achieves obviously better performance. This indicates that our correspondence structure has stronger capability in handling spatial misalignments than the adjacency-constrained search methods.
 \item The non-global method has improved Re-ID performances than the non-structure method. This further demonstrates the effectiveness of the correspondence structure learned by our approach. Meanwhile, our approach also has superior performance than the non-global method. This demonstrates the usefulness of introducing global constraint in patch matching process.
 \item The improvement of our approach is coherent on different distance metrics (KISSME~\cite{1}, kLFDA~\cite{kernel-based}, and KMFA-$R_{\chi^2}$~\cite{chen2015mirror}). This indicates the robustness of our approach in handling spatial misalignments. In practice, our proposed correspondence structure can also be combined with other features or distance metrics to further improve Re-ID performances.
 \item If comparing our approach (\emph{proposed+single}) with the non-structure method, we can see that the improvement of our approach is the most significant on the Road dataset (cf. Table~\ref{tab:cmcTable04}). This is because the Road dataset includes a large number of image pairs with significant spatial misalignments, and the effectiveness of our approach is particularly pronounced in this scenario.
\end{enumerate}

\begin{table} [t]
 \centering
 \caption{CMC results with different distance metrics on VIPeR dataset}   \label{tab:cmcTable01}
 \footnotesize
 \begin{tabular}{|l|*{5}{c|}}
  \hline
  \textbf{Rank (with KISSME)}& 1& 5& 10& 20& 30\\
  \hline
  Non-structure& 27.5& 57.0& 73.7& 83.9& 87.7\\
  Simple-average& 28.5& 57.9& 74.1& 84.2& 88.3\\
  AC+global  &  28.3 &      57.5 &        74.3 &        84.7 &      88.7\\
  Non-global& 30.8& 62.7& 77.5& 88.9& 91.7\\
  Proposed (single)   &      34.8 &      68.7 &      82.3 &      91.8 &      94.9\\
  Proposed (multi-manu)    &      35.3 &      69.8 &      83.8 &      92.2 &      94.5\\
  Proposed (multi-auto)  &  35.1 &      69.5 &        83.5 &        92.2 &      95.1\\
  Proposed (multi-GT) & {\bf 35.5}& {\bf 70.3}& {\bf 84.3}& {\bf 92.7}& {\bf 95.3}\\
  \hline
  \textbf{Rank (with kLFDA)}& 1& 5& 10& 20& 30\\
  \hline
  Non-structure& 31.5& 64.9& 79.4& 89.2& 92.4\\
  Simple-average& 30.3& 64.4& 77.3& 89.4& 91.7\\
  AC+global  &  32.1 &      69.1 &        83.2 &        91.4 &      94.1\\
  Non-global& 33.8& 69.3& 81.5& 90.3& 93.2\\
  Proposed(single)  &      35.2 &      69.7 &      83.5 &      91.9 &      94.5\\
  Proposed(multi-manu)   & {\bf 36.4}&      71.4 & {\bf 84.5}& {\bf 92.2}& {\bf 94.8}\\
  Proposed (multi-auto)  &  36.1 &      71.2 &        84.2 &        92.0 &      94.5\\
  Proposed(multi-GT)& {\bf 36.4}& {\bf 71.7}&      84.2 &      91.7 &      94.1\\
  \hline
  \textbf{Rank (with KMFA-$R_{\chi^2}$)}& 1& 5& 10& 20& 30\\
  \hline
  Non-structure&   40.3 &      74.1 &        85.3 &        90.0 &      95.2\\
  Simple-average &   39.5 &      73.7 &        84.4 &        89.8 &      95.7\\
  AC+global  &  40.8 &      74.8 &        86.1 &        90.8 &      96.3\\
  Non-global &  42.6 &      76.5 &        87.4 &        92.4 &      97.3\\
  Proposed (single) &  43.3 &      77.4 &        88.2 &        92.7 &      97.8\\
  Proposed (multi-manu) &  44.8 &      78.5 &        89.8 &        93.4 &      97.2\\
  Proposed (multi-auto)  &  44.6 &      78.2 &        89.4 &        {\bf 93.8} &      {\bf 97.6}\\
  Proposed (multi-GT) &  {\bf 45.0} &      {\bf 79.2} &        {\bf 90.3} &        93.5 &      97.4\\
  \hline
 \end{tabular}
\end{table}

\begin{table} [t]
 \centering
 \caption{CMC results with different distance metrics on PRID 450S dataset}  \label{tab:cmcTable02}
 \footnotesize
 \begin{tabular}{|l|*{5}{c|}}
  \hline
  \textbf{Rank (with KISSME)}& 1& 5& 10& 20& 30\\
  \hline
  No-structure& 39.6& 64.9& 76.0& 85.3& 89.3\\
  Simple-average& 38.2& 63.6& 75.1& 84.9& 88.9\\
  AC+global  &  40.5 &      66.3 &        76.3 &        87.8 &      90.4\\
  No-global& 42.7& 69.3& 78.2& 87.4& 91.1\\
  Proposed (single)&    44.4     &     71.6 &      82.2 &      89.8 &      93.3\\
  Proposed (multi-manu) &    45.8     &     72.8 &      82.8 &      90.3 &      93.4\\
  Proposed (multi-auto)  &  45.6 &      72.5 &        82.4 &        90.3 &      93.5\\
  Proposed (multi-GT)& {\bf 46.1}&{\bf 73.3}& {\bf 83.3}& {\bf 90.7}& {\bf 93.8}\\
  \hline
  \textbf{Rank (with kLFDA)}& 1& 5& 10& 20& 30\\
  \hline
  No-structure& 46.5& 74.5& 82.2& 88.1& 92.7\\
  Simple-average& 44.7& 73.7& 81.4& 87.4& 91.6\\
  AC+global  &  47.7 &      75.4 &        82.7 &        88.9 &      92.5\\
  No-global& 49.3& 76.5& 84.3& 89.6& 93.3\\
  Proposed (single)&         51.1 &      77.7 &      85.1 &      91.2 &      94.4\\
  Proposed (multi-manu)&          52.9 & {\bf 79.3}&      86.2 &      92.2 &      95.0\\
  Proposed (multi-auto)  &  52.7 &      78.8 &        85.7 &        91.8 &      94.8\\
  Proposed (multi-GT) & {\bf 53.2}&      79.0 & {\bf 86.9}& {\bf 92.7}& {\bf 95.6}\\
  \hline
  \textbf{Rank (with KMFA-$R_{\chi^2}$)}& 1& 5& 10& 20& 30\\
  \hline
  Non-structure&   57.8 &      79.0 &        87.3 &        92.8 &      96.2\\
  Simple-average &  57.3 &      78.5 &        86.7 &        92.6 &      95.8\\
  AC+global  &  58.2 &      79.5 &        88.2 &        93.4 &      96.7\\
  Non-global &  61.5 &      80.4 &        89.7 &        94.2 &      96.9\\
  Proposed (single) &  63.2 &      82.7 &        90.5 &        94.8 &      97.3\\
  Proposed (multi-manu) &  65.1 &      {\bf 84.9} &        91.2 &        95.4 &      {\bf 97.7}\\
  Proposed (multi-auto)  &  65.3 &      {\bf 84.9} &        91.4 &        95.1 &      97.4\\
  Proposed (multi-GT) &  {\bf 65.7} &      84.6 &      {\bf 91.8} &        {\bf 95.8} &      96.9\\
  \hline
 \end{tabular}
\end{table}

\begin{table} [t]
 \centering
 \caption{CMC results with different distance metrics on 3DPeS dataset}   \label{tab:cmcTable03}
 \footnotesize
 \begin{tabular}{|l|*{6}{c|}}
  \hline
  \textbf{Rank (with KISSME)}& 1& 5& 10& 15& 20& 30\\
  \hline
  Non-structure& 51.6& 75.8& 84.2& 88.4& 90.5& 92.6\\
  Simple-average& 50.5& 74.7& 83.2& 87.4& 89.5& 92.6\\
  AC+global  &  52.4 &      76.0 &        85.3 &        88.7 &     90.8 &      92.4\\
  Non-global& 54.7& 77.9& 87.4& 90.5& 91.6& 93.7\\
  Proposed (single) & {\bf 57.9}& {\bf 81.1}& {\bf 89.5}& {\bf 92.6}& {\bf 93.7}& {\bf 94.7}\\
  \hline
  \textbf{Rank (with kLFDA)}& 1& 5& 10& 15& 20& 30\\
  \hline
  Non-structure& 54.8& 75.6& 85.3& 89.8& 92.1& 93.1\\
  Simple-average& 53.6& 76.1& 84.7& 88.1& 91.4& 92.4\\
  AC+global  &  55.5 &      76.3 &        85.8 &        90.1 &    92.4 &      93.3\\
  Non-global& 57.6& 78.3& 87.4& 91.4& 92.9& 93.8\\
  Proposed (single) & {\bf 60.3}& {\bf 82.4}& {\bf 89.3}& {\bf 92.3}& {\bf 93.6}& {\bf 94.3}\\
  \hline
  \textbf{Rank (with KMFA-$R_{\chi^2}$)}& 1& 5& 10& 15& 20& 30\\
  \hline
  Non-structure &  56.6 &      78.5 &        86.2 &        90.4 &     92.2 &      93.3\\
  Simple-average  &  55.7 &      77.8 &        85.8 &        90.2 &     92.4 &      93.1\\
  AC+global  &  57.2 &      79.0 &        87.8 &        91.3 &     92.7 &      93.8\\
  Non-global  &  59.8 &      82.6 &        89.6 &        92.3 &     93.8 &      94.6\\
  Proposed (single)  &  {\bf 61.4} &      {\bf 84.2} &        {\bf 90.7} &       {\bf 92.7} &    {\bf 94.2} &     {\bf 94.9}\\
  \hline
 \end{tabular}
\end{table}

\begin{table} [t]
 \centering
 \caption{CMC results with different distance metrics on Road dataset}  \label{tab:cmcTable04}
 \footnotesize
 \begin{tabular}{|l|*{6}{c|}}
  \hline
  \textbf{Rank (with KISSME)}& 1& 5& 10& 15& 20& 30\\ \hline
  Non-structure& 50.5& 80.3& 87.0& 91.3& 94.2& 95.7\\
  Simple-average& 49.0& 81.7& 90.4& 92.8& 95.7& 96.2\\
  AC+global  &  53.1 &      82.4 &        89.1 &        94.4 &    95.3 &      96.4\\
  Non-global& 58.2& 85.6& 94.2& 97.1& 98.1& 98.6\\
  Proposed (single)& 61.5 & 91.8 & 95.2 & 98.1& 98.6& 99.0\\
  Proposed (multi-manu) & 63.1 & 92.2 & 95.2 &  98.2 &  {\bf 98.8} & {\bf 99.2}\\
  Proposed (multi-auto)  &  62.8 &      92.0 &    95.4 &     98.4 &    98.6 &      99.3\\
  Proposed (multi-GT) & {\bf 63.6}& {\bf 92.6}& {\bf 95.8}& {\bf 98.4}& 98.6& {\bf 99.2}\\
  \hline
  \textbf{Rank (with kLFDA)}& 1& 5& 10& 15& 20& 30\\
  \hline
  Non-structure& 58.6& 87.6& 92.4& 94.8& 96.7& 97.4\\
  Simple-average& 56.9& 85.8& 91.8& 93.6& 95.4& 96.2\\
  AC+global  &  60.3 &      88.4 &        92.8 &        95.3 &   97.3   &    97.7\\
  Non-global& 63.3& 90.7& 93.7& 96.2& 97.8& 98.4\\
  Proposed (single) &66.8        &     91.5  &    94.8   &    97.7   &  98.4 & 99.2  \\
  Proposed (multi-manu) & 68.2        &     92.8  & 96.2      &      98.1 &  98.6 & 99.4\\
  Proposed (multi-auto)  &  68.3 &      92.6 &     95.9 &     97.7 &   98.6 &      99.2\\
  Proposed (multi-GT)& {\bf 68.6}& {\bf 93.4}& {\bf 96.8}& {\bf 98.3}& {\bf 98.8}& {\bf 99.5}\\
  \hline
    \textbf{Rank (with KMFA-$R_{\chi^2}$)}& 1& 5& 10& 15& 20& 30\\
  \hline
  Non-structure &  65.4 &      90.4 &        92.7 &        94.8 &     96.1 &      96.6\\
  Simple-average  &  63.9 &      89.6 &        92.2 &        94.3 &     95.7 &      96.8\\
  AC+global  &  66.0 &      91.2 &        93.4 &        95.2 &     96.4 &      96.8\\
  Non-global  &  68.1 &      92.1 &        94.3 &        95.9 &     96.7 &      97.2\\
  Proposed (single)  &  71.4 &      92.5 &        95.6 &        96.3 &     97.1 &      97.8\\
  Proposed (multi-manu)  &  73.1 &      92.8 &        96.4 &        97.2 &     97.8 &      98.6\\
  Proposed (multi-auto) &  72.8 &      92.8 &        96.0 &        96.8 &     97.4 &    98.3\\
  Proposed (multi-GT)  &  {\bf 73.5} &     {\bf 93.3} &        {\bf 96.8} &       {\bf 97.5} &    {\bf 98.2} &     {\bf 99.0}\\
  \hline
 \end{tabular}
\end{table}

\begin{table} [t]
 \centering
 \caption{CMC results with different distance metrics on SYSU-sReID dataset}   \label{tab:cmcTable05}
 \footnotesize
 \begin{tabular}{|l|*{6}{c|}}
  \hline
  \textbf{Rank (with KISSME)} &  1 &      5&        10 &        15 &    20 &      30\\
  \hline
  Non-structure  &  33.5 &      70.1 &        80.0 &        83.4 &     88.8 &      92.8\\
  Simple-average &  32.3 &      68.9 &        78.9 &        82.6 &     87.6 &      91.2\\
  AC+global  &  33.9 &      69.3 &        78.1 &        83.8 &     88.9 &      91.8\\
  Non-global &  36.7 &      71.7 &        81.7 &        84.3 &     88.4 &      92.4\\
  Proposed (single)  &  38.6 &      73.1 &        83.7 &        85.2 &     90.4 &      92.7\\
  Proposed (multi-manu)  &  43.3 &      76.2 &        83.9 &        86.1 &     {\bf 90.7} &      {\bf 93.4}\\
  Proposed (multi-auto)  &  43.0 &      75.7 &        83.5 &        86.3 &     90.4 &      92.8\\
  Proposed (multi-GT)  &  {\bf 43.8} &      {\bf 76.8} &        {\bf 84.2} &        {\bf 86.9} &     {\bf 90.7} &      93.2\\
  \hline
  \textbf{Rank (with kLFDA)} &  1 &      5&        10 &        15 &    20 &      30\\
  \hline
  Non-structure &  35.8 &      70.5 &        83.7 &        87.9 &     92.0 &      95.6\\
  Simple-average &  36.0 &      69.7 &        81.7 &        86.4 &     91.6 &      96.1\\
  AC+global  &  36.8 &      70.9 &        83.4 &        87.3 &    91.6 &      95.4\\
  Non-global &  39.0 &      71.9 &        84.3 &        88.6 &     92.8 &      95.6\\
  Proposed (single)  &  41.2 &      73.6 &        84.8 &        89.2 &     93.2 &      96.3\\
  Proposed (multi-manu)  &  45.4 &      75.8 &        86.7 &        91.4 &     94.6 &      96.5\\
  Proposed (multi-auto)  &  45.1 &      75.5 &        86.5 &        91.6 &    {\bf 95.0} &     {\bf 97.2}\\
  Proposed (multi-GT)  &  {\bf 45.8} &     {\bf 76.3} &       {\bf 87.2} &       {\bf 92.3} &     94.3 &      96.8\\
  \hline
    \textbf{Rank (with KMFA-$R_{\chi^2}$)}& 1& 5& 10& 15& 20& 30\\
  \hline
  Non-structure &  42.3 &      71.3 &        84.5 &        88.5 &     91.8 &      95.2\\
  Simple-average  &  41.5 &      71.0 &        84.3 &        88.2 &     92.1 &      95.7\\
  AC+global  &  42.7 &      72.6 &       84.8 &       88.8 &     92.4 &      96.3\\
  Non-global  &  44.5 &      74.3 &        85.6 &        89.7 &     93.3 &      96.0\\
  Proposed (single)  &  46.2 &      76.4 &        87.2 &        90.8 &     94.2 &      96.7\\
  Proposed (multi-manu)  &  49.3 &      77.9 &        88.1 &        92.7 &     94.7 &      97.4\\
  Proposed (multi-auto) &  49.1 &      77.5 &        87.7 &        92.3 &     95.0 &    97.2\\
  Proposed (multi-GT)  & {\bf 49.6} &     {\bf 78.6} &       {\bf 88.9} &       {\bf 93.5} &    {\bf 95.4} &    {\bf 97.8}\\
  \hline
 \end{tabular}
\end{table}

\subsection{Results of using multiple correspondence structures} \label{section:multiple correspondence structure2}

We further evaluate the performance of our approach by using multiple structures to handle spatial misalignments (i.e., the multi-structure strategy described by Fig.~\ref{fig:framework of multiple structures}). Note that since the number of pedestrians for each pose is small in the 3DPeS~\cite{3dpes} dataset, which is insufficient to construct reliable local correspondences, we focus on evaluating our multi-structure strategy on the other four datasets: VIPeR~\cite{viper}, PRID 450S~\cite{prid}, Road, and SYSU-sReID~\cite{guo2014multi}.

\subsubsection{Results for correspondence structure learning} In this section, we evaluate the results of multiple correspondence structure learning based on the manually divided pose group pairs. Similar results can be observed when automatic pose group pair division (cf. Section~\ref{section:multiple correspondence structure}) is applied.

Since each of the VIPeR, PRID 450S, and Road dataset includes two major cross-view pose correspondences (e.g., front-to-right and front-to-back for VIPeR, and left-to-left and right-to-right for PRID 450S), we divide the training images in each dataset into two pose group pairs and construct two local correspondence structures from these pose group pairs.
Comparatively, since more major cross-view pose correspondences are included in the SYSU-sReID dataset (i.e., left-to-left, right-to-right, front-to-side, and back-to-side), we further divide the training images in this dataset into four pose group pairs and construct four local correspondence structures accordingly.

\begin{figure*}[t]
  \centering
  \vspace{-2mm}
  \subfloat[]{\includegraphics[width=4.9cm,height=2.4cm]{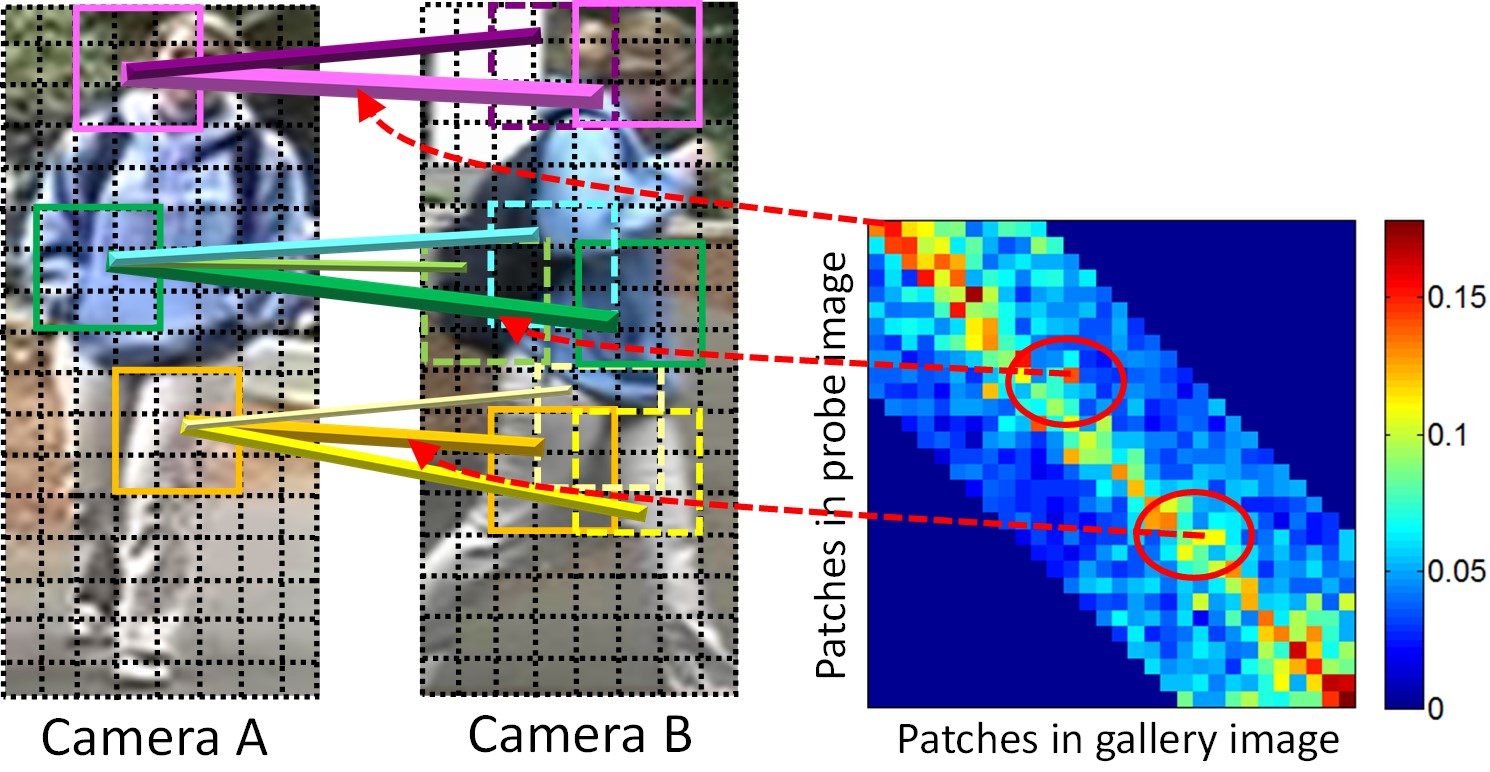}  \label{fig:multiple correspondence structure a}}
  \hspace{3mm}
  \subfloat[]{\includegraphics[width=4.9cm,height=2.4cm]{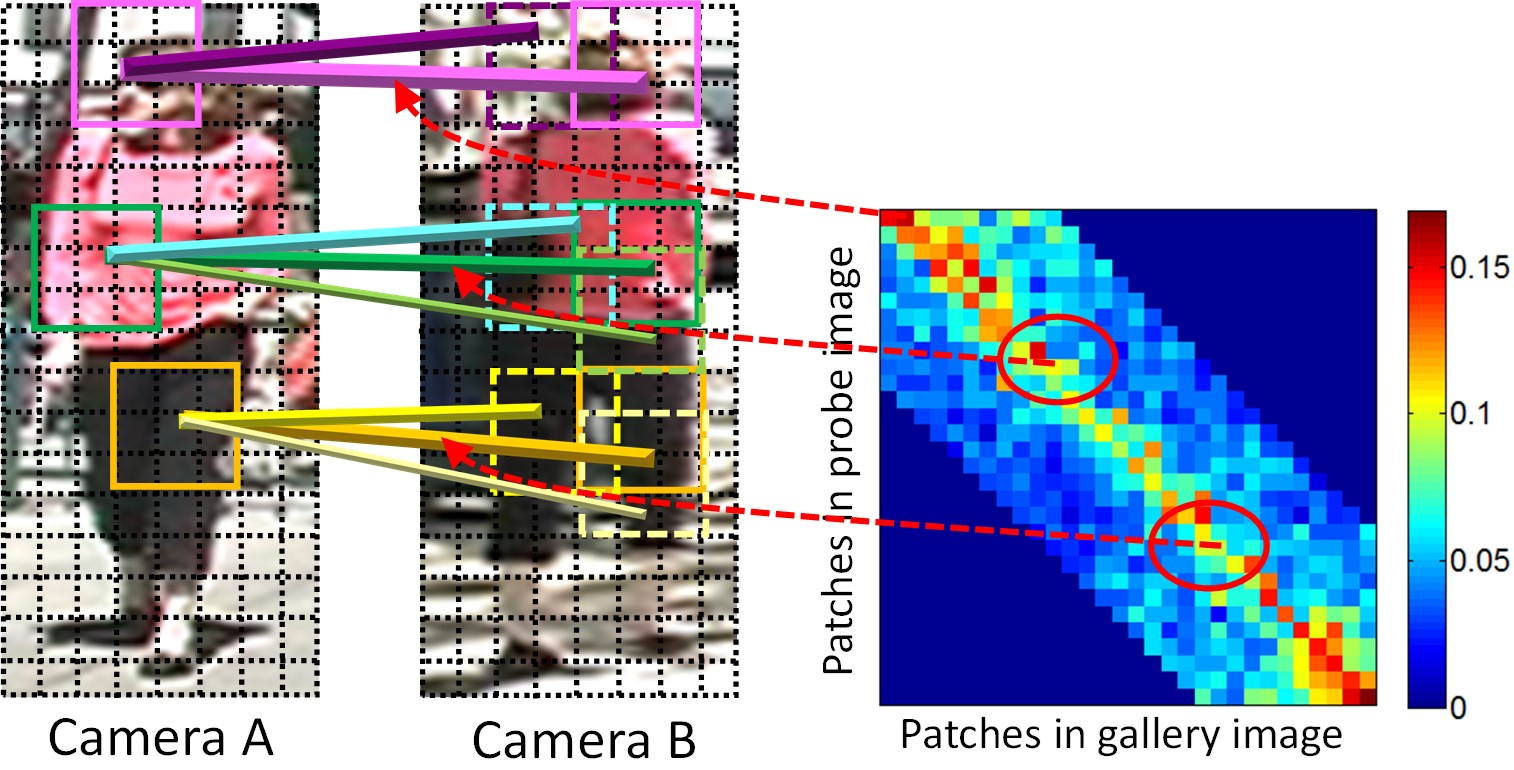}  \label{fig:multiple correspondence structure b}}
  \hspace{3mm}
  \subfloat[]{\includegraphics[width=2.4cm,height=1.92cm]{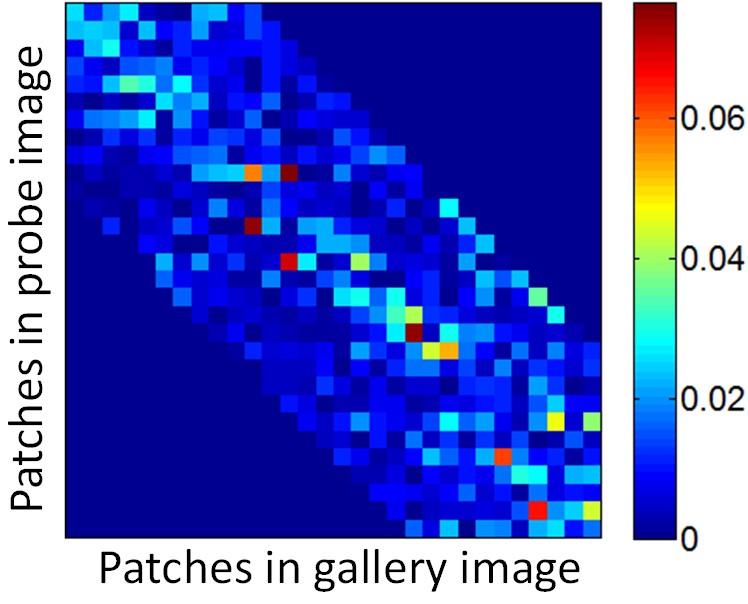} \label{fig:multiple correspondence structure c}}
  \\
  \subfloat[]{\includegraphics[width=4.9cm,height=2.4cm]{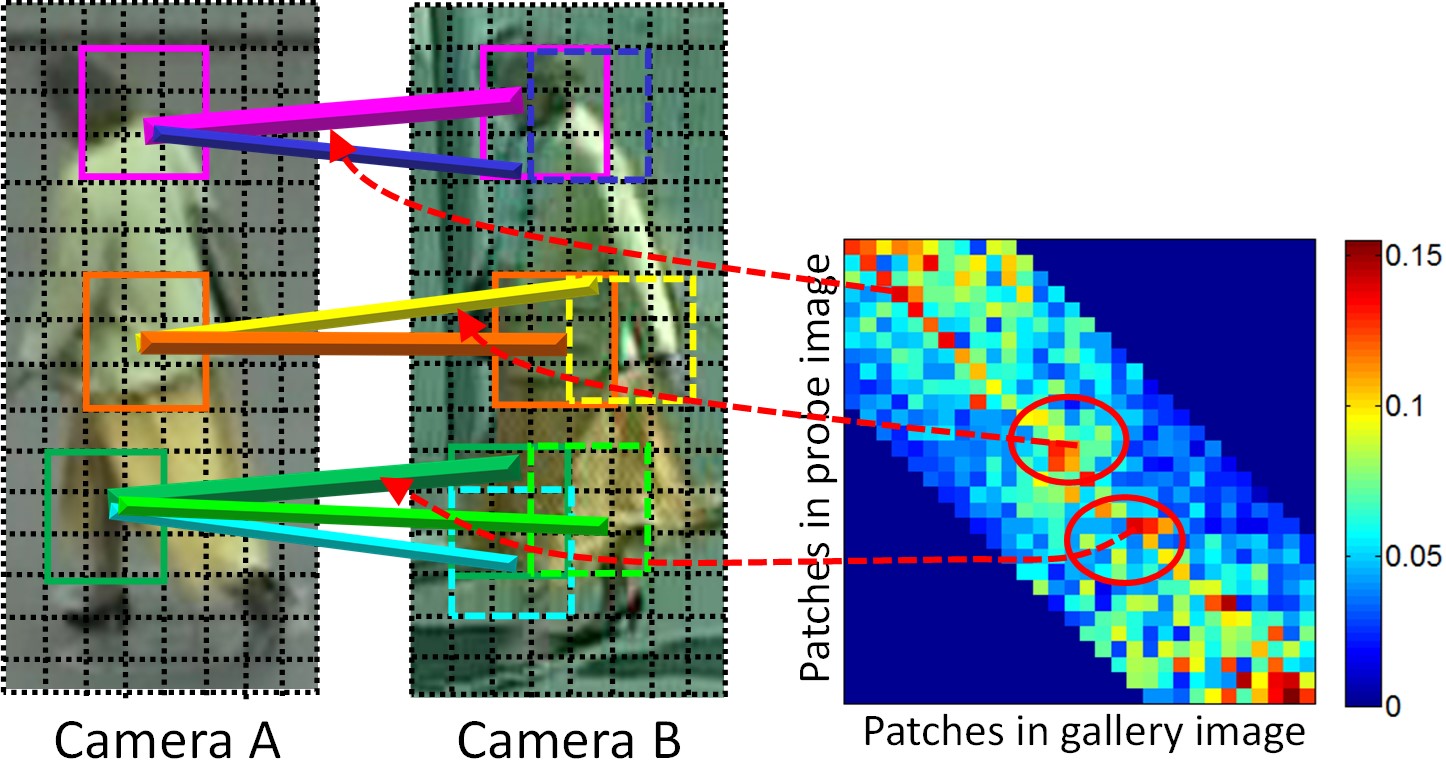}  \label{fig:multiple correspondence structure d}}
  \hspace{3mm}
  \subfloat[]{\includegraphics[width=4.9cm,height=2.4cm]{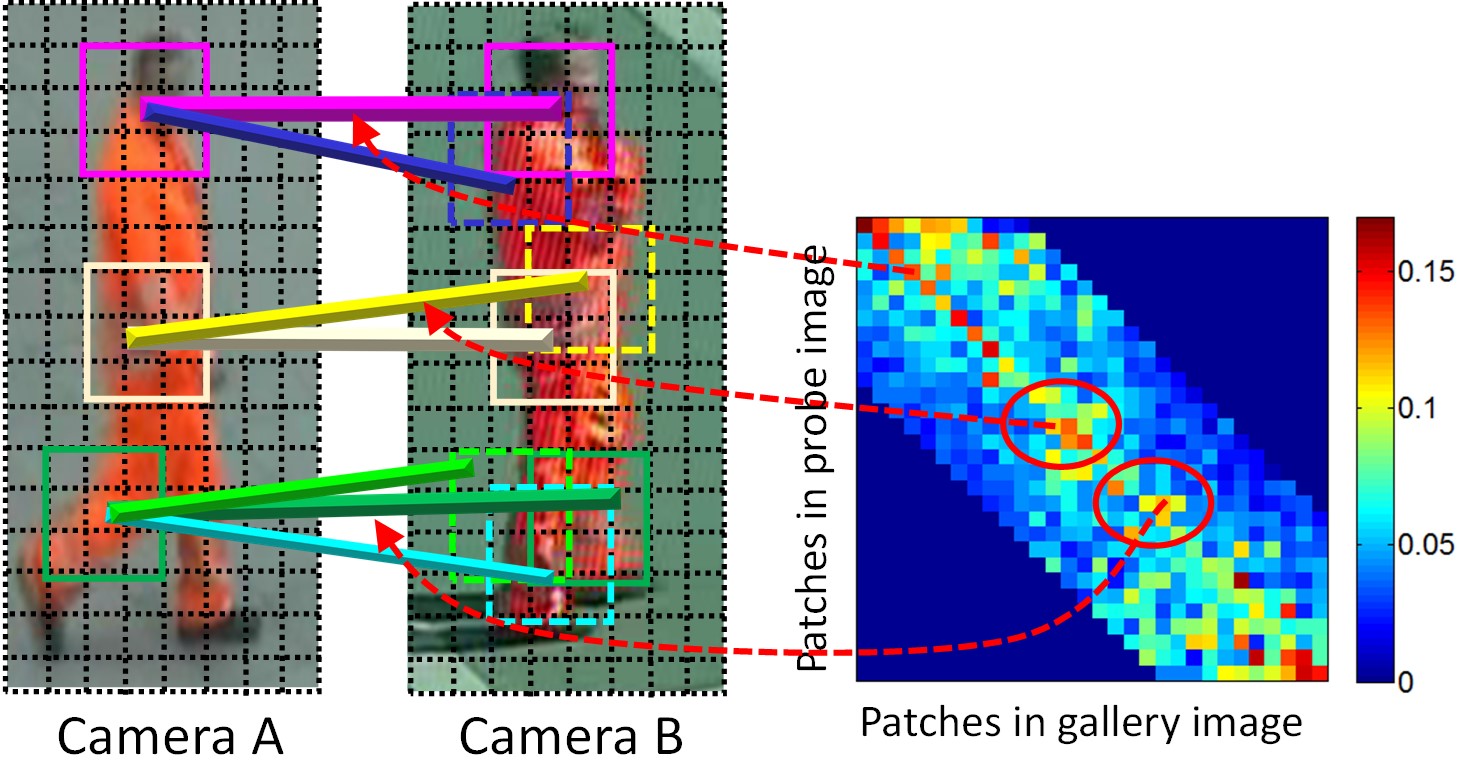}  \label{fig:multiple correspondence structure e}}
  \hspace{3mm}
  \subfloat[]{\includegraphics[width=2.4cm,height=1.92cm]{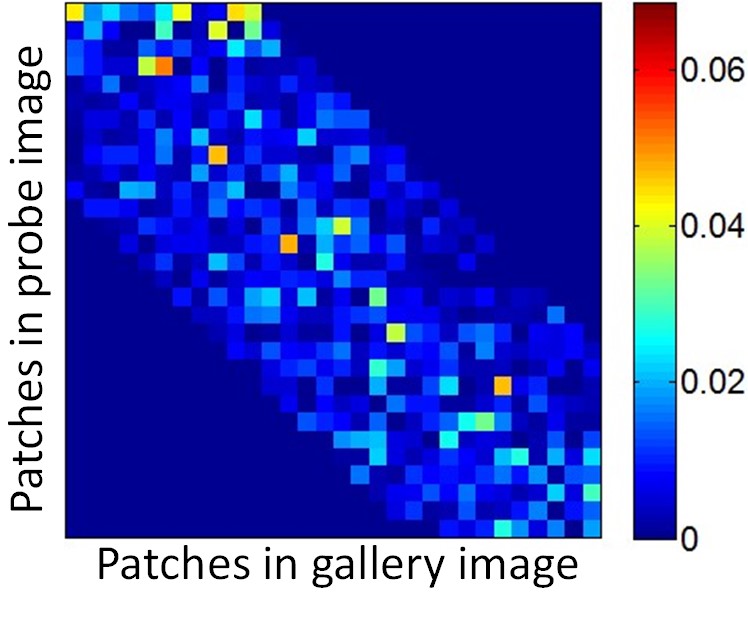} \label{fig:multiple correspondence structure f}}
  \\
  \subfloat[]{\includegraphics[width=4.9cm,height=2.4cm]{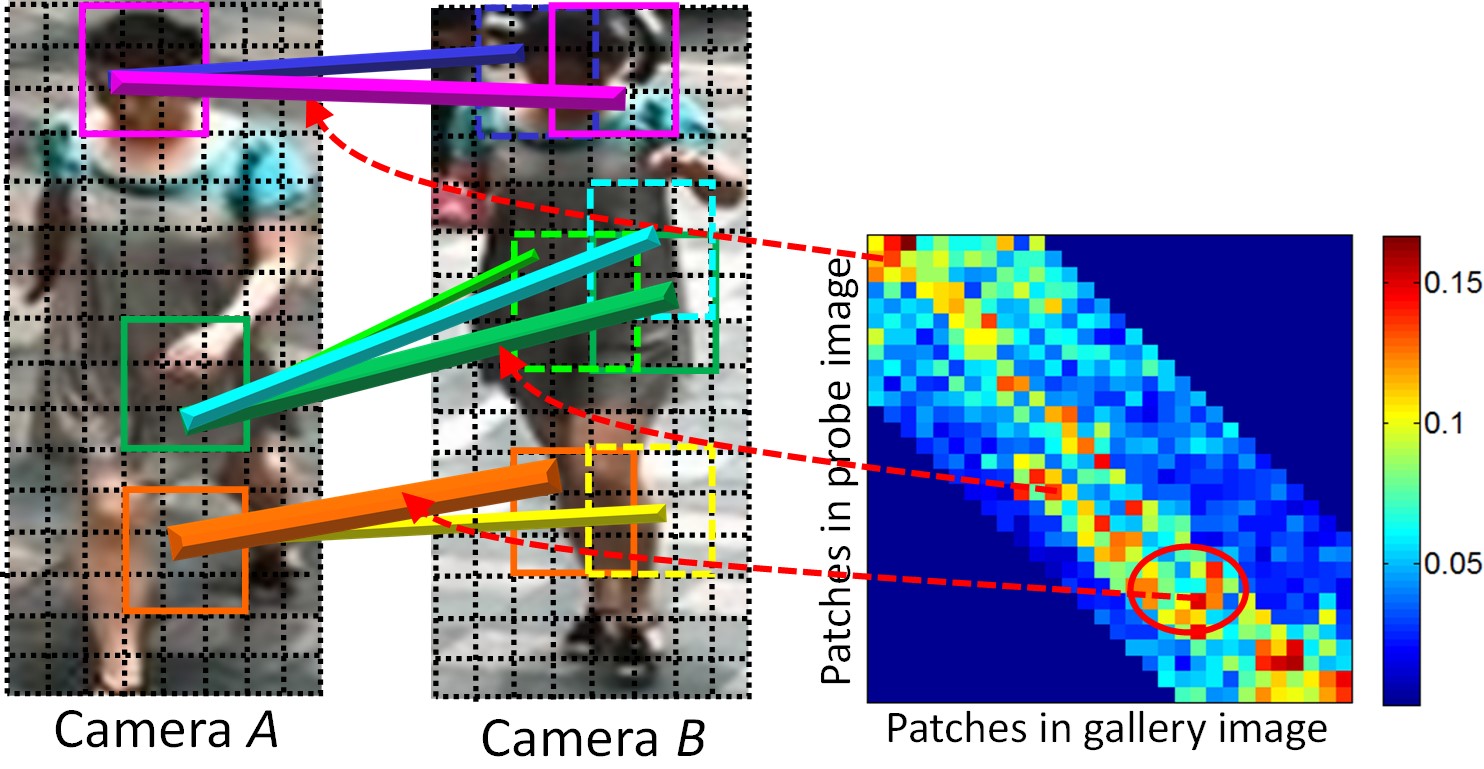}  \label{fig:multiple correspondence structure g}}
  \hspace{3mm}
  \subfloat[]{\includegraphics[width=4.9cm,height=2.4cm]{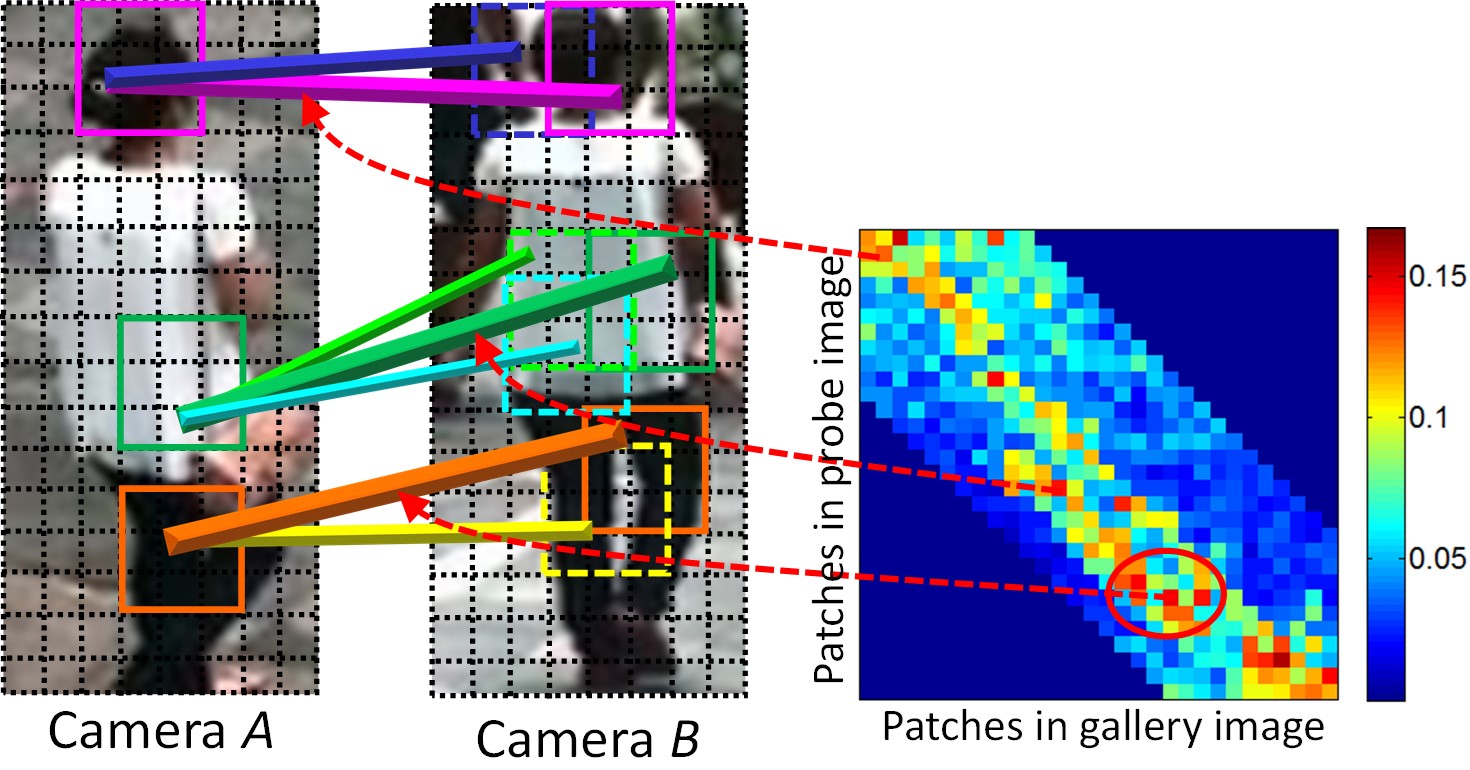}  \label{fig:multiple correspondence structure h}}
  \hspace{3mm}
  \subfloat[]{\includegraphics[width=2.4cm,height=1.92cm]{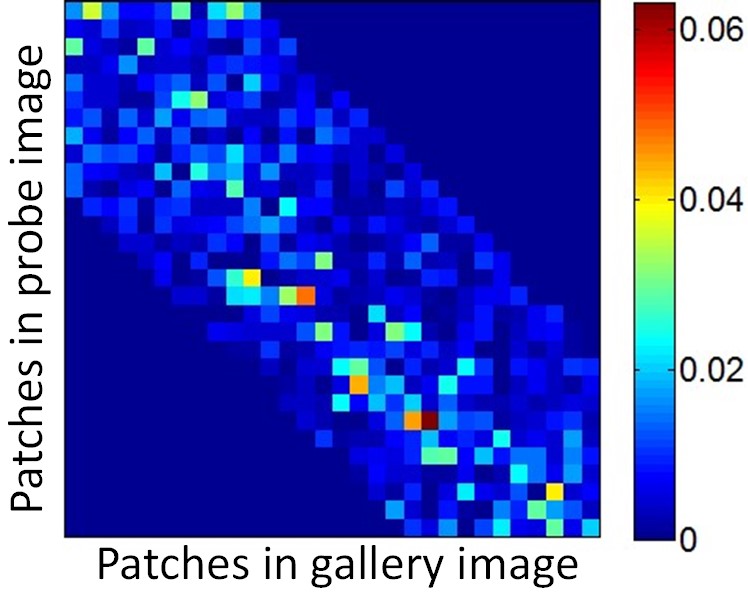} \label{fig:multiple correspondence structure i}}
    \\
  \subfloat[]{\includegraphics[width=4.9cm,height=2.4cm]{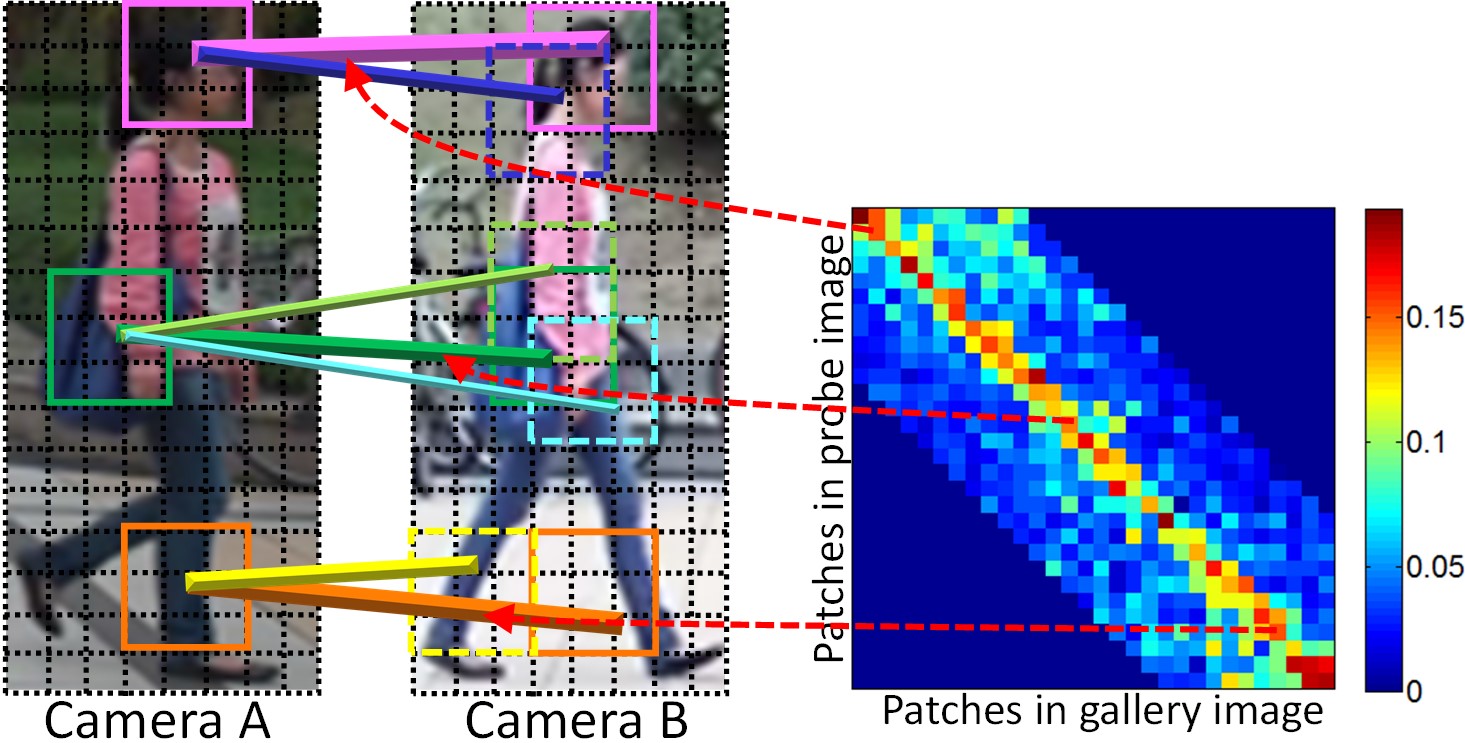}  \label{fig:multiple correspondence structure j}}
  \hspace{3mm}
  \subfloat[]{\includegraphics[width=4.9cm,height=2.4cm]{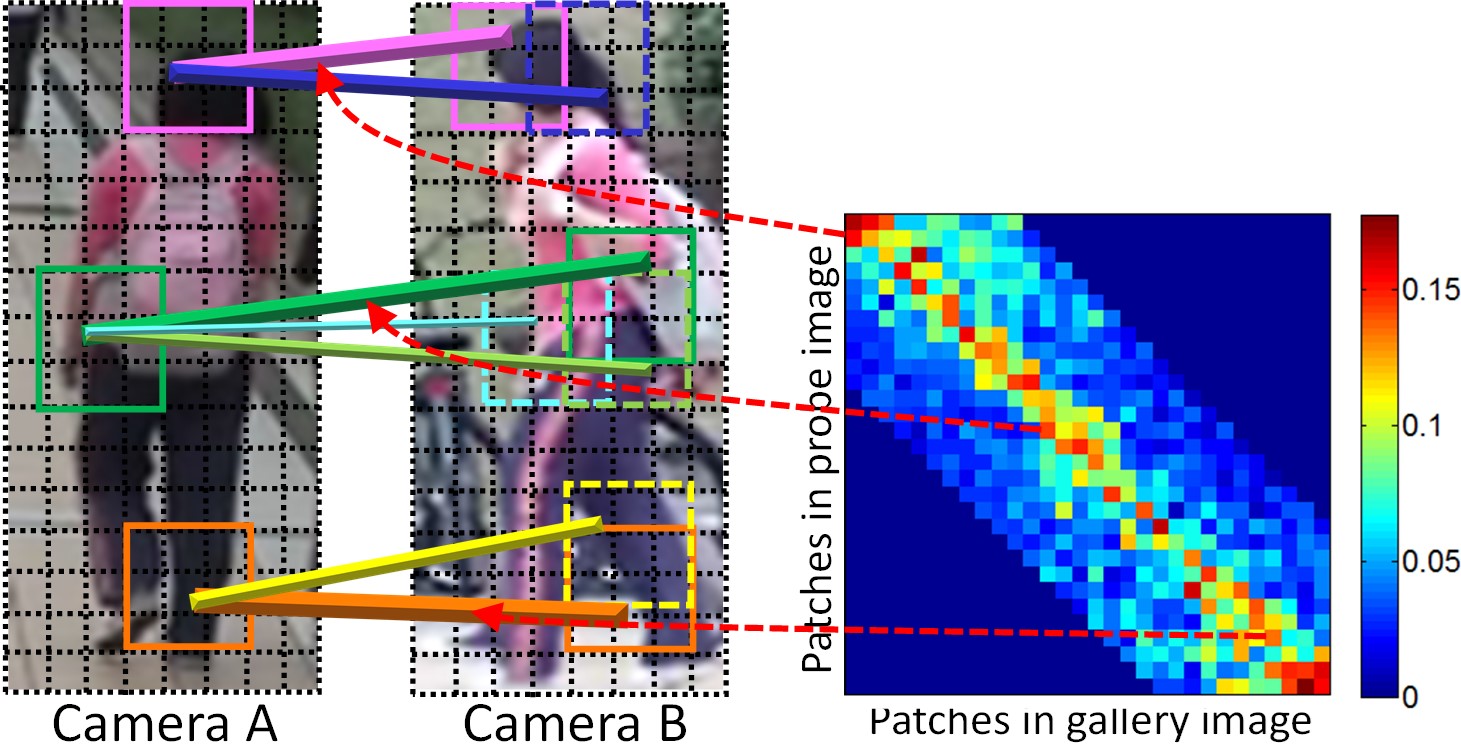}  \label{fig:multiple correspondence structure k}}
  \hspace{3mm}
  \subfloat[]{\includegraphics[width=2.4cm,height=1.92cm]{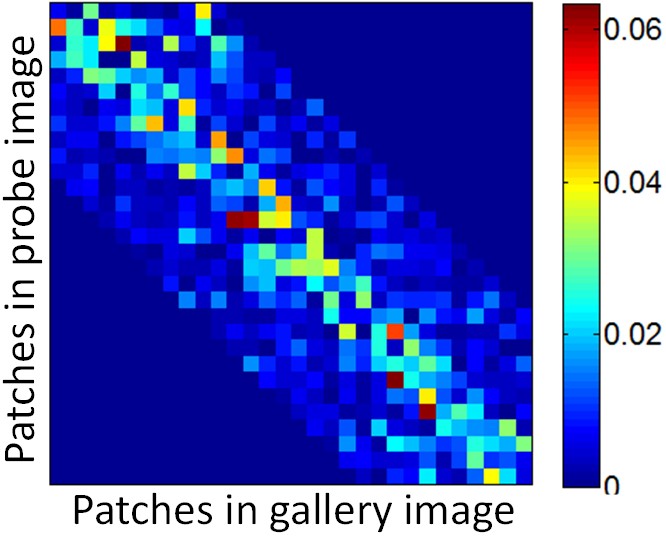} \label{fig:multiple correspondence structure l}}
  \caption{Multiple correspondence structures learned for VIPeR (a, b, c), PRID 450S (d, e, f), Road (g, h, i), SYSU-sReID (j, k, l) datasets (learned with the KISSME metric). Left and middle columns: the two local correspondence structures learned for each dataset (line width is proportional to the probability values). Right column: the absolute difference matrices between the local correspondence structures in left and middle columns. (Best viewed in color)}   \label{fig:multiple correspondence structure}
\end{figure*}

Fig.~\ref{fig:multiple correspondence structure} shows the multiple correspondence structure learning results of our approach. In Fig.~\ref{fig:multiple correspondence structure}, the left and middle columns show the two local correspondence structures learned for each dataset, and the right column shows the absolute difference matrices between the local correspondence structures in left and middle columns. Note that since the SYSU-sReID dataset includes four local correspondence structures which can form six difference matrices, we only show two local correspondence structures and their corresponding difference matrix to save space (cf. Figs~\ref{fig:multiple correspondence structure j},~\ref{fig:multiple correspondence structure k},~\ref{fig:multiple correspondence structure l}).


Fig.~\ref{fig:multiple correspondence structure} indicates that by introducing multiple correspondence structures, the spatial correspondence pattern between cameras can be modeled in a more precise way. Specifically:

\begin{enumerate}
 \item The spatial correspondence pattern between a camera pair can be jointly described by a set of local correspondence structures, where each local correspondence structure captures a sub-pattern inside the global correspondence pattern. For example, the local correspondence structure in Fig.~\ref{fig:multiple correspondence structure a} can properly emphasize the large spatial displacement due to the cross-view pose translation in the front-to-side correspondence sub-pattern. At the same time, the local correspondence structure in Fig.~\ref{fig:multiple correspondence structure b} also properly captures the front-to-back correspondence sub-pattern. In this way, our approach can have stronger capability to handle spatial misalignments with large variations.
 \item Due to the effectiveness of our correspondence structure learning process, the learned local correspondence structures can properly differentiate and highlight the subtle differences among different spatial correspondence sub-patterns. For example, the absolute difference matrix in Fig.~\ref{fig:multiple correspondence structure c} shows large values in the middle and downright corner. This implies that the major difference between the front-to-side and front-to-back correspondence sub-patterns in Figs.~\ref{fig:multiple correspondence structure a} and~\ref{fig:multiple correspondence structure b} is around the torso and leg regions. Similarly, large difference values appear in the top left and bottom right corners in Fig.~\ref{fig:multiple correspondence structure f}, indicating that the major difference between the left-to-left and right-to-right correspondence sub-patterns in Figs.~\ref{fig:multiple correspondence structure d} and~\ref{fig:multiple correspondence structure e} is around the head and leg regions.
\end{enumerate}

\subsubsection{Performances of Person Re-ID} We further perform person Re-ID experiments by using the multiple correspondence structures in Fig.~\ref{fig:multiple correspondence structure}. Note that besides the local correspondence structures in Fig.~\ref{fig:multiple correspondence structure}, the global correspondence structure learned over all training samples is also included (cf. Fig.~\ref{fig:framework of multiple structures}).

Tables~\ref{tab:cmcTable01},~\ref{tab:cmcTable02},~\ref{tab:cmcTable04}, and~\ref{tab:cmcTable05} show the CMC results of our approach with multiple correspondence structures where \emph{proposed+multi-manu} and \emph{proposed+multi-auto} represent the results of our approach under manually and automatically divided pose group pairs, respectively (cf. Section~\ref{section:multiple correspondence structure}). Moreover, in order to evaluate the impact of pose classification error, we also include the results by using the ground-truth pose classification label\footnote{The ground-truth labels are defined based on the manually divided pose group pairs.} to select the optimal correspondence structure during person Re-ID (i.e., \emph{proposed+multi-GT} in Tables~\ref{tab:cmcTable01},~\ref{tab:cmcTable02},~\ref{tab:cmcTable04}, and~\ref{tab:cmcTable05} ).

From Tables ~\ref{tab:cmcTable01},~\ref{tab:cmcTable02}, ~\ref{tab:cmcTable04}, and~\ref{tab:cmcTable05}, we can observe:

\begin{enumerate}
 \item Our approach with multiple correspondence structures (\emph{proposed+multi-manu} and \emph{proposed+multi-auto}) can achieve better results than using a single correspondence structure (\emph{proposed+single}). This further demonstrates that our multi-structure strategy can model the cross-view spatial correspondence in a more precise way, and spatial misalignments can be more effectively handled and excluded by multiple correspondence structures.
 \item Our approach, which uses automatic pose classification to select the optimal correspondence structure (\emph{proposed+multi-manu}), can achieve similar results to the method that uses the ground-truth pose classification label for correspondence structure selection (\emph{proposed+multi-GT}). This implies that the pose classification method used in our approach can provide sufficient classification accuracy to guarantee satisfactory Re-ID performances.
 \item Our approach under automatically obtained pose group pairs (\emph{proposed+multi-auto}) can achieve similar performances to the one under manually divided pose group pairs (\emph{proposed+multi-manu}). This implies that if we can find proper features to describe human poses, automatic pose grouping can also obtain proper pose group pairs and achieve satisfactory Re-ID results.
 \item The improvement of multiple correspondence structures (\emph{proposed+multi-manu} and \emph{proposed+multi-auto}) is the most obvious on the SYSU-sReID dataset. This is because the SYSU-sReID dataset includes more cross-view pose correspondence patterns. Thus, by introducing more local correspondence structures to handle these pose correspondence patterns, the cross-view spatial misalignments can be more precisely captured, leading to more obvious improvements.
\end{enumerate}

\subsection{Comparing with the state-of-art methods}

We also compare our results with the state-of-the-art methods on different datasets: RankBoost~\cite{rankboost}, LF~\cite{color_in}, JLCF~\cite{color_in}, KISSME~\cite{prid}, SalMatch~\cite{8}, DSVR FSA~\cite{part2}, svmml~\cite{svmml}, ELS~\cite{chen2015tip}, kLFDA~\cite{kernel-based}, MFA~\cite{kernel-based}, IDLA~\cite{ahmed2015cvpr}, JLR~\cite{faqiang2016cvpr}, PRISM~\cite{PRISM}, DG-Dropout~\cite{xiao2016learning}, DPML~\cite{added[2]}, LOMO~\cite{zhang2016learning}, Mirror-KMFA($R_{\chi^2}$)~\cite{chen2015mirror}, DCSL~\cite{added[4]}, deCPPs~\cite{added[1]}, MCP-CNN~\cite{cheng2016person}, and FNN~\cite{wu2016enhanced} on the VIPeR dataset; KISSME~\cite{prid}, EIML~\cite{eiml}, SCNCD~\cite{SalientColor}, SCNCDFinal~\cite{SalientColor}, svmml~\cite{svmml}, MFA~\cite{kernel-based}, kLFDA~\cite{kernel-based}, Mirror-KMFA($R_{\chi^2}$)~\cite{chen2015mirror}, and FNN~\cite{wu2016enhanced} on the PRID 450S dataset; PCCA~\cite{PCCA}, rPCCA~\cite{kernel-based}, svmml~\cite{svmml}, MFA~\cite{kernel-based}, and  kLFDA~\cite{kernel-based} on the 3DPeS dataset; eSDC-knn~\cite{6}, svmml~\cite{svmml}, MFA~\cite{kernel-based}, and kLFDA~\cite{kernel-based} on the Road dataset; and eSDC-knn~\cite{6}, svmml~\cite{svmml}, MFA~\cite{kernel-based}, kLFDA~\cite{kernel-based}, MLACM~\cite{guo2014multi}, and TA+W~\cite{added[3]} on the SYSU-sReID dataset.

\begin{table}
 \centering
 \caption{CMC results on the VIPeR dataset (Proposed(single-KMFA) and Proposed(multi-manu-KMFA) indicates our approach using a single and multiple correspondence structures with KMFA-$R_{\chi^2}$ metric)}   \label{tab:cmcTable1}
 \footnotesize
 \begin{tabular}{|@{\;\;}l@{\;\;}|*{5}{@{\;\;}c@{\;\;}|}}
  \hline
  \textbf{Rank}& 1& 5& 10& 20& 30\\
  \hline
  RankBoost~\cite{rankboost}& 23.9& 45.6& 56.2& 68.7& -\\
  JLCF~\cite{color_in}& 26.3& 51.9& 67.1& -& -\\
  KISSME~\cite{prid}& 27& -& 70& 83& -\\
  SalMatch~\cite{8}& 28.1& 52.3& -& -& -\\
  DSVR FSA~\cite{part2}& 29.4& 50.7& 61.9& 74.9& -\\
  svmml~\cite{svmml}& 30.1& 63.2& 77.4& 88.1& -\\
  ELS~\cite{chen2015tip}& 31.3& 62.1& 75.3& 86.7& -\\
  kLFDA~\cite{kernel-based}& 32.3& 65.8& 79.7& 90.9& -\\
  MFA~\cite{kernel-based}& 32.2& 66.0& 79.7& -& 90.6\\
  IDLA~\cite{ahmed2015cvpr}& 34.8& 64.5& -& -& -\\
  JLR~\cite{faqiang2016cvpr}& 35.8& 68.0& -& -& -\\
  PRISM~\cite{PRISM}& 36.7& 66.1& 79.1& 90.2& -\\
  JLCF-fusing~\cite{color_in}& 38.0& 70.3& 81.0& -& -\\
  DG-Dropout~\cite{xiao2016learning} & 38.6& -& -& -& -\\
  DPML~\cite{added[2]}& 41.5& -& 80.9& 90.5& -\\
  LOMO~\cite{zhang2016learning} & 42.3& 71.5& 82.9& 92.1& -\\
  Mirror-KMFA($R_{\chi^2}$)~\cite{chen2015mirror} & 43.0& 75.8 & 87.3&  94.8& -\\
  DCSL~\cite{added[4]} & 44.6& 73.4& 82.6& -& -\\
  svmml+SalMatch~\cite{8}& 44.6& 73.4& -& -& -\\
  deCPPs~\cite{added[1]} & 44.6& 75.1& 86.8& -& -\\
  deCPPs+MER~\cite{added[1]} & 47.5& 75.3& 86.8& -& -\\
  MCP-CNN~\cite{cheng2016person}& 47.8& 74.7& 84.8& 91.1& 94.3\\
  FNN~\cite{wu2016enhanced} & 51.1&  81.0& {\bf 91.4}& {\bf 96.9}& -\\
  LOMO-fusing~\cite{zhang2016learning} & 51.2& 82.1&  90.5&  95.9& -\\
  \hline \hline
  Proposed (single-KMFA) & 43.3 & 77.4 & 88.2 & 92.7 &  97.8\\
  Proposed (multi-manu-KMFA) &  44.8& 78.5&  89.8&  93.4&  97.2\\
  Proposed (multi-manu-KMFA)-fusing & {\bf 51.4} & {\bf 82.5} & 91.2&  96.0 & {\bf 98.1}\\
  \hline
 \end{tabular}
\end{table}

\begin{table}
 \centering
 \caption{CMC results on the PRID 450S dataset}  \label{tab:cmcTable2}
 \footnotesize
 \begin{tabular}{|l|*{5}{c|}}
  \hline
  \textbf{Rank}& 1& 5& 10& 20& 30\\
  \hline
  KISSME~\cite{prid}& 33& -& 71& 79& -\\
  EIML~\cite{eiml}& 35& -& 68& 77& -\\
  SCNCD~\cite{SalientColor}& 41.5& 66.6& 75.9& 84.4& 88.4\\
  SCNCDFinal~\cite{SalientColor}& 41.6& 68.9& 79.4& 87.8& 91.8\\
  svmml~\cite{svmml}& 42.8& 69.7& 79.2& 86.6& 91.2\\
  MFA~\cite{kernel-based}& 44.7& 71.9& 80.2& 86.9& 91.9\\
  kLFDA~\cite{kernel-based} & 46.9& 73.1& 81.7& 87.2& 91.4\\
  Mirror-KMFA($R_{\chi^2}$)~\cite{chen2015mirror} &  55.4&  79.3&  87.8 &  93.9& - \\
  FNN~\cite{wu2016enhanced} & {\bf 66.6}& {\bf 86.8}& {\bf 92.8}& {\bf 96.9}& -\\
  \hline \hline
  Proposed (single-KMFA) &  63.2 & 82.7 & 90.5 & 94.8 & 97.3\\
  Proposed (multi-manu-KMFA) & 65.1 & 84.9 & 91.2 & 95.4 & {\bf 97.7}\\
  \hline
 \end{tabular}
\end{table}

\begin{table}
\centering
\caption{CMC results on the 3DPeS dataset} \label{tab:cmcTable3}
\footnotesize
 \begin{tabular}{|l|*{6}{c|}}
  \hline
  \textbf{Rank}& 1& 5& 10& 20& 30\\ \hline
  svmml~\cite{svmml}& 34.7& 66.4& 78.8& 88.5& -\\
  PCCA~\cite{PCCA}& 41.6& 70.5& 81.3&  90.4& -\\
  rPCCA~\cite{kernel-based}& 47.3& 75.0& 84.5&  91.9& -\\
  MFA~\cite{kernel-based}& 48.4& 72.4& 81.5&  89.8& -\\
  kLFDA~\cite{kernel-based}& 54.0& 77.7& 85.9&  92.4& -\\
  DG-Dropout~\cite{xiao2016learning} & 56.0& -& -& -& -\\
  \hline \hline
  Proposed (single-KMFA)~~~~~~~ & {\bf 61.4}& {\bf 84.2}& {\bf 90.7}& {\bf 94.2}& {\bf 94.9}\\
  \hline
 \end{tabular}
\end{table}

\begin{table}
 \centering
 \caption{CMC results on the Road dataset} \label{tab:cmcTable4}
 \footnotesize
 \begin{tabular}{|l|*{6}{c|}}
  \hline
  \textbf{Rank}& 1& 5& 10& 20& 30\\ \hline
  eSDC-knn~\cite{6}& 52.4& 74.5& 83.7& 89.9& 91.8\\
  svmml~\cite{svmml}& 57.2& 85.2& 92.1& 96.1& 97.6\\
  MFA~\cite{kernel-based}& 58.3& 85.5& 92.4& 96.2& 97.3\\
  kLFDA~\cite{kernel-based}& 59.1& 86.5& 91.8& 95.8& 98.1\\
  \hline \hline
  Proposed (single-KMFA) & 71.4 & 92.5 & 95.6  & 97.1 & 97.8\\
  Proposed (multi-manu-KMFA) & {\bf 73.1}& {\bf 92.8}& {\bf 96.4}& {\bf 97.8}& {\bf 98.6} \\
  \hline
 \end{tabular}
\end{table}

\begin{table}
 \centering
 \caption{CMC results on the SYSU-sReID dataset (Note: The TA-W method uses the same feature and distance metric as the Proposed (single-KMFA) method)} \label{tab:cmcTable5}
 \footnotesize
 \begin{tabular}{|l|*{6}{c|}}
  \hline
  \textbf{Rank}& 1& 5& 10& 20& 30\\ \hline
  KISSME~\cite{prid}& 28.6& 62.0& 72.8& 82.4& 85.0\\
  svmml~\cite{svmml}& 30.8& 63.5& 74.9& 84.2& 88.3\\
  MFA~\cite{kernel-based}& 32.4& 65.2& 75.4& 85.7& 89.4\\
  kLFDA~\cite{kernel-based}& 33.4& 65.9& 76.3& 86.3& 90.0\\
  MLACM~\cite{guo2014multi} & 32.5& 57.7& 68.3& 78.2& --\\
  TA-W~\cite{added[3]} & 43.0 & 73.1 & 85.3 & 92.2 & 96.1\\
  \hline \hline
  Proposed (single-KMFA) & 46.2 & 76.4 & 87.2  & 94.2 & 96.7\\
  Proposed (multi-manu-KMFA) & {\bf 49.3}& {\bf 77.9}& {\bf 88.1}& {\bf 94.7}& {\bf 97.4} \\
  \hline
 \end{tabular}
\end{table}

\begin{figure*}[t]
  \centering
  \subfloat[VIPeR]{\includegraphics[width=3.6cm,height=2.7cm]{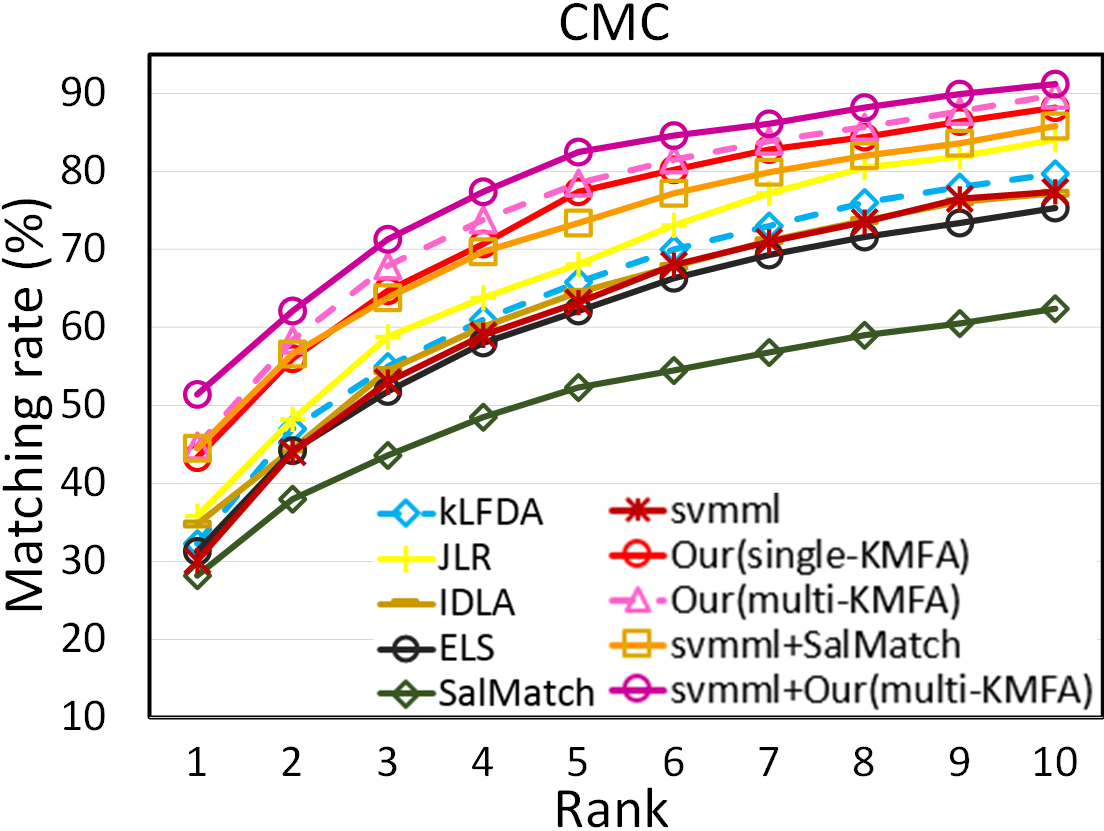} \label{fig:cmc curve1 a}}
  \subfloat[PRID 450S]{\includegraphics[width=3.6cm,height=2.7cm]{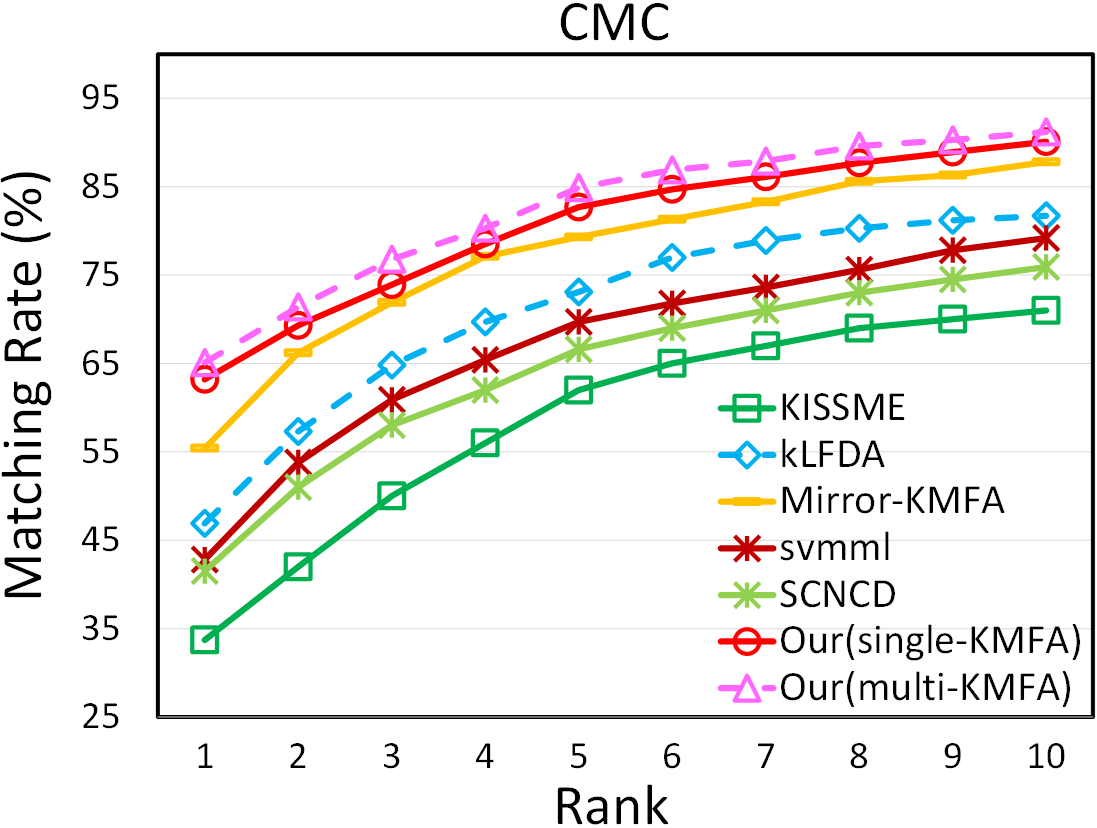} \label{fig:cmc curve1 b}}
  \subfloat[3DPeS]{\includegraphics[width=3.6cm,height=2.7cm]{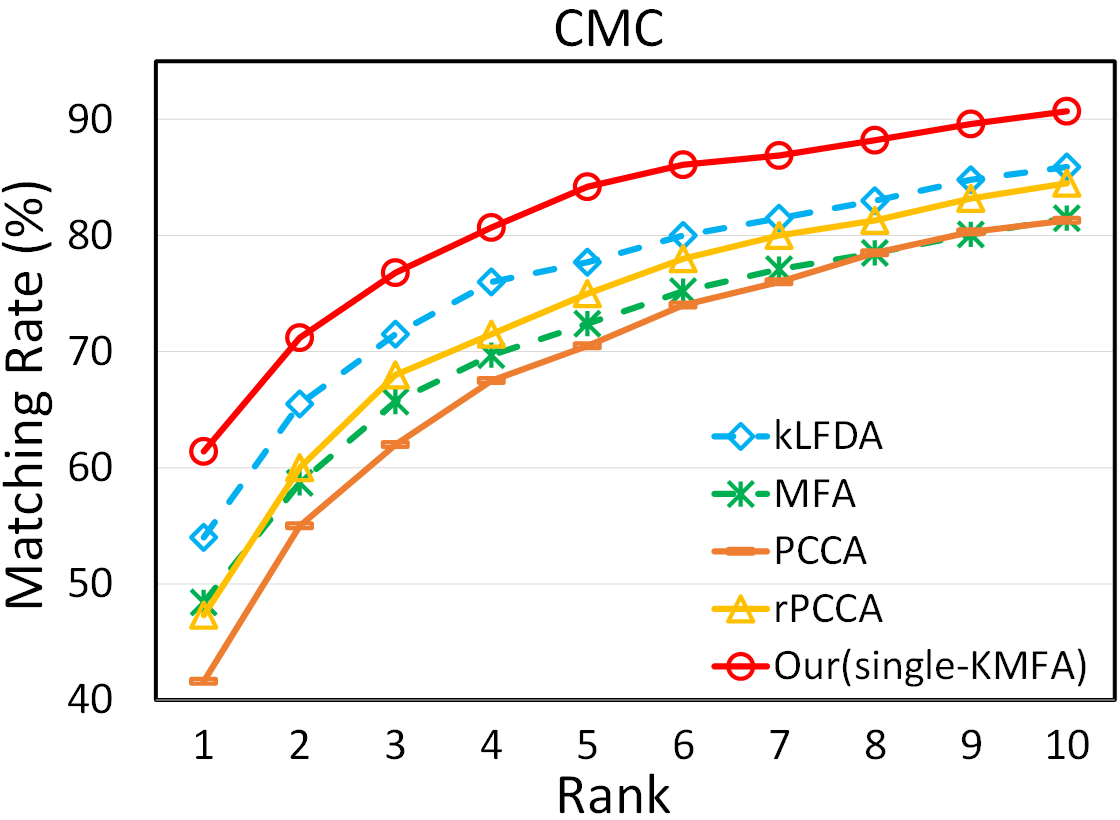} \label{fig:cmc curve1 c}}
  \subfloat[Road]{\includegraphics[width=3.6cm,height=2.7cm]{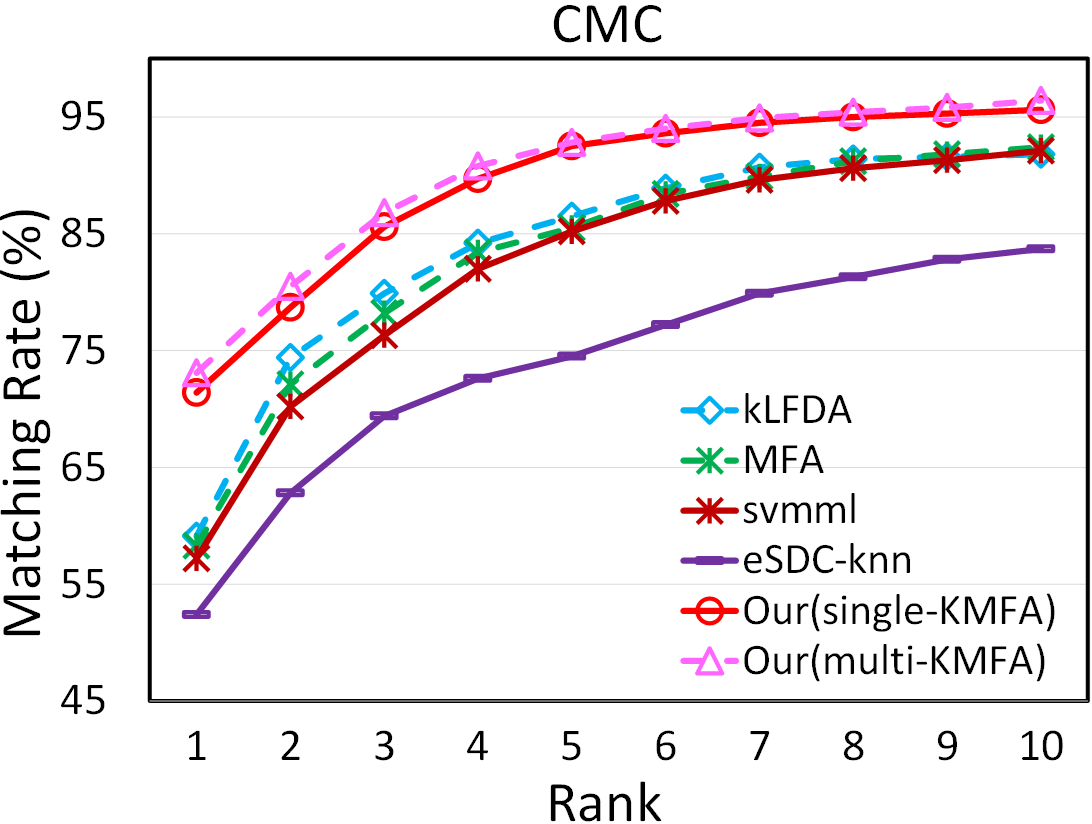} \label{fig:cmc curve1 d}}
  \subfloat[SYSU-sReID]{\includegraphics[width=3.6cm,height=2.7cm]{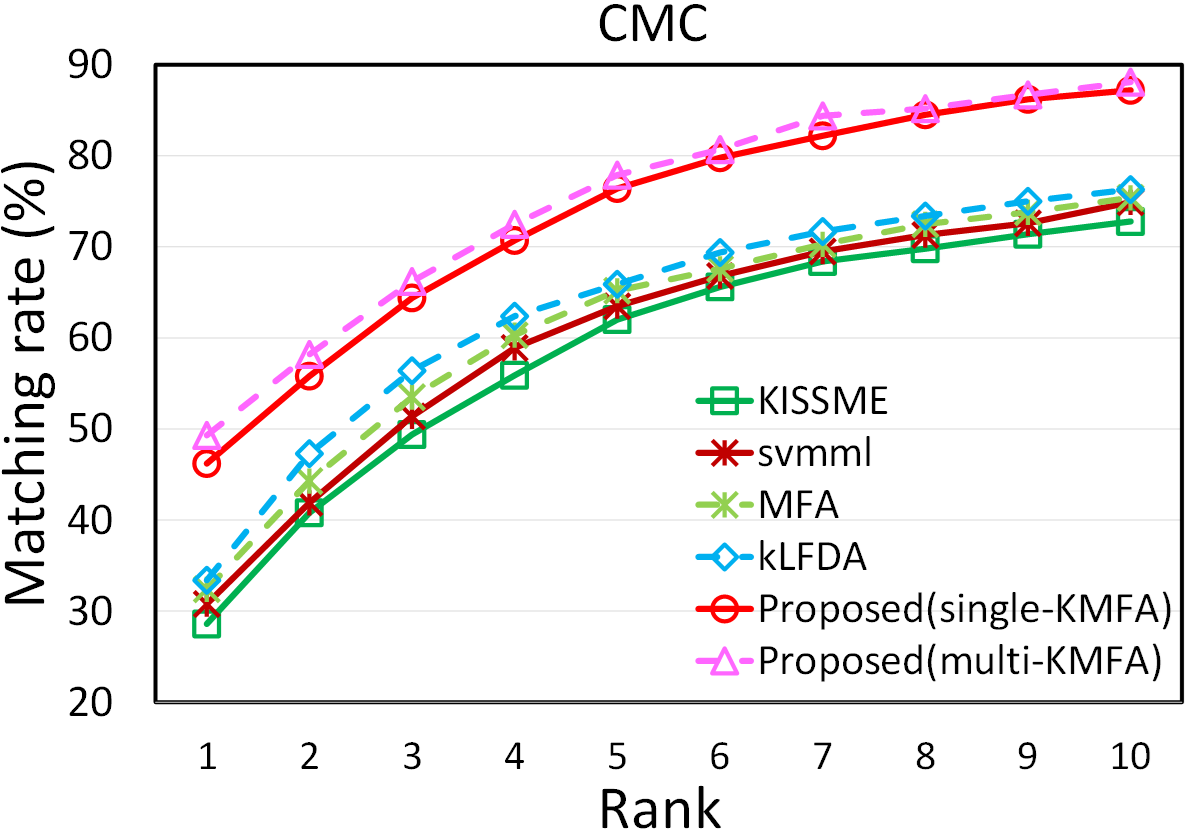} \label{fig:cmc curve1 e}}
  \caption{CMC curve comparison on different datasets. (a): VIPeR, (b): PRID 450S, (c): 3DPeS, (d): Road, (e): SYSU-sReID.}  \label{fig:cmc curve1}
\end{figure*}

Note that since the rotation and orientation parameters used in the TA+W method~\cite{added[3]} are derived from camera calibration and human motion information which is not available on SYSU-sReID dataset, we manually label these parameters so as to have a fair comparison with our approach (i.e., guarantee that \cite{added[3]}'s performance will not be degraded due to improper rotation and orientation parameters).

The CMC results of different methods are shown in Tables~\ref{tab:cmcTable1}--\ref{tab:cmcTable5} and Fig.~\ref{fig:cmc curve1}. Moreover, since many works also reported fusion results on the VIPeR dataset (i.e., adding multiple Re-ID results together to achieve a higher Re-ID result)~\cite{6,8,color_in}, we also show a fusion result of our approach (\emph{svmml+Proposed+multi-manu-KMFA}) and compare it with the other fusion results in Table~\ref{tab:cmcTable1}. Tables~\ref{tab:cmcTable1}--\ref{tab:cmcTable5} show that:

\begin{enumerate}
 \item Our approach has better Re-ID performances than most of the state-of-the-art methods. This further demonstrates the effectiveness of our approach.
 \item Our approach also achieves similar or better results than some deep-neural-network-based methods (e.g., IDLA~\cite{ahmed2015cvpr}, JLR~\cite{faqiang2016cvpr}, and DG-Dropout~\cite{xiao2016learning} in Table~\ref{tab:cmcTable1}). This implies that spatial misalignment is indeed a significant factor affecting the performance of person Re-ID. If the spatial misalignment problem can be properly handled, satisfactory results can be achieved with relatively simple hand-crafted features. Moreover, since our approach is independent of the specific features used for Re-ID, it can easily accommodate future advances by combining with the deep-neural-network-based features (such as~\cite{wu2016enhanced}) without changing the core protocols.
 \item Since the TA+W method~\cite{added[3]} creates multiple weight matrices for pedestrians with different orientations, it can be viewed as another version of using multiple matching structures. However, our approach (\emph{proposed+multi-manu}) still outperforms the TA+W method~\cite{added[3]} (cf. Table~\ref{tab:cmcTable5}). This further indicates that our multi-structure scheme is effective in modeling the cross-view spatial correspondence.
\end{enumerate}

\subsection{Time complexity \& Effects of different parameter values}

\subsubsection{Time complexity} We also evaluate the running time of training \& testing excluding dense feature extraction in Table~\ref{tab:cmcTable6}. Testing process with (\emph{Test+Proposed}) and without (\emph{Test+Non-global}) Hungarian method are both evaluated. Moreover, for the testing process, we list two time complexity values: (1) the running time for the entire process (\emph{Test-all image pairs}), and (2) the average running time for computing the similarity of a single image pair (\emph{Test-per image pair}).

We can notice that the running time of both training and testing are acceptable. Table~\ref{tab:cmcTable6} also reveals that Hungarian method takes the major time cost of testing. This implies that our testing process can be further optimized by replacing Hungarian method with other more efficient algorithms~\cite{mills2007dynamic}.

\begin{table*}
 \centering
 \caption{Running time on five datasets (Evaluated on a PC with 4-core CPU and 2G RAM; \emph{hr, min, sec, ms} refer to \emph{hour, minute, second} and \emph{millisecond}, respectively.)} \label{tab:cmcTable6}
 \footnotesize
 \begin{tabular}{|l|*{5}{r@{.}l@{\;}l|}}
  \hline
  \textbf{Datasets}& \multicolumn{3}{c|}{VIPeR} & \multicolumn{3}{c|}{PRID 450S} & \multicolumn{3}{c|}{3DPeS}  & \multicolumn{3}{c|}{Road} & \multicolumn{3}{c|}{SYSU-sReID}\\ \hline
  Train-all image pairs & 2&06&hr  & 1&22&hr   & 0&89&hr   & 1&07&hr  & 1&43&hr \\
  Test-all image pairs (Non-global) & 57&97&sec  & 34&64&sec  & 24&87&sec  & 29&45&sec  & 39&06&sec\\
  Test-per image pair (Non-global) &0&58&ms  & 0&68&ms   & 0&52&ms   & 0&68&ms  & 0&62&ms\\
  Test-all image pairs (Proposed) & 6&64&min   & 3&49&min   & 2&61&min    & 3&03&min  & 4&28&min\\
  Test-per image pair (Proposed) & 3&99&ms  & 4&14&ms   & 3&26&ms   & 4&20&ms  & 4&08&ms\\
  \hline
 \end{tabular}
\end{table*}


\begin{figure}[t]
  \centering
  \subfloat[Results with different $T_c$ \& $T_d$]{\includegraphics[width=3.8cm,height=2.4cm]{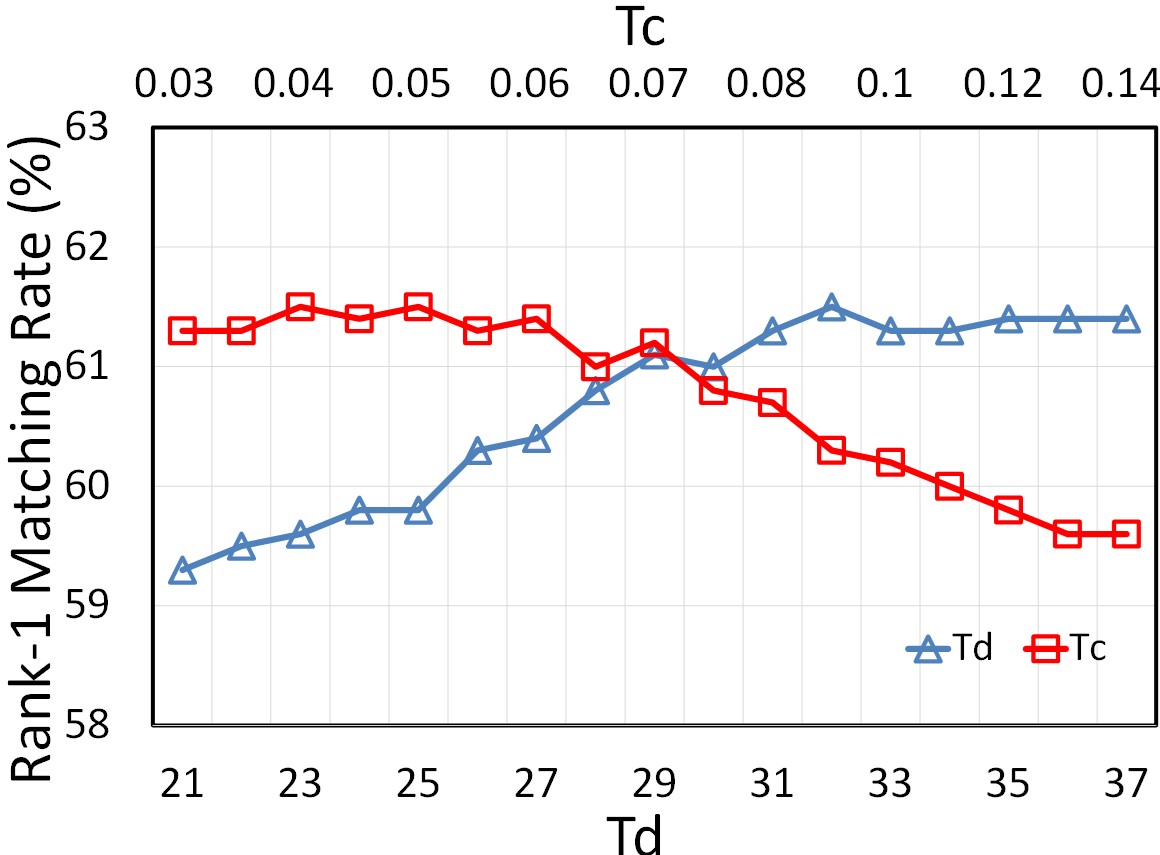} \label{fig:parameter evaluation d}}
  \hspace{2mm}
  \subfloat[Results with different patch sizes]{\includegraphics[width=4.0cm,height=2.1cm]{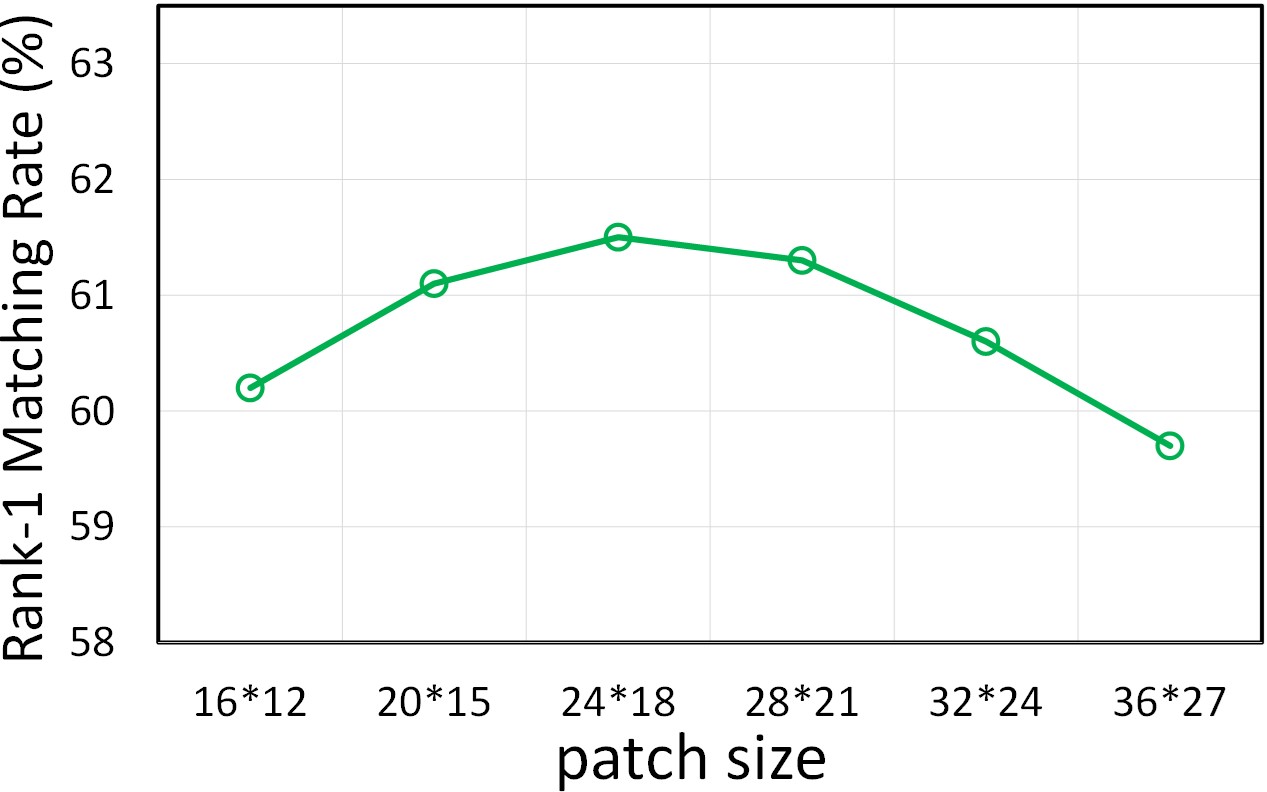} \label{fig:parameter evaluation c}}
  \caption{ The effect of different parameter values (The results are evaluated by the rank-1 CMC score on the Road dataset using KISSME distance metric).} \label{fig:parameter evaluation}
\end{figure}

\subsubsection{Effect of different parameter values} Finally, we evaluate the effect of three parameters: (1) $T_c$ which is the threshold of keeping the learned matching probability value $P_{ij}$ (cf. Eq.~\ref{equation:equ1}); (2) $T_d$ which is the search range of handling spatial displacement (cf. Eq.~\ref{equation:equ5a}); and (3) Patch size which is the size of the basic element for calculating image-wise similarity.

\begin{figure}[t]
  \centering
  \subfloat[$T_d=11$]{\includegraphics[width=4.25cm,height=2.3cm]{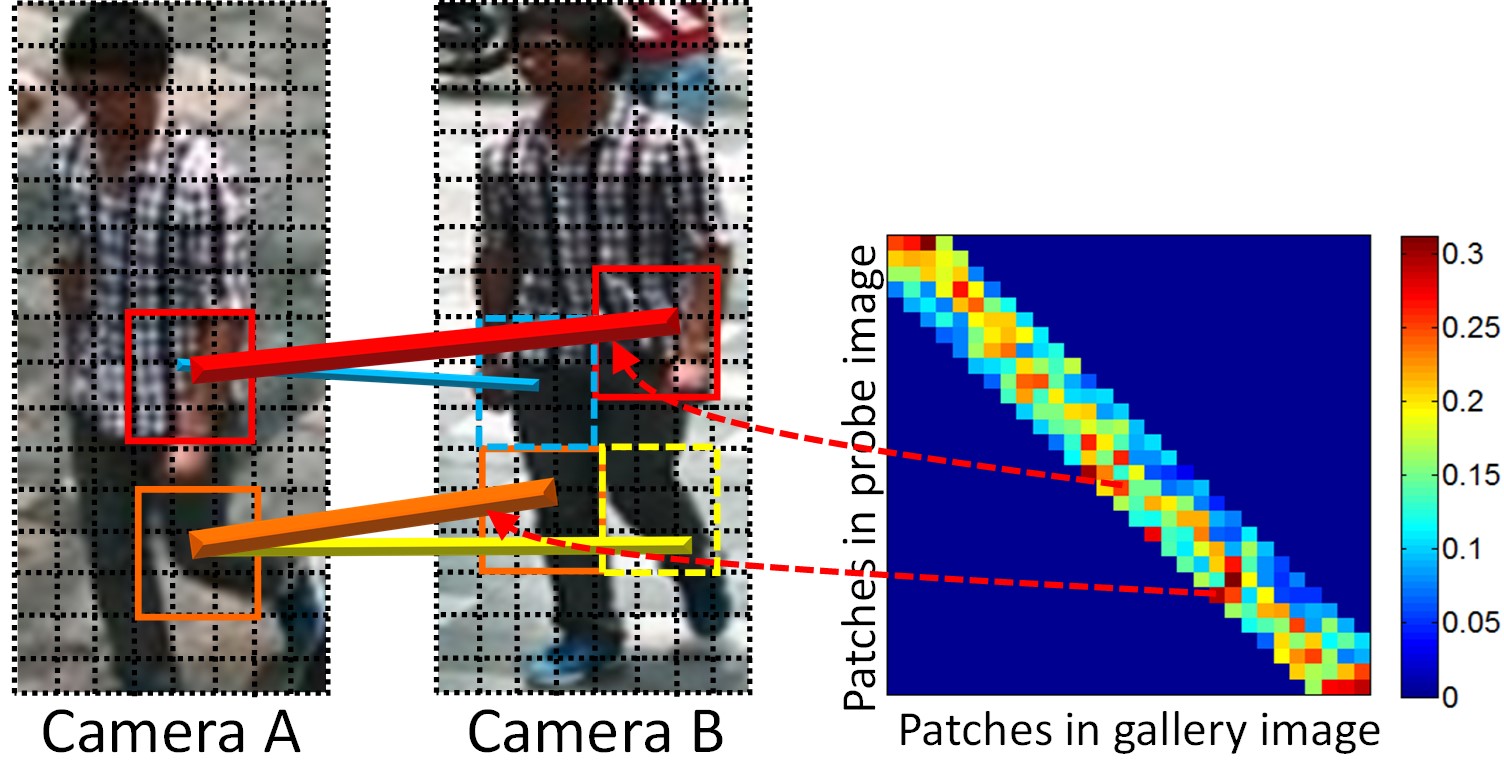} \label{fig:structures with various Td a}}
  \hspace{0mm}
  \subfloat[$T_d=32$]{\includegraphics[width=4.25cm,height=2.3cm]{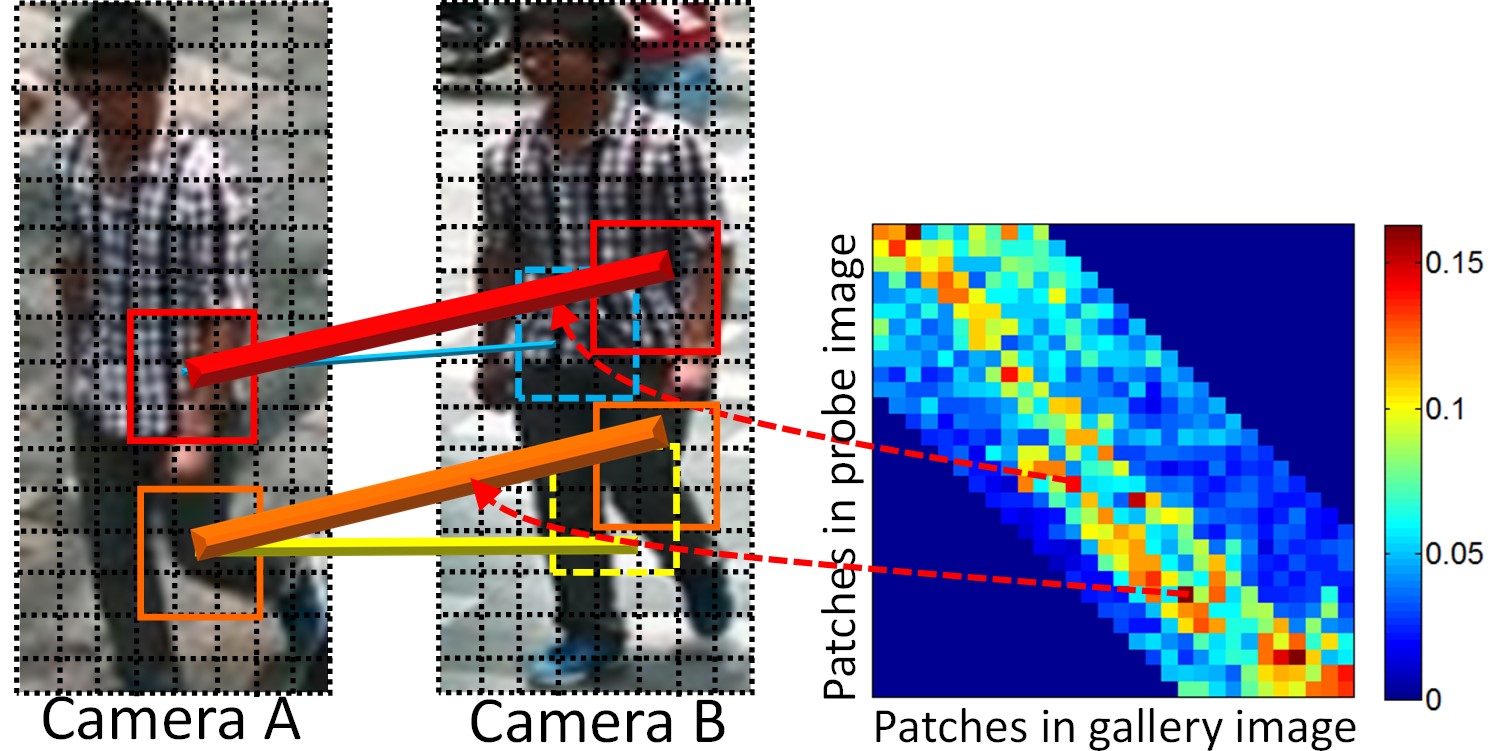} \label{fig:structures with various Td c}}
  \caption{Comparison of correspondence structures of the Road dataset with different search ranges $T_d$. (The correspondence structures are learned with the KISSME metric. Best viewed in color)} \label{fig:structures with various Td}
\end{figure}

Figures~\ref{fig:parameter evaluation d} to~\ref{fig:parameter evaluation c} show the rank-1 CMC scores~\cite{cmc} of our approach under different $T_c$, $T_d$, and patch size values on the Road dataset. Moreover, Fig.~\ref{fig:structures with various Td} further shows the learned correspondence structures under different search ranges $T_d$. Figures~\ref{fig:parameter evaluation} and~\ref{fig:structures with various Td} show that:

\begin{enumerate}
 \item Basically, $T_c$ should be kept small since a large $T_c$ will exclude many useful matching probabilities in the correspondence structure and lead to decreased Re-ID results. However, when $T_c$ becomes extremely small, the Re-ID performance may also be slightly affected since some noisy matching probabilities may be included in the correspondence structure. According to our experiments, 0.03---0.07 is a proper range for $T_c$ and we use $T_c=0.05$ throughout our experiments.
 \item $T_d$ normally should be set large enough in order to handle large spatial misalignments. However, a large $T_d$ will also increase the complexity of person Re-ID. According to our experiments, satisfactory results can be achieved when $T_d$ is larger than $32$. Therefore, in our experiments, we set $T_d=32$ to guarantee performance while minimizing computation complexity.
 \item The patch size should not be too small or too large. If the patch size is too small, the visual information included in each patch becomes very limited. This may reduce the reliability of patch matching results. On the other hand, if the patch size is too large, the number of patches in an image will be obviously reduced. This will affect the flexibility of our approach to handle spatial misalignments. According to our experiments, $20\times 15$---$28\times 21$ is a proper patch size range for  images with $128 \times 48$ sizes.
\end{enumerate}

\section{Conclusion} \label{section:conclusion}

In this paper, we propose a novel framework for addressing the problem of cross-view spatial misalignments in person Re-ID. Our framework consists of three key ingredients: 1) introducing the idea of correspondence structure and learning this structure via a novel boosting method to adapt to arbitrary camera configurations; 2) a constrained global matching step to control the patch-wise misalignments between images due to local appearance ambiguity; 3) a multi-structure strategy to handle spatial misalignments in a more precise way. Extensive experimental results on benchmark show that our approach achieves the state-of-the-art performance.

\bibliographystyle{IEEEtran}
\bibliography{egbib}


\end{document}